\documentclass[lettersize,journal]{IEEEtran}
\usepackage[linesnumbered,ruled,vlined]{algorithm2e}
\usepackage{graphicx}
\usepackage{makecell}
\usepackage{multirow} 
\usepackage{tikz}
\usepackage{amsmath,amsthm}
\usepackage{subfigure}
\usepackage{diagbox}
\usepackage{bm}
\usepackage{amsfonts}
\usepackage{amssymb}
\usepackage{verbatim}
\usepackage{url}
\usepackage{booktabs}

\usepackage{rotating}
\usepackage{color}
\usepackage{xcolor}
\usepackage{colortbl} 
\usepackage[switch]{lineno}
\usepackage[OT1]{fontenc} 
\usepackage{filecontents}
\usepackage{mathrsfs}
\usepackage{mdframed}
\usepackage{float}
\usepackage[most]{tcolorbox}
\usepackage{enumitem} 

\usepackage{hhline}
\usepackage[colorlinks,linkcolor=blue,citecolor=blue,urlcolor=blue]{hyperref}

\newenvironment{packeditemize}{
	\begin{list}{$\bullet$}{
			\setlength{\labelwidth}{4pt}
			\setlength{\itemsep}{0pt}
			\setlength{\leftmargin}{\labelwidth}
			\addtolength{\leftmargin}{\labelsep}
			\setlength{\parindent}{0pt}
			\setlength{\listparindent}{\parindent}
			\setlength{\parsep}{0pt}
			\setlength{\topsep}{1pt}}}{\end{list}}

\definecolor{light_cyan}{rgb}{0.53, 0.75, 0.77}
\definecolor{light_blue}{rgb}{0.466, 0.655, 0.94}
\definecolor{light_pink}{rgb}{0.96, 0.55, 0.565}
\definecolor{light_yellow}{rgb}{0.96, 0.83, 0.32}

\AtBeginDocument{%
  }

\begin{document}

\title{Parallel Unlearning in Inherited Model Networks}

\author{Xiao~Liu,
        Mingyuan~Li,
        Guangsheng~Yu,
        Lixiang~Li,
        Haipeng~Peng,
        and~Ren~Ping~Liu
        
\IEEEcompsocitemizethanks{
\IEEEcompsocthanksitem This work was supported in part by the National Key Research and Development Program of China under Grant 2024YFB2906503, in part by the National Natural Science Foundation of China under Grant 62032002, by the 111 Project under Grant B21049, and by the Doctoral Student Innovation Fund under Grant CX2023217. \textit{(Corresponding authors:
Guangsheng Yu; Lixiang Li.)}
\IEEEcompsocthanksitem X. Liu, M. Li, L. Li, and H. Peng are with Information Security Center, State Key Laboratory of Networking and Switching Technology, Beijing University of Posts and Telecommunications, Beijing 100876, China. \protect
E-mail: \{liuxiao68, henryli\_i, lixiang, penghaipeng\}@bupt.edu.cn
\IEEEcompsocthanksitem G. Yu was with CSIRO Data61, Sydney, Australia, at the time of writing. \protect
E-mail: reimusaber@gmail.com
\IEEEcompsocthanksitem R. P. Liu is with the Global Big Data Technologies Centre, University of Technology Sydney, Australia. \protect
E-mail: renping.liu@uts.edu.au
\IEEEcompsocthanksitem The affiliations of Guangsheng Yu and Ren Ping Liu reflect their institutions at the time of primary contribution and the contribution period prior to the updated Australian research security guidelines (2024–2025). There has been no further involvement following their transition to current positions, and they did not receive, directly or indirectly, any funding, compensation, in-kind assistance, material support, or resources from their affiliated institutions for this study, nor from these grants or any other source.


}
}


\maketitle

\begin{abstract}
Unlearning is challenging in generic learning frameworks with the continuous growth and updates of models exhibiting complex inheritance relationships. This paper presents a novel unlearning framework that enables fully parallel unlearning among models exhibiting inheritance.
We use a chronologically Directed Acyclic Graph (DAG) to capture various unlearning scenarios occurring in model inheritance networks. 
Central to our framework is the Fisher Inheritance Unlearning (FIUn) method, designed to enable efficient parallel unlearning within the DAG. FIUn utilizes the Fisher Information Matrix (FIM) to assess the significance of model parameters for unlearning tasks and adjusts them accordingly. To handle multiple unlearning requests simultaneously, we propose the Merging-FIM (MFIM) function, which consolidates FIMs from multiple upstream models into a unified matrix. This design supports all unlearning scenarios captured by the DAG, enabling one-shot removal of inherited knowledge while significantly reducing computational overhead.
Experiments confirm the effectiveness of our unlearning framework. For single-class tasks, it achieves complete unlearning with \textbf{0\%} accuracy for unlearned labels while maintaining \textbf{94.53\%} accuracy for retained labels. For multi-class tasks, the accuracy is \textbf{1.07\%} for unlearned labels and \textbf{84.77\%} for retained labels. Our framework accelerates unlearning by \textbf{99\%} compared to alternative methods.
\end{abstract}


\section{Introduction}
\label{section:1}

In the rapidly evolving landscape of data and application complexity, models require frequent updates to adapt to new data, constraints, and standards, highlighting the critical need for robust data privacy measures, auditing of illicit information, and adherence to regulatory and industry standards~\cite{jayaraman}. 
The process of ``model forgetting'' or ``unlearning'' emerges as essential, involving the removal of specific information from trained models to address privacy concerns, ensuring models do not retain or utilize sensitive data in violation of regulations like the General Data Protection Regulation (GDPR)~\cite{voigt2017eu}. Moreover, unlearning facilitates data correction by eliminating the influence of erroneous or biased inputs, thereby not only protecting user privacy but also enhancing the model's accuracy, fairness, and overall quality.

Many existing unlearning advances (e.g.,~\cite{wang2025gru,yang2025exploring}) typically focus on removing the influence of specific data from a single model, without considering relationships among models. However, in an Inherited Model Network (IMN), where models are organized as a Directed Acyclic Graph (DAG), unlearning becomes significantly more complex: updates to one model must be consistently propagated to all its downstream inherited models. This structure introduces cascading effects and computational challenges that are unique to IMNs. This challenge arises in scenarios such as customizing a base model to meet various Machine Learning Operations (MLOps) lifecycle requirements~\cite{mbma} or distilling it to fit computational resource constraints~\cite{polino2018model}. These situations can be abstracted into prevalent learning frameworks such as Federated Learning~(FL)~\cite{li2020review}, Distributed Data-Parallel Learning (DDPL)~\cite{ddap}, Incremental Learning (IL)~\cite{IL}, and Transfer Learning (TL)~\cite{TL}. In such cases, unlearning even a single model necessitates unlearning the entire subgraph rooted at the origin to ensure that all descendant models remain consistent and relevant.

\noindent\textbf{Research gaps.} 
Various machine unlearning techniques have been proposed. 
Unfortunately, the inheritance relationships among related models prevent most existing unlearning methods from being executed in parallel, including re-training~\cite{retrain}, gradient ascent~\cite{ga}, or knowledge distillation methods~\cite{distil1Unl}.
On the other hand, Sharding, Isolation, Slicing, and Aggregated Training (SISA)~\cite{sisa,yu2024splitunlearning} is a commonly used efficient method that reduces the dependence between data by dividing it into multiple segments and training multiple models in parallel. However, SISA is also inapplicable when there are inheritance relationships between models~\cite{liu2024}. 
This is because SISA needs to be performed sequentially from the starting point of updating to the end of the subgraph, incurring significant overhead and resource consumption during unlearning.
In this sense, there is a pressing need for an \underline{efficient} (i.e., {\textit{unlearning speed}}) and \underline{effective} (i.e., \textit{accuracy}) way to perform unlearning in the presence of model inheritance relationships.

\noindent\textbf{Research questions.} To address this research gap, we need to identify the commonalities across various prevalent frameworks in unlearning and determine how a new unlearning design can be universally applied. We raise two key Research Questions (RQs):

\smallskip
\smallskip

\textbf{RQ1.} \textit{How can unlearning requests be quickly allocated to models that need updates and what unique features could be realized during unlearning in a large-scale and interdependent model network?}

\smallskip
\textbf{RQ2.} 
\textit{How can the model unlearning tasks be efficiently and effectively executed, especially when the unlearning of one model impacts subsequent models?}

\smallskip
\smallskip

In response to these two RQs, we investigate the topological structure and unlearning impact of inherited models in four prevalent frameworks: FL, DDPL, IL, and TL. Specifically, we study the key factors affecting the efficiency, accuracy, and consistency of model unlearning. These include the selection of model parameters, changes in data distribution, and utilization of computational resources.

\smallskip
\noindent\textbf{Contributions.}
{\color{black}This paper introduces a novel parallel unlearning framework for efficiently handling unlearning across numerous dependent models with complex inheritance relations, such as in FL, DDPL, IL, and TL.
The contributions are summarized as follows:

\begin{packeditemize}
    \item We capture various unlearning scenarios in IMNs, which are chronologically DAG-based. By analyzing scenarios with single- and multi-root unlearning requests, and the impact of path depths and multi-path structures on unlearning, we provide precise and efficient identification of models to be unlearned.

     \item We introduce the Fisher Inheritance Unlearning (FIUn) method to enable efficient parallel unlearning within DAGs. FIUn compares the Fisher Information Matrices (FIMs) of the initial unlearning models (upon the data to be unlearned) with the FIMs of all inherited models, identifies the model parameters that require updates, and then proportionally adjusts parameters in parallel across the inherited models based on their unlearning impact.

\item To handle multiple unlearning requests simultaneously, FIUn incorporates a Merging-FIM (MFIM) function that consolidates FIMs from multiple upstream models into a single unified matrix. This allows FIUn to address all unlearning scenarios, facilitating one-shot removal of inherited knowledge while greatly reducing computational overhead.

\end{packeditemize}

Extensive experiments indicates that our method significantly outperforms all comparison methods in terms of unlearning speed on widely-accepted public image and text datasets. Compared to re-training, our method achieves comparable levels of unlearning accuracy. Even on large-scale models, our method demonstrates a certain degree of effectiveness and applicability.
A preliminary version of this work has been posted on arXiv.\footnote{This work is based on a preprint post: \url{https://arxiv.org/abs/2408.08493}.}
}

The rest of this paper is organized as follows. The related works are reviewed in Section~\ref{sec:related work}. The preliminaries are in Section~\ref{sec:preliminaries}.
The explored topological structure frameworks are discussed in Section~\ref{sec:paradigm}. The designed parallel unlearning method in Section~\ref{sec:method}. The experimental settings are presented in Section~\ref{sec:experimental settings}, followed by the evaluation in Section~\ref{sec:experimental results}. Section~\ref{sec:conclusion}  concludes this work.

\section{Related Work}\label{sec:related work}

Although FL~\cite{li2020federated}, DDPL~\cite{ddap}, IL~\cite{IL}, and TL~\cite{TL} have addressed challenges such as data privacy~\cite{li2021survey}, training efficiency~\cite{dean2012large}, knowledge retention and unlearning~\cite{han2023anomaly}, and cross-domain knowledge transfer~\cite{rajapaksha2023improving}, systematic exploration combining these frameworks is limited. Most studies have only attempted simple method combinations without addressing their shared foundation model inheritance.
This aspect is crucial in large-scale unlearning, where computational overhead and time consumption increase with inheritance depth, even when unlearning is parallelized at the same depth. This remains true across existing unlearning methods, which are discussed below.

\smallskip
\noindent\textbf{Federated Unlearning.} 
Wang et al.~\cite{wang2024server} proposed a server-initiated federated unlearning framework, SIFU, which identifies and quantifies the impact of low-quality client data. By combining gradient
ascent with high-quality data augmentation, the method efficiently eliminates the negative influence of such data on the global model. Tao et al.~\cite{tao2024communication} introduced the Total Variation (TV) Stability to measure model sensitivity to small data perturbations and modified the FedAvg algorithm to improve communication efficiency while enabling precise client- and sample-level unlearning. To address performance degradation caused by gradient conflicts, Pan et al.~\cite{pan2025federated} proposed the FedOSD method, which designs a dedicated unlearning loss function and optimized the model along the steepest descent
direction closest to the target client’s gradient, without interfering with others, achieving a balance between unlearning and utility. Gao et al.~\cite{gao2024verifi} introduced the VeriFi framework, a framework that supports active verification of unlearning effects and includes a more efficient unlearning algorithm,
uS2U, along with two non-invasive verification methods, vFM and vEM, enhancing the security and
verifiability of federated unlearning.

%


\smallskip 
\noindent\textbf{Distributed Data-Parallel Unlearning.} In DDPL knowledge unlearning, handling the unlearning process is restricted to the data shard on the specific device, followed by model aggregation and iterative training. If unlearning is required from a specific shard, only the model trained on that shard is re-trained. Approximate machine unlearning directly modifies trained model parameters, employing methods such as 1) Gradient-based approaches~\cite{golatkar2020eternal}, which update model gradients to gradually diminish the data's influence, and 2) Data perturbation~\cite{tarun2023fast}, which introduces noise to reduce specific data points' impact on the model. These methods facilitate effective knowledge unlearning in DDPL.

\smallskip
\noindent\textbf{Incremental Unlearning.}
Research on unlearning in IL is still a developing area, but some recent methods show promise. Projected-Gradient Unlearning (PGU)~\cite{hoang2024learn} calculates gradients for the unlearning dataset, projects them onto the Core Gradient Space (CGS) derived from the training set, and subtracts the projected component. The residuals, now orthogonal to the CGS, are used to update model parameters. Effective for one-time and batch-wise unlearning, PGU is well-suited for incremental learning processes.
Zuo et al.~\cite{zuo2024ecil} proposed a class-level unlearning method called eCIL-MU, which uses embedding to
map data into vectors and store them in a vector database. This enables efficient and non-destructive
unlearning by reusing the structure of class-incremental learning tasks.

\smallskip
\noindent\textbf{Transfer Unlearning.} Research on unlearning specific target task knowledge has been relatively scarce in TL. To address this challenge, Sepahvand et al.~\cite{sepahvand2024data} proposed fine-tuning pre-trained models through data selection. Subsequently, the entire network was fine-tuned on the target task using gradient descent.

\noindent\textbf{Machine Unlearning.}
Nguyen et al.~\cite{nguyen2020variational} proposed a loss function minimizing Kullback-Leibler divergence to ensure certified data removal in Bayesian models, using two techniques to correct posterior inaccuracies and prevent catastrophic forgetting for small data subsets. Ginart et al.~\cite{ginart2019making} introduced Q-k-means and DC-k-means algorithms for k-means clustering, using quantized cluster centers and divide-and-conquer to efficiently update models without retraining, ensuring precise data removal while maintaining clustering quality.
Kurmanji et al.~\cite{kurmanji2023towards} presented SCRUB, a teacher-student model optimizing student weights to diverge from the teacher's output on the forget dataset ($D_f$) while aligning with the retain dataset ($D_r$), balancing forgetting and utility for bias removal, confusion resolution, and privacy. Zhu et al.~\cite{zhu2024decoupling} proposed TARF, using annealed gradient ascent and target-aware gradient descent to decouple specific concepts, preserving performance on unaffected data. Both modified model weights for efficient approximate unlearning without strict certified removal.

\smallskip
\noindent\textbf{Ours.}
Existing unlearning methods struggle with high computational costs, resource demands, and implementation challenges, particularly in large-scale settings. Our FIUn method enables fully parallel unlearning by adapting single-model techniques~\cite{foster2024fast,golatkar2020eternal} for IMNs, improving efficiency, resource utilization, and adaptability while achieving effectiveness comparable to re-training.

\section{Preliminaries}
\label{sec:preliminaries}
This section introduces the concepts, notation, and formulas of this paper, including machine unlearning and FIM.

\subsection{Machine Unlearning}
Machine unlearning refers to removing the influence of specific data subsets from a trained model while maintaining its performance on the remaining data~\cite{foster2024zero}. 

Let $D = \{(x_i, y_i)\}_{i=1}^N$ denote the full training dataset, where $x_i$ is the $i$-th input and $y_i \in \{1, \dots, K\}$ is its corresponding label. Let $D^f \subset D$ be the subset of samples to be unlearned (forget set), and $D^r = D \setminus D^f$ be the remaining samples (retain set). Let $\phi_\theta : X \rightarrow Y$ be the model parameterized by $\theta \in \mathbb{R}^m$, and $\phi_{\theta'}$ be the updated model after unlearning.
The unlearning process is considered successful if:
(i) $\phi_{\theta'} \perp D^f$, meaning that the predictions of $\phi_{\theta'}$ are statistically independent of the removed data $D^f$, ideally resembling those of a model trained only on $D^r$;
(ii) $L(\phi_{\theta'}, D^r) \approx L(\phi_\theta, D^r)$, ensuring the model retains performance on the remaining data.

In practice, this entails adjusting the model weights $\phi_\theta$ to obtain $\phi_{\theta'}$ that removes the influence of $D^f$ while preserving knowledge from $D^r$. Crucially, model parameters are inherently heterogeneous—different weights contribute unequally to different classes, samples, or tasks~\cite{koh2017understanding}. Effective unlearning therefore requires selective and targeted parameter adaptation. This motivates a range of gradient-based and sensitivity-aware approaches—including influence functions, Fisher Information, and other parameter importance metrics—that identify and update only the most relevant components in a principled and efficient manner.

\subsection{Fisher Information Matrix}
FIM is a useful tool in statistics for assessing the accuracy of parameter estimates~\cite{golatkar2020eternal}. It quantifies the impact of parameter changes on the probability distribution of observed data. 
Given a probability density function $p\left( x|\theta  \right) $ conditioned on the model parameter $\theta $, an FIM
$I\left( \theta  \right) $ is the expected value of the second derivative of the negative log-likelihood function $\ell\left( \theta  \right)$, as given by
\begin{equation}\label{eq.FIM1}
I\left( \theta  \right) =\mathbb{E} \left[ -\frac{\partial^{2} \ell \left( \theta  \right) }{\partial \theta^{2} }  \right] ,\  \text{where}\  \ell \left( \theta  \right) =\ln p\left( x|\theta  \right) .
\end{equation}

To simplify computation, in practice, the diagonal of FIM can approximate the second derivative of the loss function~\cite{steven1993fundamentals}. An equivalent form of $I\left( \theta  \right) $ is given by \begin{equation}\label{eq.FIM2}
I(\theta) =\mathbb{E} \left[ \left( \left( \frac{\partial \ell \left( \theta  \right) }{\partial \theta }  \right) \left( \frac{\partial \ell \left( \theta  \right) }{\partial \theta }  \right)^{\top }  \right)  \right] \  .
\end{equation}

The FIM holds significant importance in the theory of machine unlearning. The natural gradient regards the FIM as a Riemannian metric tensor of the parameter space, used to define the optimal direction of parameter perturbation~\cite{amari1998natural}. Furthermore, the FIM is the expected form of the Hessian, aligning with the concept of influence functions for analyzing the impact of individual samples~\cite{koh2017understanding}.
From an information-theoretic perspective, a larger FIM value indicates that the corresponding parameter carries more discriminative information about the training data and is therefore more critical to the model’s behavior.

An intuitive baseline is to simply remove the columns in the last FC layer weights corresponding to the classes to be unlearned. However, this approach has significant limitations: (i) it is only applicable to class-level unlearning and cannot be extended to sample-level unlearning tasks; (ii) simple eliminating weight columns disrupts the decision boundaries of the retained data, leading to substantial accuracy degradation~\cite{lee2023undo}; and (iii) in IMNs, this method only affects the current node and fails to ensure consistent propagation of unlearning information across all inherited models.
To address these issues, this paper employs the FIM to quantify the importance of model parameters to the training data. By comparing the FIMs of the training dataset and the unlearning dataset, we can identify the critical parameters related to the knowledge to be unlearned.

\section{Capturing Model-Inheritance}\label{sec:paradigm}

In this section, we explore the topological structure of four prevalent frameworks, i.e., FL, DDPL, IL, and TL, and analyze the impact of unlearning on the inherited models within these frameworks. We propose using DAG, which can visualize and capture the model inheritance and update relationships in these learning frameworks in a unified manner.

\subsection{Diverse Learning Frameworks and Tasks}

\begin{figure}[t]
 \centering
 \begin{subfigure}[FL vanilla and DDPL]{   \centering\includegraphics[width=0.2\textwidth,height=1.98cm]{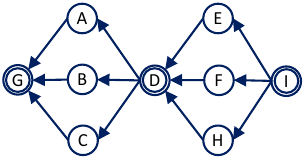}
         \label{fig:FL}    
     }
     \end{subfigure}
     \begin{subfigure}[Multi-layer/server FL]{
\centering\includegraphics[width=0.185\textwidth,height=2.19cm]{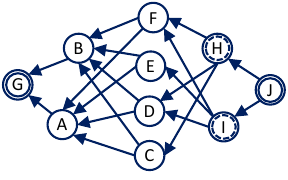}
\label{fig:multilayer}
     }
     \end{subfigure}
     \begin{subfigure}[Decentralized FL]{
        \centering
\includegraphics[width=0.21\textwidth,height=2cm]{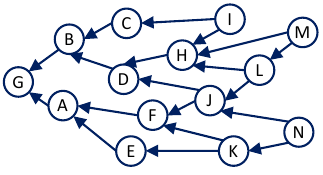}
     \label{fig:DAG_FL}
    }
     \end{subfigure}
      \begin{subfigure}[IL and TL]{
         \centering
\includegraphics[width=0.13\textwidth,height=1.9cm]{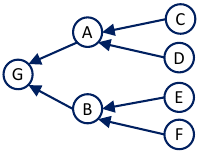}
         \label{fig:IL}
     }
     \end{subfigure}
     
      \vspace{0.3cm}
    \includegraphics[width=0.4\textwidth]{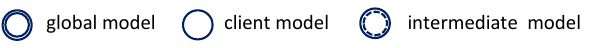}

  \caption{\small These diagrams show learning frameworks modeled as a DAG, where one user may create multiple models, and multiple models may be created by different users.}
  \label{fig:total_Experi_Scenarios}
\end{figure}

\noindent\textbf{Federated Learning.} A central server creates a global model, which participants download to locally update the model parameters based on their local data. The participants then send back their updated parameters.  The central server aggregates the parameters to update and redistribute the global model. This process repeats until the model converges~\cite{krauss2024automatic}. As shown in Fig.~\ref{fig:FL}, the topology of FL consists of nodes representing model states and edges representing parameter transmission and aggregation. Updates to model $A$ trigger updates in the global model $D$ and then to $E$, $F$, $H$, and~$I$. 

 \smallskip
\noindent\textbf{Multi-Layer FL.} This variant introduces an intermediate aggregation layer to traditional FL~\cite{li2021survey}, as depicted in Fig.~\ref{fig:multilayer}. This approach adds multiple servers that perform intermediate aggregation between the clients and the global server, thus reducing the burden on the global server, improving system scalability, and enhancing model training.

 \smallskip
\noindent\textbf{Decenralized FL.} As shown in Fig.~\ref{fig:DAG_FL}, multiple clients update and aggregate models hierarchically in DAG-FL~\cite{yu2023ironforge}. Each node can receive model updates from multiple parent nodes and send the updated results to multiple child nodes. This structure enables efficient model training and more flexible communication, better adapting to complex network environments and different application needs.

  \smallskip
\noindent\textbf{Distributed Data-Parallel Learning.} Training tasks are divided into sub-tasks and distributed across computing nodes or servers~\cite{liu2024fishers,dean2012large}. Each node processes its sub-task independently and periodically synchronizes parameters to update the global model. This iterative process is essential in DDPL topologies—similar to the FL topology in Fig.~\ref{fig:FL}—where sub-tasks $E$, $F$, and $H$ and the aggregated model $I$ must be updated with every new iteration of the global model $D$.


  \smallskip
\noindent\textbf{Incremental Learning.} IL entails training a model initially on a dataset and then progressively updating it as new data arrives, ensuring adaptability to the latest information~\cite{han2023anomaly}. IL can be visualized as shown in Fig.~\ref{fig:IL}, where nodes represent different updated states of the model, and edges depict the transition between model states after data updates. If the initial state model $G$ continues to update, then the subsequent models $A$ and $B$ that receive new data, as well as their related models, also need updates.

\smallskip
\noindent\textbf{Transfer Learning.}
TL pre-trains a base model on one dataset and adapts it to a new task by fine-tuning, typically on the last layers, using task-specific data~\cite{rajapaksha2023improving}. Through continued fine-tuning, the model aligns with the new task and achieves optimal performance. The TL topology mirrors the IL topology in Fig.~\ref{fig:IL}, following a similar update process.

\smallskip
In these learning frameworks, inheritance between models is crucial: when a base model is updated, all its child models must also be updated. We use a DAG topology to represent these dependencies, ensuring that updates to one model propagate through its entire subgraph of descendants.

\subsection{Inherited Model Networks}
\begin{figure}[t]
\centering
\includegraphics[width=6.5cm]{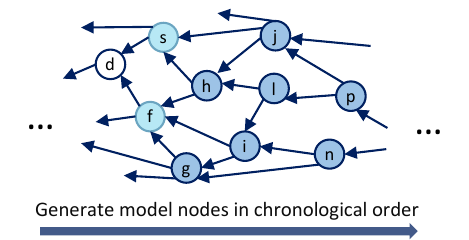}
	\caption{\small The dark blue nodes are all model nodes inheriting from the starting nodes $n_{s}$ and $n_{f}$ in light blue. A model node $n_{\ast}$ corresponds to a model $w_{\ast}$. $w_{s}$ and $w_{f}$ can originate from different users, with the possibility of simultaneous unlearning requests.}
 \label{fig:relationship}

\end{figure}

We use a DAG to represent and analyze model inheritance and updates across frameworks like FL, DDPL, IL, and TL. Time is integrated into the representation: each node corresponds to a publication moment, and edges follow the chronological order from left to right. This captures both inheritance dependencies and the temporal sequence of model publication, providing a clear view of complex model networks.


As shown in Fig.~\ref{fig:relationship}, we model the system as a weighted directed graph $\mathcal G = (N, E)$. The nodes $N={n_{1},n_{2},\ldots,n_{m}}$ represent model states or tasks, each corresponding to a set of parameters. The edges $E={e_{ij}}$ capture inheritance relations, such as parameter transfer or task inheritance. In FL and DDPL, edges may denote parameter transfer from local to global models, while in TL they represent task inheritance from pre-trained to fine-tuned models. Edge direction is determined by temporal order, reflecting when nodes are published, and by reference relations, indicating when one node explicitly refers to another.

Based on the source and state of the model, we further categorize the nodes into two types:

\begin{packeditemize}
\item \noindent\textbf{Discovery Nodes} are models trained on data with unlearned labels, allowing adaptation to updated classification capabilities within their DAG subgraph. Root model nodes $n_s$ and $n_f$ exemplify this type in Fig.~\ref{fig:relationship}.

\item \noindent\textbf{Inherited Nodes} are all non-discovery nodes in the subgraph that connect to discovery or other inherited nodes, focusing on classification refinement. Nodes $n_h,n_j,n_l,n_p,n_g,n_i,n_n$ belong to this type.
\end{packeditemize}

\subsection{DAG-Assisted Unlearning}\label{subsec:umig-unlearning}

With an emphasis on knowledge inheritance and unlearning tasks, DAG allows a model to retain and utilize previously learned useful information when receiving new data, achieving an effective combination of old and new knowledge. DAG also allows the model to remove specific data for continuous optimization and performance improvement.

The DAG-assisted unlearning consists primarily of the following two processes:
\begin{packeditemize}
    \item \noindent\textbf{Locating unlearning subgraphs. } When a user wants to unlearn certain data, an unlearn request is initiated. The system first identifies the nodes related to that user, which are referred to as discovery nodes, which then spawn the subgraphs associated with the discovery nodes.

    \item \noindent\textbf{Updating models within unlearning subgraphs.} All model nodes, including discovery and inherited nodes within the identified unlearning subgraphs, are updated to unlearn the parts of their models affected by the data.

\end{packeditemize}

These discovery nodes serve as the roots for the unlearning subgraphs. As depicted in Fig.~\ref{fig:relationship}, the unlearning subgraphs may overlap since inherited nodes may inherit from multiple discovery nodes. For example, the discovery nodes, $n_s$ and $n_f$, lead to two overlapping subgraphs with inherited nodes. Overlapping inherited nodes, such as $n_h, n_j, n_l, n_p$, would require the elimination of knowledge from multiple discovery nodes when unlearning due to their connections to different unlearning subgraphs. This situation is known as a \textbf{multi-root scenario}. In contrast, inherited nodes within one unlearning subgraph, such as $n_g, n_i, n_n$ in a \textbf{single-root scenario}, only need to eliminate the impact from a single discovery node.

Inheritance is not just a simple mapping from the discovery nodes to the inherited nodes. Instead, it is achieved through multiple paths and varying depths. Each path represents a route of knowledge transfer via iteration, and the number of paths and differences in depth determine the complexity of the inheritance relationship. 

A complex structure emerges when multiple paths are at varying depths from an inherited node to its corresponding discovery node(s), e.g., $n_f, n_g, n_i$ or $n_s, n_h, n_j$. If unlearning operations are conducted simultaneously in terms of depth (in other words, nodes at the same depth from the root(s) are processed simultaneously and rely on sequential dependencies), nodes such as $n_i$ and $n_j$ may require multiple updates. This situation is termed an \textbf{imbalanced-path scenario}, contrasting with a \textbf{balanced-path scenario} (e.g., $n_f, n_h, n_i, n_l$), where each node requires only a single update due to uniform path lengths and synchronous processing at each depth. In cases where unlearning is conducted asynchronously, regardless of depth, both scenarios may require multiple updates.

\smallskip 

\noindent\underline{\textbf{Answer to RQ1:}}
We use a DAG to visualize model inheritance. Each node marks a specific time, and edges follow the chronological order of model publication, giving a clear view of complex inheritance networks. The DAG also helps quickly identify models whose knowledge should be removed and systematically collects inheritance information to simplify unlearning. We analyze different learning frameworks, focusing on upstream models and inheritance paths to capture key unlearning scenarios: \textbf{single-root}, \textbf{multi-root}, \textbf{balanced-path}, and \textbf{imbalanced-path}.


\begin{figure*} [!ht]
        \centering
	\includegraphics[width=0.85\textwidth ]{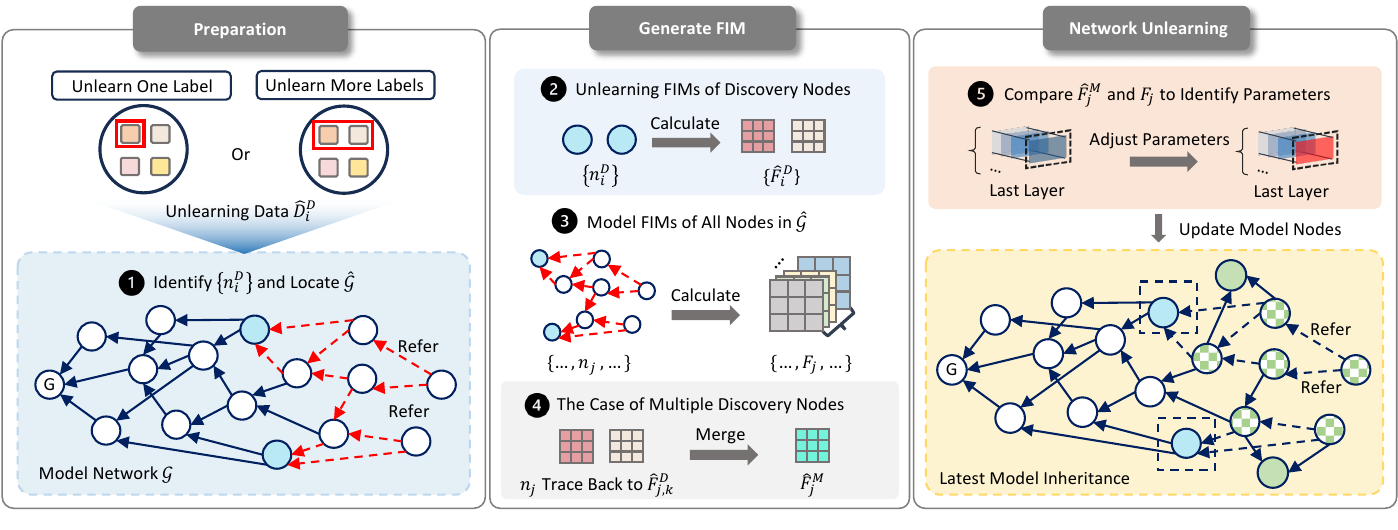}
	\caption{\small The unlearning process in $\mathcal G$ consists of five steps: 
 1) Preparation. The discovery nodes $\{n^{\scriptscriptstyle D}_i\}$ are identified, with unlearning data $\hat D^{\scriptscriptstyle D}_i$ containing one or more labels. Subsequently, the unlearning graph $\mathcal{\hat G}$ is located which root at $\{n^{\scriptscriptstyle D}_i\}$.
 2) Calculate unlearning the unlearning FIMs $\{\hat{F}_{i}^{\scriptscriptstyle D}\}$ of $\{n^{\scriptscriptstyle D}_i\}$.
 3) Calculate the model FIM $ F_{j}$ of node $n_j$ in $\mathcal{\hat G}$. 
 4) Merge unlearning FIMs $\hat F^{\scriptscriptstyle D}_{j,k}$ for $n_j$ in the case of multiple discovery nodes according to the topology. 
 5) Update $n_j$ by comparing the merged unlearning FIM $\hat{F}_{j}^{\scriptscriptstyle M}$ and $F_{j}$. Steps 2, 3, 4 and 5 apply to all nodes in $\mathcal{\hat G}$ .}
 
 \label{fig:system_model}

 \end{figure*}

\section{FIUn-Based Parallel Unlearning}\label{sec:method}
This section introduces a novel parallel unlearning method, Fisher Inheritance Unlearning (FIUn), designed to address the challenges associated with machine unlearning in large-scale DAG model networks.
A trusted learning network is considered where all learning contributors follow the designed unlearning process.
The FIUn method operates under the general expression provided by the DAG and leverages FIMs. It initiates the unlearning process by identifying specific model parameters that need updates, making the process more systematic. Derived from models in the unlearning graph, the training data, and the requested 
unlearning data, the FIMs are crucial in pinpointing essential model parameters and recommending their updated values. 

Notably, the FIUn method allows for the independent application of unlearning steps to each impacted model, facilitating a swift and effective parallel unlearning across extensive model networks.
The FIUn method is depicted in Fig.~\ref{fig:system_model}, and the procedural details are given in Algorithm~\ref{Algo.2}.

\begin{algorithm}[t]
\footnotesize
\caption{Fisher Inheritance Unlearning (FIUn)} 
\label{Algo.2}
\SetAlgoLined

\textbf{Input:} Model network $\mathcal G=(N,E)$, Unlearing task $\mathcal T$

\textbf{Parameter:} $\gamma$, $\tau$, $\eta$  

 Identify discovery nodes ${n^{\scriptscriptstyle D}_i}$ and the resulting unlearning graph $\mathcal {\hat G}$\\
\For{$n^{\scriptscriptstyle D}_i$ in $\{n^{\scriptscriptstyle D}_i\}$ parallel}
{
    Calculate unlearning FIM $\hat F^{\scriptscriptstyle D}_i$ with~\eqref{eq.FIM_unlearning}  
}

\For{$n_j$ in $\mathcal {\hat G}$ parallel}
{
    Calculate model FIM $F_j$ with~\eqref{eq.FIM_model}  \\
    Merge unlearning FIMs $\{\hat F^{\scriptscriptstyle D}_{j,k}\}$ with~\eqref{eq.merge} \\
    Update model parameters of node $n_j$ with~\eqref{eq.Identify}
}

\end{algorithm}


\subsection{FIUn}\label{subsec:fiun}

The main idea of FIUn is to identify critical model parameters by comparing the FIM from the unlearning dataset with the FIM from the training dataset. The differences between these matrices highlight the necessary adjustments in model parameters to achieve effective unlearning. The FIUn process unfolds as follows.

  \smallskip
\noindent\textbf{Preparation.}
Upon receiving an unlearning task $\mathcal{T}$ initiated by a user, FIUn first identifies the set of discovery nodes $\{n^{\scriptscriptstyle D}_i\}$ related to that user from the model network $\mathcal G$, with $n^{\scriptscriptstyle D}_i$ representing the $i$-th discovery node.
Subsequently, FIUn delineates the unlearning graph $\mathcal{\hat{G}}$, which is rooted at the identified discovery nodes. FIUn
then updates all nodes in the unlearning graph to unlearn the
requested unlearning knowledge.

  \smallskip
\noindent\textbf{Step-1: Calculate unlearning FIMs of discovery nodes.}
The FIUn method initiates by assessing the impact of the unlearning data on the model parameters of the discovery nodes, utilizing the FIM as specified in~\eqref{eq.FIM_unlearning}. This assessment is conducted using the model parameters and the unlearning data (see line 5, Algorithm~\ref{Algo.2}). 

The unlearning FIM for the $i$-th identified discovery node, $n^{\scriptscriptstyle D}_i$, is denoted by $\hat F^{\scriptscriptstyle D}_i$ and is calculated as 
\begin{equation}
\label{eq.FIM_unlearning}
\hat F^{\scriptscriptstyle D}_i\!\!=\!\mathbb{E} \!\left[ \!\left(\left( \frac{\partial \ln p(\hat D_{i}^{\scriptscriptstyle D}|w^{\scriptscriptstyle L}_{i} )}{\partial w^{\scriptscriptstyle L}_{i}}  \right)
\left( \frac{\partial \ln p(\hat D_{i}^{\scriptscriptstyle D}|w^{\scriptscriptstyle L}_{i})}{\partial w^{\scriptscriptstyle L}_{i} }\!  \right)^{\intercal }  \right)\Bigg|_{w^{\ast }_{\hat D_{i}^{\scriptscriptstyle D}} }\! \right] , 
\end{equation}
where $\hat D^{\scriptscriptstyle D}_i$ represents the unlearning data for $n^{\scriptscriptstyle D}_i$, $w^{\scriptscriptstyle L}_{i}$ denotes the last layer parameters of the model $w_i$ at node $n^{\scriptscriptstyle D}_i$, and $w^{\ast }_{\hat D_{i}^{\scriptscriptstyle D}}$ refers to the optimal parameters learned on $\hat D_{i}^{\scriptscriptstyle D}$.

The unlearning FIM highlights the importance of last-layer parameters for specific unlearning tasks \cite{lee2023layer}. In FIUn, FIM calculations are restricted to the last layer while freezing other layers, thereby reducing the computational complexity inherent in traditional FIM-based methods. Focusing on the last layer targets the most relevant features, as it directly influences the model’s final decision. This approach effectively facilitates unlearning goals while maintaining stable performance on unaffected data.

   \smallskip
\noindent\textbf{Step-2: Calculate model FIMs of all nodes in the unlearning graph.} The FIUn method proceeds by calculating the model FIMs of all models within the unlearning graph to identify the impact of model parameters in each node (see line~8, Algorithm~\ref{Algo.2}).

For any node $n_j$ in the unlearning graph $\mathcal{\hat{G}}$, the model FIM $F_{j}$ is given by
\begin{equation}\label{eq.FIM_model}
F_{j}\!=\!\mathbb{E} \!\left[\! \left( \left( \frac{\partial \ln p(D_{j}|w^{\scriptscriptstyle L}_{j} )}{\partial w^{\scriptscriptstyle L}_{j} }  \right) \left( \frac{\partial \ln p(D_{j}|w^{\scriptscriptstyle L}_{j} )}{\partial w^{\scriptscriptstyle L}_{j} }  \right)^{\intercal }  \right)\! \Bigg |_{w^{\ast }_{D_{j}} } \right]\!,
\end{equation}
where $D_j$ is the training data for $n_j$,  $w^{\scriptscriptstyle L}_{j}$ denotes the last layer parameters of the model $w_j$ at node $n_j$, and $w^{\ast }_{D_{j}}$ denotes the optimal parameters learned on $ D_{j}$.

The model FIM quantifies the knowledge about the training dataset embedded in the model parameters and highlights the significance of each parameter. Similar to the unlearning FIM, only the FIM for the last layer parameter is calculated, aiming to minimize computational complexity.

  \smallskip
\noindent\textbf{Step-3: Merge unlearning FIMs for all nodes in the unlearning graph.} The FIUn method estimates the unlearning FIM for all nodes within the unlearning graph, including both discovery and inherited nodes (see line 9, Algorithm~\ref{Algo.2}). The unlearning FIMs from discovery nodes are merged to form a collective unlearning FIM for each node in the unlearning graph. This merger is crucial as the linkages between the unlearning data and parameters are embedded within the model and can persist through model inheritance~\cite{foster2024fast}, especially in multi-root scenarios of DAG, where an inherited node may trace back to several discovery nodes. 
Sequential FIM calculation of the unlearned data from multiple roots is required for existing FIM methods~\cite{foster2024fast,golatkar2020eternal}, but this significantly reduces efficiency in IMNs. To address this, \textit{a novel Merging-FIM (MFIM) function} is introduced specifically for this context, denoted by $\Phi\left(\cdot\right)$, and is defined as 
\begin{equation}
\label{eq.merge}
\hat F^{\scriptscriptstyle M}_{j}=\Phi\left( \{\hat F^{\scriptscriptstyle D}_{j,k}\}\right),    
\end{equation}
where $\hat F^{\scriptscriptstyle M}_{j}$ is the merged unlearning FIM for the $j$-th node in the unlearning graph, the set $\{ \hat F^{\scriptscriptstyle D}_{j,k}\}$ collects the unlearning FIMs of discovery nodes within $\mathcal{\hat G}$ that can be traced back from node $n_j$, and $k$ denotes the $k$-th discovery node reachable from $n_j$. 
In this paper, the element-wise maximum is employed for the MFIM function. Specifically, $\hat F^{\scriptscriptstyle M}_{j}$ assembles the largest elements among the unlearning FIMs of discovery nodes in multi-root scenarios and takes the only unlearning FIM in single-root scenarios.

\begin{figure}[t]
\centering
\includegraphics[width=8cm]{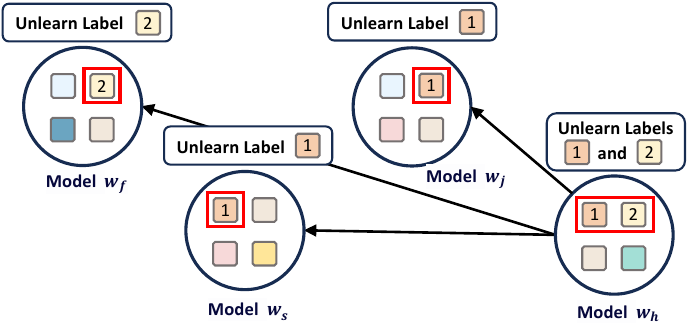}
	\caption{\small Three discovery nodes need to undergo the unlearning process, namely model $w_j$, model $w_s$ and model $w_f$. The labels to be unlearned are label 1 and label 2. However, model $w_j$ and model $w_s$ only have label 1, while model $w_f$ only has label 2. Model $w_h$ inherits from these three discovery nodes, so we merge the unlearning FIMs of the discovery nodes and perform the unlearning of label 1 and label 2 on model $w_h$ together.}
 \label{fig:case}

\end{figure}

As shown in Fig.~\ref{fig:case}, the model $w_h$ inherits from models $w_j$, $w_s$, and $w_f$ and needs to remove requested unlearning data, though each discovery node contains only a portion of the unlearning data.
The FIM of each model reflects the sensitivity of its parameters to the unlearning data, which can be interpreted as parameter importance. By taking the maximum of the corresponding elements across these unlearning FIMs, the merged FIM encompasses all unlearning tasks from the discovery nodes and identifies the parameters that need updating to achieve complete unlearning in a single process.

  \smallskip
\noindent\textbf{Step-4: Update models in the unlearning graph.}
In the final step, the FIUn method identifies which parameters need to be updated by comparing the merged FIM with the model FIM, as described in line 10 of Algorithm~\ref{Algo.2}. The comparison, based on the Selective Synaptic Dampening (SSD)~\cite{foster2024fast}, is also utilized to scale the parameters. For node $n_j$ in the unlearning graph, the parameters are updated as 
\begin{equation}\label{eq.Identify}
\hat w^{\scriptscriptstyle L}_{j,l} =\begin{cases}
\min(\tau\frac{F_{j,l}}{\hat F^{\scriptscriptstyle M}_{j,l}} , \eta) w^{\scriptscriptstyle L}_{j,l}, &\text{if }\frac{\hat F^{\scriptscriptstyle M}_{j,l}}{F_{j,l}}>\gamma;\\ 
w^{\scriptscriptstyle L}_{j,l}, &\text{if} \ \frac{\hat F^{\scriptscriptstyle M}_{j,l}}{F_{j,l}}\le\gamma,\end{cases}
\end{equation}
where $w^{\scriptscriptstyle L}_{j,l}$ and $\hat w^{\scriptscriptstyle L}_{j,l}$ are the $l$-th parameter in the last layer before and after the unlearning update, respectively. Similarly, $F_{j,l}$ and $\hat F^{\scriptscriptstyle M}_{j,l}$ are the $l$-th element of the model FIM and merged unlearning FIM of node $n_j$, respectively. The hyperparameters, including $\tau$, $\eta$, and $\gamma$, balance the unlearning impact on the unlearning data and the accuracy of the retained data.   
 
During the model updating phase, only the parameters of the last layer are updated while the other layers remain frozen, consistent with our design of the unlearning FIM and model FIM, where only the FIM of the last layer is calculated. Elements within the FIM indicate the significance of each parameter relative to the dataset. The threshold $\gamma$ identifies the critical parameters that are significant for the unlearning task, and these parameters are scaled down accordingly.
The hyperparameter $\eta$ ensures minimal change to achieve the desired unlearning effect. Non-critical parameters are left unchanged to maintain accuracy on data that does not require unlearning.

Specific datasets can also be retained by extending the formulations in~(\ref{eq.merge}) and~(\ref{eq.Identify}). In this setting, the unlearning FIM is replaced with a retention FIM, and the role of the parameter $\gamma$ shifts from suppressing to amplifying the weights of critical parameters. Since larger FIM values indicate that the corresponding parameters carry more discriminative information about the training data, we adjust these key parameters proportionally, thereby strengthening the model’s memory and dependence on the retained dataset.

\subsection{Discussion}

From the perspective of the FIM, the impact of data on specific model parameters is consistently localized at precise positions within the parameter space~\cite{foster2024fast}. The stable localization of the FIM across iterations helps precisely locate the parameter locations for unlearning at the class, client, and sample levels using data tied to specific label sets. By fine-tuning $\tau$, $\eta$, and $\gamma$ in~\eqref{eq.Identify} to match the sample size of the class being unlearned, the task transits flexibly between the class-level~\cite{foster2024fast} and sample-level unlearning~\cite{golatkar2020eternal}. Specifically, when the unlearning FIM mirrors the model's FIM with a class-level setting, it becomes a client-level unlearning task.
All types of unlearning are performed irrespective of the succession of inheritance relationships.
Consequently, employing FIUn disrupts the sequential dependencies traditionally required for unlearning, thus enabling fully parallel unlearning processes across any number of subgraphs.

Models across subgraphs and at different depths within the same subgraph can undergo parallel unlearning, enabled by \textit{Hyper-Distance}, where learned knowledge's parameter impact is independent of model inheritance depth. Each model integrates predecessors' label sets to preserve inheritance; focusing on label differences achieves full parallel unlearning, making depths/paths irrelevant without compromising inheritance integrity.
This is possible because the FIUn method utilizes FIMs for unlearning, enabling a class-wise inheritance approach. 

\smallskip 
\noindent\underline{\textbf{Answer to RQ2:}}
The FIUn method enables the independent removal of knowledge from each inherited model. This method utilizes the FIM to obtain the \textit{Hyper-Distance} property which breaks sequential dependencies among models, enabling fully parallel unlearning on inherited models and significantly decreasing the execution time of unlearning tasks. To address the complexity arising from multiple upstream models, the MFIM function is developed to aggregate unlearning FIMs from those models, facilitating the efficient, one-shot removal of inherited knowledge.

\smallskip
We also map the FIUn method and the MFIM function to the DAG-assisted unlearning scenarios described in Sec.~\ref{subsec:umig-unlearning}. This shows how the proposed method efficiently facilitates unlearning through the property of \textit{Hyper-Distance}.

\smallskip
\noindent\textbf{Converting Imbalance into Balance.}
This scenario is particularly significant since existing unlearning methods face unavoidable sequential dependencies. FIUn utilizes the \textit{Hyper-Distance} property to bypass the effects of depths and paths in the inheritance topology. As a result, imbalanced-path scenarios are handled as efficiently as balanced ones, requiring just one update operation per inherited node, regardless of whether unlearning is synchronous or asynchronous.

\smallskip
\noindent\textbf{Converting Multi-root into Single-root.}
The MFIM function is proposed to refine multi-root scenarios. This function merges the FIMs of the corresponding unlearned label sets from multiple discovery nodes into a single FIM, as shown in Fig.~\ref{fig:case}. Specifically, overlapping inherited nodes require only one update operation to eliminate the impact of multiple discovery nodes by conducting a parameter-wise merger, where only the most impactful parameters take effect in this FIM. This function eliminates the need for multiple comparisons among various unlearning FIMs in multi-root scenarios, hence enhancing unlearning efficiency.

\subsection{Implementation}
\noindent\textbf{System Settings.}
FIUn implementation varies based on system architecture:
\begin{packeditemize}
    \item\noindent\textit{Centralized:} Model FIMs are computed and sent with the published model to the central server. Upon an unlearning request, the central server merges the unlearning FIMs from the discovery nodes and executes parallel unlearning.
    \item\noindent\textit{Decentralized:} 
    Each unlearning FIM is publicly shared by discovery nodes. Inherited node owners can then combine these FIMs with their local models as needed, enabling efficient and effective decentralized unlearning.
\end{packeditemize}

Although FIUn focuses on efficient parallel unlearning, it can be extended with auditability mechanisms
to ensure verifiable unlearning in both centralized and decentralized settings.
A practical approach is to integrate FIM commitments that record
the Fisher information signatures before and after unlearning.
Each node can compute a lightweight hash or zero-knowledge proof
of its local FIM difference $\Delta F = F - \hat{F}^D$, 
which is then submitted to a verification server or blockchain ledger.
In a \textit{centralized} mode, the server validates updates by comparing FIM commitments to the expected parameter deltas from~(6). In a \textit{decentralized} mode, commitments are written on-chain, enabling independent validators to check that unlearning matches the declared forgetting requests without revealing model parameters. This yields an auditable trail for the ``right to be forgotten'' while preserving FIUn’s efficiency and scalability.

\smallskip
\noindent\textbf{Complexity.}
The computational complexity of an unlearning task comprises the complexities of Steps-1, 3, and 4 (i.e., calculate unlearning FIM, merge unlearning FIMs and update models, cf. Section~\ref{subsec:fiun}). The complexity of Step-2 is not considered, as the computation of the model FIM can be performed beforehand.

The overall complexity for the unlearning task is \(\mathcal{O}(|\hat D| |L_{\mathrm{all}}| + |\hat D| |L| + |\{n^{\scriptscriptstyle D}_i\}| + |L|)\). Breaking down the complexities, Equations~\eqref{eq.FIM_unlearning} and~\eqref{eq.FIM_model} update $|L|$ parameters per data entry, with complexities \(\mathcal{O}(|\hat D^{\scriptscriptstyle D}_i| |L|)\) and \(\mathcal{O}(|D_j||L|)\), respectively, where $|\hat D^{\scriptscriptstyle D}_i|\le|\hat D|$ and $|L|$ is the number of parameters in the last layer. Additionally, Equations~\eqref{eq.FIM_unlearning} and~\eqref{eq.FIM_model} also requires a forward propagation process through the entire model for each data entry to compute the necessary outputs, with complexity \(\mathcal{O}(|\hat D| |L_{\mathrm{all}}|)\), where $|L_{\mathrm{all}}|$ is the number of parameters in the entire model. 

Equation~\eqref{eq.merge} includes a comparison and merger of $K$ unlearning FIMs, leading to a complexity of \(\mathcal{O}(K)\), where $K$ is the number of traceable discovery nodes in the unlearning graph $\mathcal {\hat G}$ from node $n_j$ and $K\le|\{n^{\scriptscriptstyle D}_i\}|$. Equation~\eqref{eq.Identify} updates $|L|$ model parameters, leading to a complexity of \(\mathcal{O}(|L|)\). As the computation of the unlearning FIM for discovery nodes (including the forward propagation) and the model updates can proceed in parallel, the overall complexity of the unlearning tasks is \(\mathcal{O}(|\hat D| |L_{\mathrm{all}}| + |\hat D| |L| + |\{n^{\scriptscriptstyle D}_i\}| + |L|)\).

\section{Experimental Settings}\label{sec:experimental settings}

This section first presents the datasets used in the experiments, the models selected, and the performance metrics for accuracy evaluation, followed by the topology setups and the benchmarks for unlearning to evaluate the FIUn method.

 \begin{figure}[!ht]
\centering
\includegraphics[width=7cm]{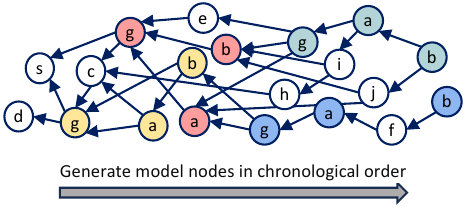}
	\caption{ Models topologies considered in the experiments.}
 \label{fig:total_Experi_Scenarios_2}
 \vspace{-1em}
\end{figure}

\begin{table}[t]
\centering
\caption{Hyperparameters.}
\label{table: hyperparameter}
\small
\begin{tabular}{p{0.95\linewidth}}
\toprule
\textbf{Defaults (all models unless overridden):} optimizer = \textbf{SGD}, learning rate = \textbf{0.01}, epochs = \textbf{100}, batch size = \textbf{64}. \\
\midrule
\textbf{Overrides:}\\
FIUn: lr = 0.1; $\tau$ = 1; $\gamma$ = 1; $\eta$ = 0.1; use layer = last. \\
AlexNet: epochs = 200. \\
BERT: optimizer = Adam; lr = 2e-5; epochs = 30. \\
GCNN: optimizer = Adam; lr = 0.001; epochs = 30. \\
ResNet18: lr = 0.1. \\
\bottomrule
\end{tabular}
\end{table}

\begin{table*}
\centering
\caption{Baseline of Various Framework Performance on CIFAR100 Unlearning Speed Comparison}
\label{table:fed_baseline}
\resizebox{0.78\textwidth}{!}{
\begin{tabular}{ccc|ccc|ccc|ccc|ccc|ccc} 
\toprule
\multirow{3}{*}{\textbf{Model}}              & \multirow{3}{*}{\begin{tabular}[c]{@{}c@{}}\textbf{Framework}\end{tabular}} & \multirow{3}{*}{\#$C_f$} & \multicolumn{15}{c}{\textbf{Cumulative Unlearning Time (s)}}                                                                                                                                                                                                                                                                                                                                                            \\ 
\cmidrule{4-18}
                                    &                                                                      &                          & \multicolumn{3}{c|}{Gradient Ascent}                                                 & \multicolumn{3}{c|}{Fine-tuning}                                                        & \multicolumn{3}{c|}{Distill}                                                         & \multicolumn{3}{c|}{Re-training}                                & \multicolumn{3}{c}{FIUn}                                                \\ 
\cmidrule{4-18}
                                    &                                                                      &                          & \multicolumn{1}{c}{$w_g$} & \multicolumn{1}{c}{$w_a$} & \multicolumn{1}{c|}{$w_b$} & \multicolumn{1}{c}{$w_g$} & \multicolumn{1}{c}{$w_a$} & \multicolumn{1}{c|}{$w_b$} & \multicolumn{1}{c}{$w_g$} & \multicolumn{1}{c}{$w_a$} & \multicolumn{1}{c|}{$w_b$} & \multicolumn{1}{c}{$w_g$} & \multicolumn{1}{c}{$w_a$} & $w_b$ & \multicolumn{1}{c}{$w_g$} & \multicolumn{1}{c}{$w_a$} & $w_b$          \\ 
\midrule
\multirow{8}{*}{\rotcell{AlexNet}}  & \multirow{2}{*}{FL}                                                  & 1                        & 0.97                       & 1.05                       & 1.26                       & 0.54                       & 0.75                       & 1.08                       & 82.5                       & 153.65                     & 294.32                     & 29.70                      & 59.06                      & 89.62 & \cellcolor{yellow!14}\textbf{0.09}              & \cellcolor{yellow!14}\textbf{0.11}              & \cellcolor{yellow!14}\textbf{0.13}  \\
                                    &                                                                      & \cellcolor{blue!8}10                       & \cellcolor{blue!8}1.1                        & \cellcolor{blue!8}1.27                       & \cellcolor{blue!8}1.68                       & \cellcolor{blue!8}0.53                       & \cellcolor{blue!8}1.95                       & \cellcolor{blue!8}2.31                       & \cellcolor{blue!8}75.3                       & \cellcolor{blue!8}147.6                      & \cellcolor{blue!8}326.8                      & \cellcolor{blue!8}21.96                      & \cellcolor{blue!8}42.37                      & \cellcolor{blue!8}63.82 & \cellcolor{yellow!14}\textbf{0.11}              & \cellcolor{yellow!14}\textbf{0.14}              & \cellcolor{yellow!14}\textbf{0.15}  \\ 
\cmidrule{2-18}
                                    & \multirow{2}{*}{IL}                                                  & 1                        & 1.26                       & 1.32                       & 1.39                       & 1.25                       & 2.09                       & 3.81                       & 108.07                     & 212.36                     & 424.96                     & 22.95                      & 43.83                      & 63.45 & \cellcolor{yellow!14}\textbf{0.09}              & \cellcolor{yellow!14}\textbf{0.39}              & \cellcolor{yellow!14}\textbf{0.39}  \\
                                    &                                                                      & \cellcolor{blue!8}10                       & \cellcolor{blue!8}1.29                       & \cellcolor{blue!8}1.39                       & \cellcolor{blue!8}1.54                       & \cellcolor{blue!8}1.26                       & \cellcolor{blue!8}2.14                       & \cellcolor{blue!8}3.24                       & \cellcolor{blue!8}109.3                      & \cellcolor{blue!8}214.35                     & \cellcolor{blue!8}421.85                     & \cellcolor{blue!8}17.35                      & \cellcolor{blue!8}38.32                      & \cellcolor{blue!8}57.94 & \cellcolor{yellow!14}\textbf{0.12}              & \cellcolor{yellow!14}\textbf{0.28}              & \cellcolor{yellow!14}\textbf{0.29}  \\ 
\cmidrule{2-18}
                                    & \multirow{2}{*}{DDPL}                                                  & 1                        & 1.42                       & 1.47                       & 1.56                       & 1.53                       & 1.83                       & 1.79                       & 326.64                     & 108.13                     & 121.98                     & 30.13                      & 31.26                      & 31.14 & \cellcolor{yellow!14}\textbf{0.23}              & \cellcolor{yellow!14}\textbf{0.14}              & \cellcolor{yellow!14}\textbf{0.09}  \\
                                    &                                                                      & \cellcolor{blue!8}10                       & \cellcolor{blue!8}1.56                       & \cellcolor{blue!8}1.63                       & \cellcolor{blue!8}1.75                       & \cellcolor{blue!8}1.74                       & \cellcolor{blue!8}2.03                       & \cellcolor{blue!8}2.01                       & \cellcolor{blue!8}114.32                     & \cellcolor{blue!8}111.3                      & \cellcolor{blue!8}111.19                     & \cellcolor{blue!8}\cellcolor{blue!8}21.03                      & \cellcolor{blue!8}22.96                      & \cellcolor{blue!8}22.10 & \cellcolor{yellow!14}\textbf{0.24}              & \cellcolor{yellow!14}\textbf{0.14}              & \cellcolor{yellow!14}\textbf{0.10}  \\ 
\cmidrule{2-18}
                                    & \multirow{2}{*}{TL}                                                  & 1                        & 1.14                       & 1.21                       & 1.3                        & 1.16                       & 2.19                       & 1.98                       & 112.87                     & 220.13                     & 226.31                     & 20.31                      & 40.49                      & 42.36 & \cellcolor{yellow!14}\textbf{0.09}              & \cellcolor{yellow!14}\textbf{0.11}              & \cellcolor{yellow!14}\textbf{0.11}  \\
                                    &                                                                      & \cellcolor{blue!8}10                       & \cellcolor{blue!8}1.26                       & \cellcolor{blue!8}1.34                       & \cellcolor{blue!8}1.36                       & \cellcolor{blue!8}1.25                       & \cellcolor{blue!8}2.06                       & \cellcolor{blue!8}2.01                       & \cellcolor{blue!8}109.44                     & \cellcolor{blue!8}217.32                     & \cellcolor{blue!8}223.91                     & \cellcolor{blue!8}19.42                      & \cellcolor{blue!8}39.46                      & \cellcolor{blue!8}39.35 & \cellcolor{yellow!14}\textbf{0.08}              & \cellcolor{yellow!14}\textbf{0.13}              & \cellcolor{yellow!14}\textbf{0.11}  \\ 
\midrule
\multirow{8}{*}{\rotcell{ResNet18}} & \multirow{2}{*}{FL}                                                  & 1                        & 2.45                       & 2.89                       & 3.42                       & 0.74                       & 1.32                       & 1.74                       & 70.31                      & 129.44                     & 247.6                      & 30.36                      & 61.40                      & 90.94 & \cellcolor{yellow!14}\textbf{0.68}              & \cellcolor{yellow!14}\textbf{1.00}              & \cellcolor{yellow!14}\textbf{0.99}  \\
                                    &                                                                      & \cellcolor{blue!8}10                       & \cellcolor{blue!8}2.02                       & \cellcolor{blue!8}2.58                       & \cellcolor{blue!8}3.67                       & \cellcolor{blue!8}0.85                       & \cellcolor{blue!8}1.54                       & \cellcolor{blue!8}1.84                       & \cellcolor{blue!8}75.12                      & \cellcolor{blue!8}150.35                     & \cellcolor{blue!8}267.34                     & \cellcolor{blue!8}22.03                      & \cellcolor{blue!8}44.83                      & \cellcolor{blue!8}66.30 & \cellcolor{yellow!14}\textbf{0.76}              & \cellcolor{yellow!14}\textbf{1.14}              & \cellcolor{yellow!14}\textbf{1.04}  \\ 
\cmidrule{2-18}
                                    & \multirow{2}{*}{IL}                                                  & 1                        & 1.52                       & 1.63                       & 1.84                       & 1.35                       & 2.65                       & 3.86                       & 98.51                      & 170.66                     & 317.53                     & 26.35                      & 47.12                      & 66.93 & \cellcolor{yellow!14}\textbf{0.30}              & \cellcolor{yellow!14}\textbf{0.80}              & \cellcolor{yellow!14}\textbf{0.85}  \\
                                    &                                                                      & \cellcolor{blue!8}10                       & \cellcolor{blue!8}1.33                       & \cellcolor{blue!8}1.45                       & \cellcolor{blue!8}1.62                       & \cellcolor{blue!8}1.47                       & \cellcolor{blue!8}2.24                       & \cellcolor{blue!8}3.74                       & \cellcolor{blue!8}81.74                      & \cellcolor{blue!8}143.53                     & \cellcolor{blue!8}275.34                     & \cellcolor{blue!8}18.93                      & \cellcolor{blue!8}39.93                      & \cellcolor{blue!8}58.93 & \cellcolor{yellow!14}\textbf{0.32}              & \cellcolor{yellow!14}\textbf{0.88}              & \cellcolor{yellow!14}\textbf{0.89}  \\ 
\cmidrule{2-18}
                                    & \multirow{2}{*}{DDPL}                                                  & 1                        & 1.18                       & 1.25                       & 1.34                       & 1.57                       & 1.43                       & 1.51                       & 72.34                      & 71.87                      & 71.39                      & 32.53                      & 31.95                      & 32.18 & \cellcolor{yellow!14}\textbf{0.64}              & \cellcolor{yellow!14}\textbf{0.34}              & \cellcolor{yellow!14}\textbf{0.30}  \\
                                    &                                                                      & \cellcolor{blue!8}10                       & \cellcolor{blue!8}1.34                       & \cellcolor{blue!8}1.42                       & \cellcolor{blue!8}1.51                       & \cellcolor{blue!8}2.04                       & \cellcolor{blue!8}1.85                       & \cellcolor{blue!8}1.97                       & \cellcolor{blue!8}71.41                      & \cellcolor{blue!8}66.56                      & \cellcolor{blue!8}70.41                      & \cellcolor{blue!8}22.60                      & \cellcolor{blue!8}21.92                      & \cellcolor{blue!8}24.19 & \cellcolor{yellow!14}\textbf{0.68}              & \cellcolor{yellow!14}\textbf{0.34}              & \cellcolor{yellow!14}\textbf{0.34}  \\ 
\cmidrule{2-18}
                                    & \multirow{2}{*}{TL}                                                  & 1                        & 1.39                       & 1.48                       & 1.52                       & 1.37                       & 3.67                       & 3.25                       & 95.4                       & 163.54                     & 168.51                     & 22.21                      & 41.52                      & 42.93 & \cellcolor{yellow!14}\textbf{0.15}              & \cellcolor{yellow!14}\textbf{0.24}              & \cellcolor{yellow!14}\textbf{0.24}  \\
                                    &                                                                      & \cellcolor{blue!8}10                       & \cellcolor{blue!8}1.35                       & \cellcolor{blue!8}1.51                       & \cellcolor{blue!8}1.46                       & \cellcolor{blue!8}1.5                        & \cellcolor{blue!8}5.53                       & \cellcolor{blue!8}5.54                       & \cellcolor{blue!8}81.15                      & \cellcolor{blue!8}145.73                     & \cellcolor{blue!8}151.52                     & \cellcolor{blue!8}19.10                      & \cellcolor{blue!8}42.60                      & \cellcolor{blue!8}41.36 & \cellcolor{yellow!14}\textbf{0.16}              & \cellcolor{yellow!14}\textbf{0.27}              & \cellcolor{yellow!14}\textbf{0.27}  \\
\bottomrule
\end{tabular}
}
\end{table*}

\begin{table*}
\centering
\setlength{\extrarowheight}{0pt}
\addtolength{\extrarowheight}{\aboverulesep}
\addtolength{\extrarowheight}{\belowrulesep}
\setlength{\aboverulesep}{0pt}
\setlength{\belowrulesep}{0pt}
\caption{Transfer Unlearning Performance on CIFAR100}
\label{table:transfer_unlearn1}
\resizebox{0.78\textwidth}{!}{
\begin{tabular}{ccc|ccc|ccc|ccc|cccccc} 
\toprule
\multirow{3}{*}{\textbf{Model}}     & \multirow{3}{*}{\#$C_f$} & \multirow{3}{*}{\textbf{Metrics}}                  & \multicolumn{3}{c|}{\multirow{2}{*}{Original~(\%)}}                                                                   & \multicolumn{3}{c|}{\multirow{2}{*}{Re-training~(\%)}}                                                             & \multicolumn{3}{c|}{\multirow{2}{*}{FIUn~(\%)}}                                                                                                  & \multicolumn{6}{c}{\textbf{Cumulative Unlearning Time (s)}}                                                                                                                                                                                                                         \\ 
\cmidrule{13-18}
                                    &                          &                                                    & \multicolumn{3}{c|}{}                                                                                                 & \multicolumn{3}{c|}{}                                                                                              & \multicolumn{3}{c|}{}                                                                                                                            & \multicolumn{3}{c|}{Re-training}                                             & \multicolumn{3}{c}{FIUn}                                                                                                                                                                             \\ 
\cmidrule{4-18}
                                    &                          &                                                    & $w_g$                                 & $w_a$                                 & $w_b$                                 & $w_g$                                & $w_a$                                & $w_b$                                & $w_g$                                          & $w_a$                                          & $w_b$                                          & $w_g$                  & $w_a$                  & \multicolumn{1}{c|}{$w_b$} & $w_g$                                                           & $w_a$                                                           & $w_b$                                                            \\ 
\midrule
\multirow{6}{*}{\rotcell{AlexNet}}  & \multirow{2}{*}{1}       & $AD_r\uparrow$                                     & 97.16                                 & 95.61                                 & 92.20                                 & 98.17                                & 95.98                                & 91.32                                & {\cellcolor[rgb]{1,0.992,0.859}}\textbf{91.40} & {\cellcolor[rgb]{1,0.992,0.859}}\textbf{86.24} & {\cellcolor[rgb]{1,0.992,0.859}}\textbf{83.99} & \multirow{2}{*}{20.31} & \multirow{2}{*}{40.49} & \multirow{2}{*}{42.36}     & {\cellcolor[rgb]{1,0.992,0.859}}                                & {\cellcolor[rgb]{1,0.992,0.859}}                                & {\cellcolor[rgb]{1,0.992,0.859}}                                 \\
                                    &                          & {\cellcolor[rgb]{0.922,0.922,1}}${AD}_f\downarrow$ & {\cellcolor[rgb]{0.922,0.922,1}}96.66 & {\cellcolor[rgb]{0.922,0.922,1}}99.99 & {\cellcolor[rgb]{0.922,0.922,1}}99.99 & {\cellcolor[rgb]{0.922,0.922,1}}0.00 & {\cellcolor[rgb]{0.922,0.922,1}}0.00 & {\cellcolor[rgb]{0.922,0.922,1}}0.00 & {\cellcolor[rgb]{1,0.992,0.859}}\textbf{0.00}  & {\cellcolor[rgb]{1,0.992,0.859}}\textbf{0.00}  & {\cellcolor[rgb]{1,0.992,0.859}}\textbf{0.00}  &                        &                        &                            & \multirow{-2}{*}{{\cellcolor[rgb]{1,0.992,0.859}}\textbf{0.09}} & \multirow{-2}{*}{{\cellcolor[rgb]{1,0.992,0.859}}\textbf{0.11}} & \multirow{-2}{*}{{\cellcolor[rgb]{1,0.992,0.859}}\textbf{0.11}}  \\ 
\hhline{~-----------------}
                                    & \multirow{2}{*}{2}       & $AD_r\uparrow$~                                    & 97.16                                 & 95.61                                 & 92.20                                 & 96.88                                & 96.68                                & 93.16                                & {\cellcolor[rgb]{1,0.992,0.859}}\textbf{82.36} & {\cellcolor[rgb]{1,0.992,0.859}}\textbf{80.47} & {\cellcolor[rgb]{1,0.992,0.859}}\textbf{79.99} & \multirow{2}{*}{21.20} & \multirow{2}{*}{40.56} & \multirow{2}{*}{41.92}     & {\cellcolor[rgb]{1,0.992,0.859}}                                & {\cellcolor[rgb]{1,0.992,0.859}}                                & {\cellcolor[rgb]{1,0.992,0.859}}                                 \\
                                    &                          & {\cellcolor[rgb]{0.922,0.922,1}}${AD}_f\downarrow$ & {\cellcolor[rgb]{0.922,0.922,1}}99.99 & {\cellcolor[rgb]{0.922,0.922,1}}99.99 & {\cellcolor[rgb]{0.922,0.922,1}}99.99 & {\cellcolor[rgb]{0.922,0.922,1}}0.00 & {\cellcolor[rgb]{0.922,0.922,1}}0.00 & {\cellcolor[rgb]{0.922,0.922,1}}0.00 & {\cellcolor[rgb]{1,0.992,0.859}}\textbf{0.00}  & {\cellcolor[rgb]{1,0.992,0.859}}\textbf{0.00}  & {\cellcolor[rgb]{1,0.992,0.859}}\textbf{0.00}  &                        &                        &                            & \multirow{-2}{*}{{\cellcolor[rgb]{1,0.992,0.859}}\textbf{0.08}} & \multirow{-2}{*}{{\cellcolor[rgb]{1,0.992,0.859}}\textbf{0.13}} & \multirow{-2}{*}{{\cellcolor[rgb]{1,0.992,0.859}}\textbf{0.11}}  \\ 
\hhline{~-----------------}
                                    & \multirow{2}{*}{4}       & $AD_r\uparrow$                                     & 97.16                                 & 95.61                                 & 92.20                                 & 94.67                                & 96.45                                & 93.45                                & {\cellcolor[rgb]{1,0.992,0.859}}\textbf{71.59} & {\cellcolor[rgb]{1,0.992,0.859}}\textbf{71.43} & {\cellcolor[rgb]{1,0.992,0.859}}\textbf{76.06} & \multirow{2}{*}{20.30} & \multirow{2}{*}{39.76} & \multirow{2}{*}{40.60}     & {\cellcolor[rgb]{1,0.992,0.859}}                                & {\cellcolor[rgb]{1,0.992,0.859}}                                & {\cellcolor[rgb]{1,0.992,0.859}}                                 \\
                                    &                          & {\cellcolor[rgb]{0.922,0.922,1}}${AD}_f\downarrow$ & {\cellcolor[rgb]{0.922,0.922,1}}96.61 & {\cellcolor[rgb]{0.922,0.922,1}}98.30 & {\cellcolor[rgb]{0.922,0.922,1}}98.30 & {\cellcolor[rgb]{0.922,0.922,1}}0.00 & {\cellcolor[rgb]{0.922,0.922,1}}0.00 & {\cellcolor[rgb]{0.922,0.922,1}}0.00 & {\cellcolor[rgb]{1,0.992,0.859}}\textbf{0.00}  & {\cellcolor[rgb]{1,0.992,0.859}}\textbf{0.00}  & {\cellcolor[rgb]{1,0.992,0.859}}\textbf{0.00}  &                        &                        &                            & \multirow{-2}{*}{{\cellcolor[rgb]{1,0.992,0.859}}\textbf{0.09}} & \multirow{-2}{*}{{\cellcolor[rgb]{1,0.992,0.859}}\textbf{0.11}} & \multirow{-2}{*}{{\cellcolor[rgb]{1,0.992,0.859}}\textbf{0.11}}  \\ 
\midrule
\multirow{6}{*}{\rotcell{ResNet18}} & \multirow{2}{*}{1}       & $AD_r\uparrow$                                     & 99.99                                 & 99.99                                 & 99.99                                 & 99.99                                & 99.99                                & 99.99                                & {\cellcolor[rgb]{1,0.992,0.859}}\textbf{89.09} & {\cellcolor[rgb]{1,0.992,0.859}}\textbf{93.51} & {\cellcolor[rgb]{1,0.992,0.859}}\textbf{95.22} & \multirow{2}{*}{22.21} & \multirow{2}{*}{41.52} & \multirow{2}{*}{42.93}     & {\cellcolor[rgb]{1,0.992,0.859}}                                & {\cellcolor[rgb]{1,0.992,0.859}}                                & {\cellcolor[rgb]{1,0.992,0.859}}                                 \\
                                    &                          & {\cellcolor[rgb]{0.922,0.922,1}}${AD}_f\downarrow$ & {\cellcolor[rgb]{0.922,0.922,1}}99.99 & {\cellcolor[rgb]{0.922,0.922,1}}99.99 & {\cellcolor[rgb]{0.922,0.922,1}}99.99 & {\cellcolor[rgb]{0.922,0.922,1}}0.00 & {\cellcolor[rgb]{0.922,0.922,1}}0.00 & {\cellcolor[rgb]{0.922,0.922,1}}0.00 & {\cellcolor[rgb]{1,0.992,0.859}}\textbf{0.00}  & {\cellcolor[rgb]{1,0.992,0.859}}\textbf{0.00}  & {\cellcolor[rgb]{1,0.992,0.859}}\textbf{0.00}  &                        &                        &                            & \multirow{-2}{*}{{\cellcolor[rgb]{1,0.992,0.859}}\textbf{0.15}} & \multirow{-2}{*}{{\cellcolor[rgb]{1,0.992,0.859}}\textbf{0.24}} & \multirow{-2}{*}{{\cellcolor[rgb]{1,0.992,0.859}}\textbf{0.24}}  \\ 
\hhline{~-----------------}
                                    & \multirow{2}{*}{2}       & $AD_r\uparrow$                                     & 99.99                                 & 99.99                                 & 99.99                                 & 99.99                                & 99.99                                & 99.99                                & {\cellcolor[rgb]{1,0.992,0.859}}\textbf{85.89} & {\cellcolor[rgb]{1,0.992,0.859}}\textbf{94.89} & {\cellcolor[rgb]{1,0.992,0.859}}\textbf{95.79} & \multirow{2}{*}{21.19} & \multirow{2}{*}{41.72} & \multirow{2}{*}{43.20}     & {\cellcolor[rgb]{1,0.992,0.859}}                                & {\cellcolor[rgb]{1,0.992,0.859}}                                & {\cellcolor[rgb]{1,0.992,0.859}}                                 \\
                                    &                          & {\cellcolor[rgb]{0.922,0.922,1}}${AD}_f\downarrow$ & {\cellcolor[rgb]{0.922,0.922,1}}99.99 & {\cellcolor[rgb]{0.922,0.922,1}}99.99 & {\cellcolor[rgb]{0.922,0.922,1}}99.99 & {\cellcolor[rgb]{0.922,0.922,1}}0.00 & {\cellcolor[rgb]{0.922,0.922,1}}0.00 & {\cellcolor[rgb]{0.922,0.922,1}}0.00 & {\cellcolor[rgb]{1,0.992,0.859}}\textbf{0.00}  & {\cellcolor[rgb]{1,0.992,0.859}}\textbf{0.00}  & {\cellcolor[rgb]{1,0.992,0.859}}\textbf{0.00}  &                        &                        &                            & \multirow{-2}{*}{{\cellcolor[rgb]{1,0.992,0.859}}\textbf{0.17}} & \multirow{-2}{*}{{\cellcolor[rgb]{1,0.992,0.859}}\textbf{0.23}} & \multirow{-2}{*}{{\cellcolor[rgb]{1,0.992,0.859}}\textbf{0.26}}  \\ 
\hhline{~-----------------}
                                    & \multirow{2}{*}{4}       & $AD_r\uparrow$                                     & 99.96                                 & 99.99                                 & 99.99                                 & 99.99                                & 99.99                                & 99.99                                & {\cellcolor[rgb]{1,0.992,0.859}}\textbf{80.20} & {\cellcolor[rgb]{1,0.992,0.859}}\textbf{94.82} & {\cellcolor[rgb]{1,0.992,0.859}}\textbf{97.03} & \multirow{2}{*}{22.03} & \multirow{2}{*}{42.26} & \multirow{2}{*}{43.25}     & {\cellcolor[rgb]{1,0.992,0.859}}                                & {\cellcolor[rgb]{1,0.992,0.859}}                                & {\cellcolor[rgb]{1,0.992,0.859}}                                 \\
                                    &                          & {\cellcolor[rgb]{0.922,0.922,1}}${AD}_f\downarrow$ & {\cellcolor[rgb]{0.922,0.922,1}}99.99 & {\cellcolor[rgb]{0.922,0.922,1}}99.99 & {\cellcolor[rgb]{0.922,0.922,1}}99.99 & {\cellcolor[rgb]{0.922,0.922,1}}0.00 & {\cellcolor[rgb]{0.922,0.922,1}}0.00 & {\cellcolor[rgb]{0.922,0.922,1}}0.00 & {\cellcolor[rgb]{1,0.992,0.859}}\textbf{0.00}  & {\cellcolor[rgb]{1,0.992,0.859}}\textbf{0.00}  & {\cellcolor[rgb]{1,0.992,0.859}}\textbf{0.00}  &                        &                        &                            & \multirow{-2}{*}{{\cellcolor[rgb]{1,0.992,0.859}}\textbf{0.17}} & \multirow{-2}{*}{{\cellcolor[rgb]{1,0.992,0.859}}\textbf{0.25}} & \multirow{-2}{*}{{\cellcolor[rgb]{1,0.992,0.859}}\textbf{0.25}}  \\
\bottomrule
\end{tabular}
}
\end{table*}

\subsection{Datasets and Models}
\noindent\textbf{Datasets.} The experiment uses three datasets to evaluate the proposed FIUn method. These three datasets are benchmarks for image classification tasks, encompassing diverse image classification tasks.

\begin{packeditemize}
\item \textbf{CIFAR100.} 
This dataset includes 60,000 color images across 100 classes, each with 600 images, sized at $32\times 32$ pixels. It comprises 50,000 training and 10,000 test images. Each class belongs to one of 20 superclasses, with both a coarse (superclass) and fine (specific class) label per image.
       
\item \textbf{TinyImageNet.}
This dataset includes 200 classes, with 500 training images, 50 validation images, and 50 test images per class, totaling 100,000 images. The images are resized to $64\times 64$ pixels, larger than CIFAR100 but smaller than the original ImageNet.

\item \textbf{Yahoo! Answers.}
his text dataset is a large public Q\&A dataset widely used in natural language processing (NLP) research for tasks like text classification, question answering, and sentiment analysis. It contains categorized question-answer pairs from Yahoo! Answers.

\end{packeditemize}

 \smallskip
\noindent\textbf{Models.} We use the AlexNet, ResNet18, DenseNet161, BERT-base-cased language model and GCNN to evaluate the impact of our proposed FIUn method on the accuracy of the unlearned label set $C_f$ and the retained label set $C_r$.

 \smallskip
\noindent\textbf{Performance Metrics.} We use the training datasets to check the accuracy of the model to evaluate its practicality in experiments. This is a common and reasonable approach since it directly evaluates if the removed information still influences the model, as widely accepted in the field~\cite{tarun2023fast,golatkar2020eternal}.

\begin{packeditemize}
\item \noindent\textbf{Accuracy on unlearned labels ($AD_{f}$).} The accuracy of the unlearned label set in the unlearning model should ideally approach 0\%. This represents a scenario where the softmax layer assigns negligible probabilities to unlearned classes. When applying the Top-$N$ strategy, the likelihood of selecting unlearned classes becomes extremely small, effectively resulting in 0\% accuracy. We evaluate this by testing the unlearning model on training datasets to assess the effectiveness of the unlearning process.

\item \noindent\textbf{Accuracy on retained labels ($AD_{r}$).} 
The accuracy of the retained label set should closely match the original model's accuracy before unlearning.

\item \noindent\textbf{Cumulative unlearning time.} The cumulative time required for each model to unlearn labels during the training process in machine learning and deep learning.

\item \noindent\textbf{Accuracy difference index ($\bigtriangleup_{acc}$).}  The difference between $AD_r$ and $AD_f$ directly measures unlearning effectiveness, with a larger difference indicating more effective unlearning, up to a maximum value of 1.
\end{packeditemize}



\begin{figure*}[!ht]
 \centering
  \subfigure[FL in ResNet18 of \#$C_f=1$]{
         \centering
         \includegraphics[width=0.23\textwidth]{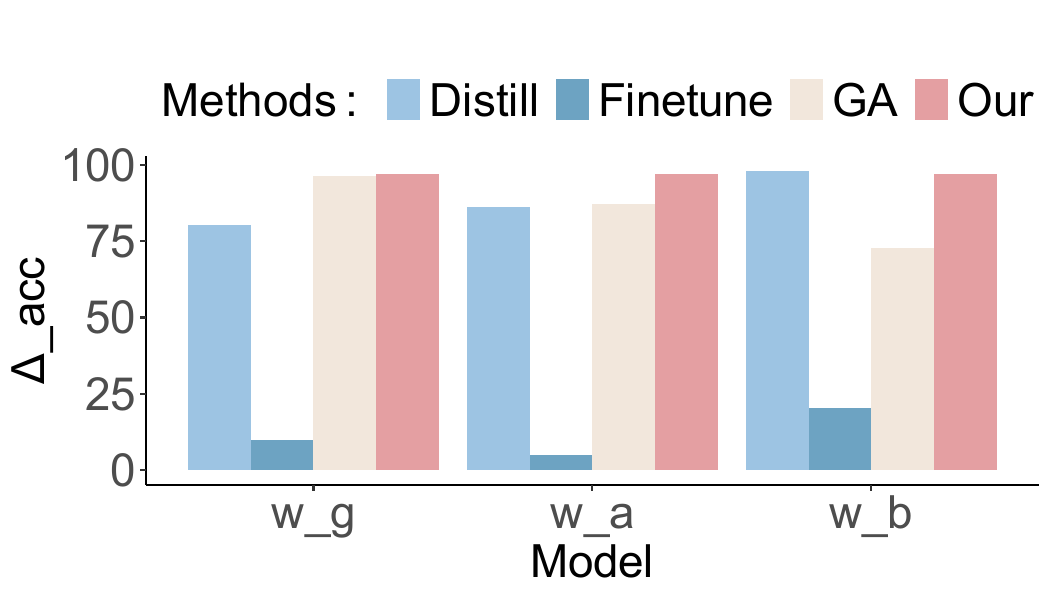}
         
     }
     \subfigure[FL in ResNet18 of \#$C_f=10$]{
         \centering\includegraphics[width=0.23\textwidth]{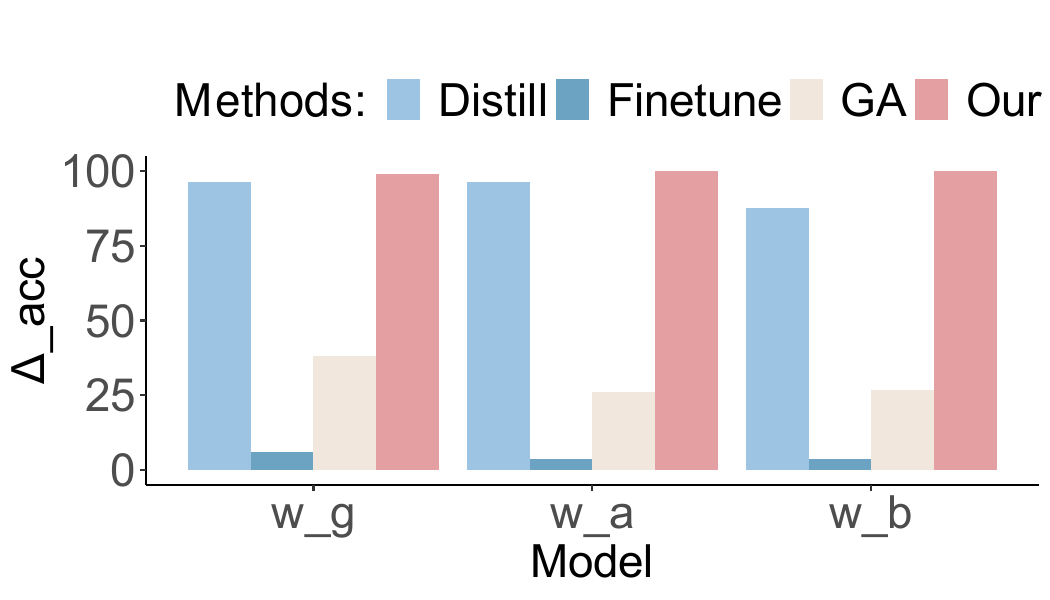}

     }
       \subfigure[FL in AlexNet of \#$C_f=1$]{
         \centering
         \includegraphics[width=0.23\textwidth]{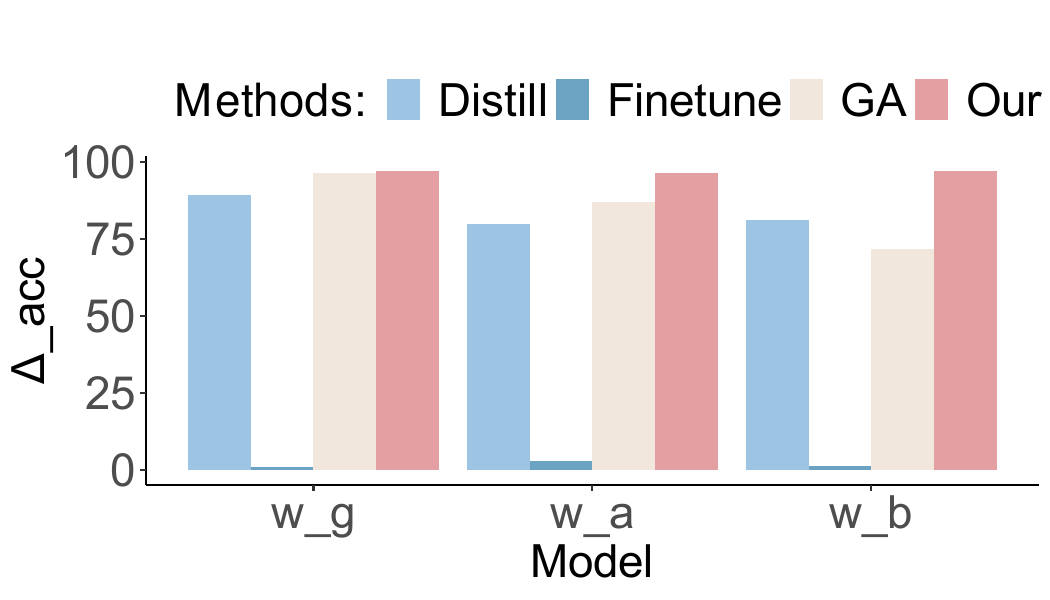}

     }
    \subfigure[FL in AlexNet of \#$C_f=10$]{
        \centering
        \includegraphics[width=0.23\textwidth]{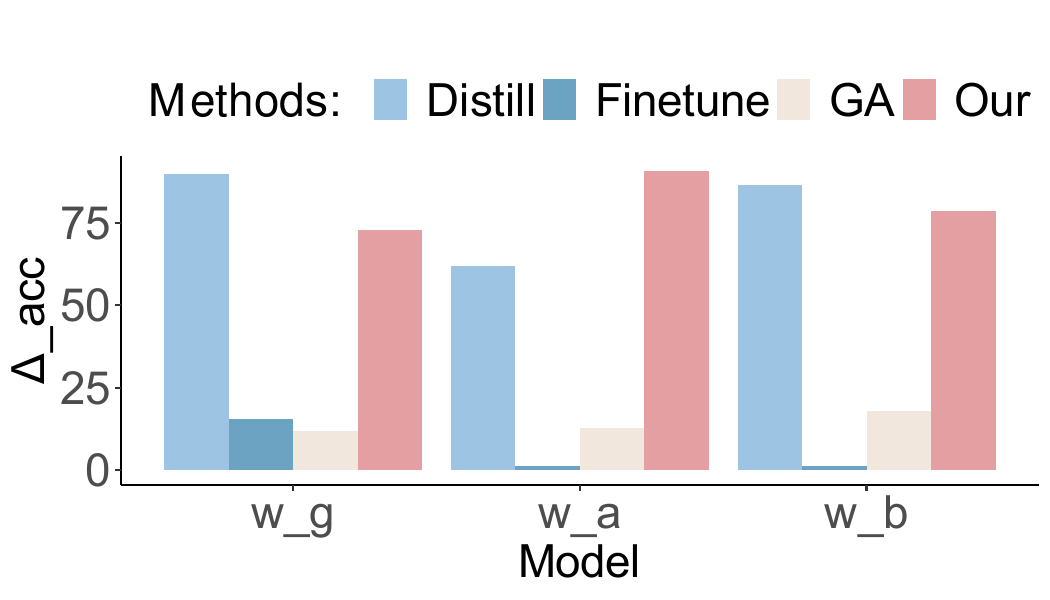}
     
    }

    \subfigure[DDPL in ResNet18 of \#$C_f=1$]{
        \centering
        \includegraphics[width=0.23\textwidth]{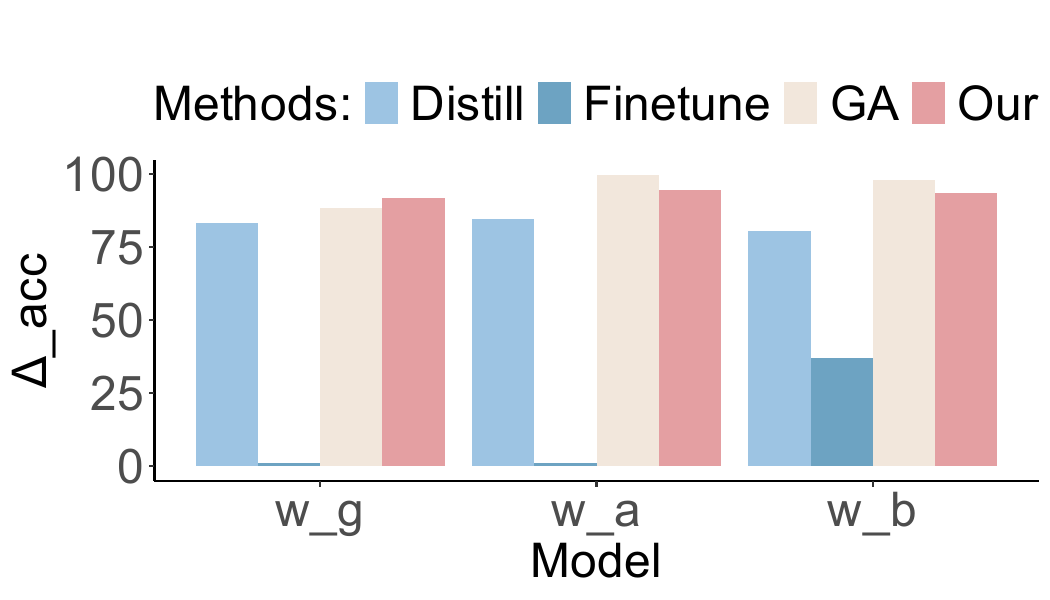}
    }
   \subfigure[DDPL in ResNet18 of \#$C_f=10$]{
        \centering
        \includegraphics[width=0.23\textwidth]{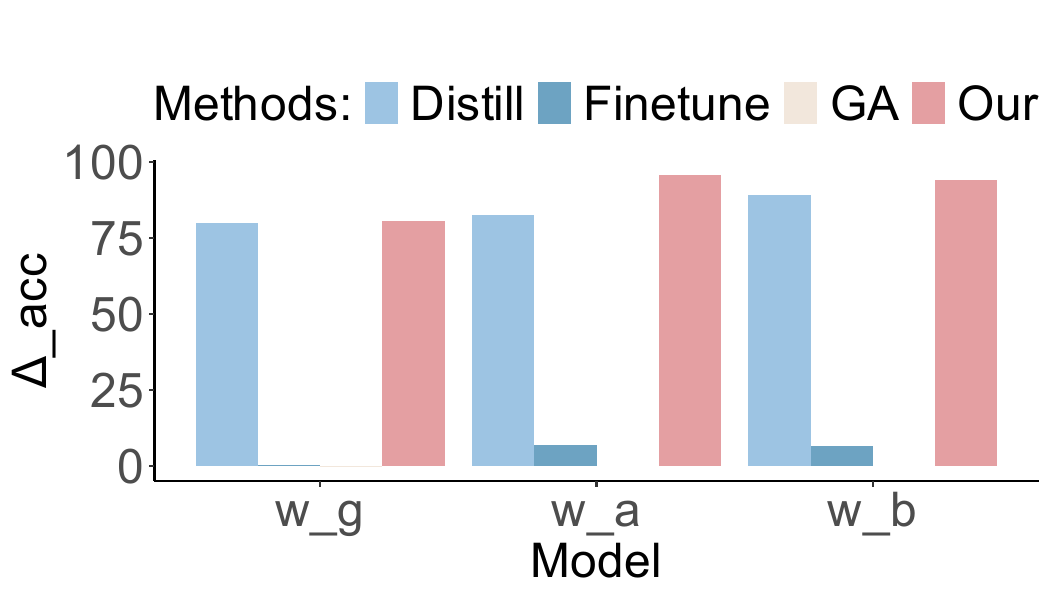}
    }
     \subfigure[DDPL in AlexNet of \#$C_f=1$]{
        \centering
        \includegraphics[width=0.23\textwidth]{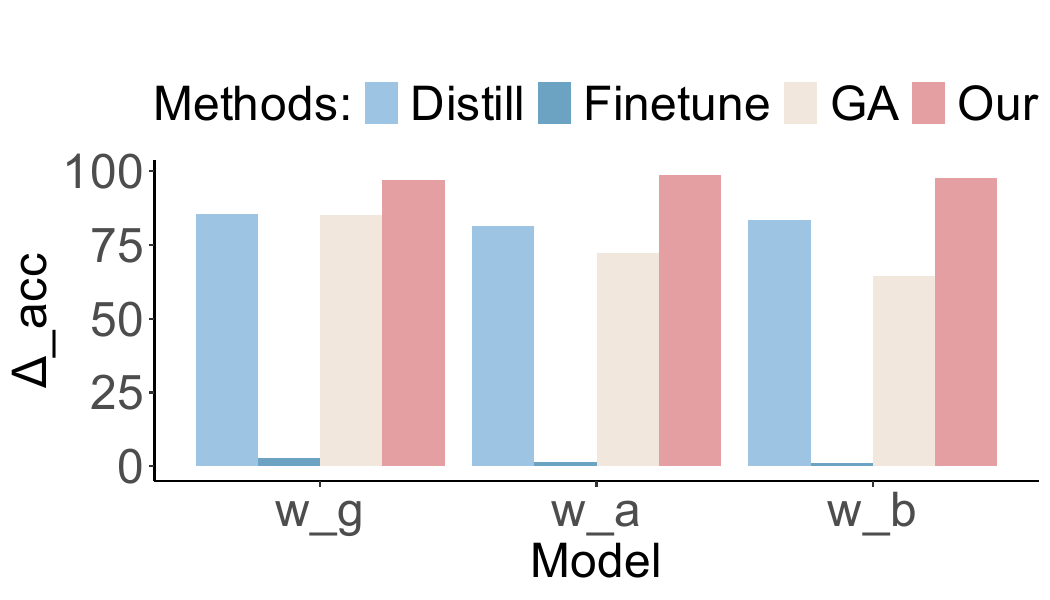}
    }
        \subfigure[DDPL in AlexNet of \#$C_f=10$]{
        \centering
        \includegraphics[width=0.23\textwidth]{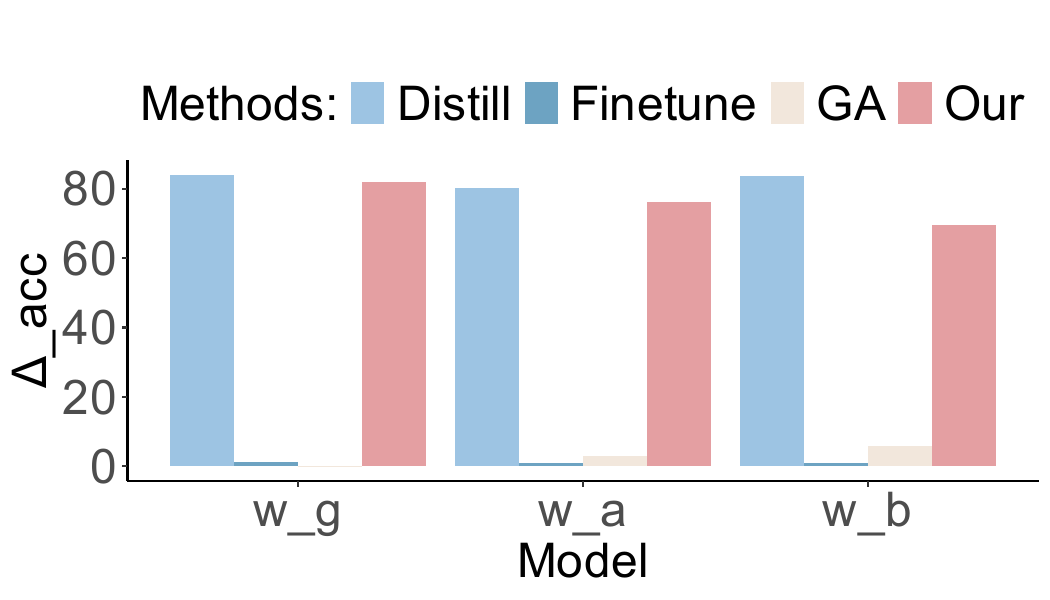}
    }

  \caption{Baseline Unlearning Accuracy of FL and DDPL Frameworks on CIFAR100.}

  \label{fig:baseline1}
  \vspace{-1em}
\end{figure*}

\begin{figure*}[!ht]
 \centering

 \subfigure[IL in ResNet18 of \#$C_f=1$]{
        \centering
        \includegraphics[width=0.23\textwidth]{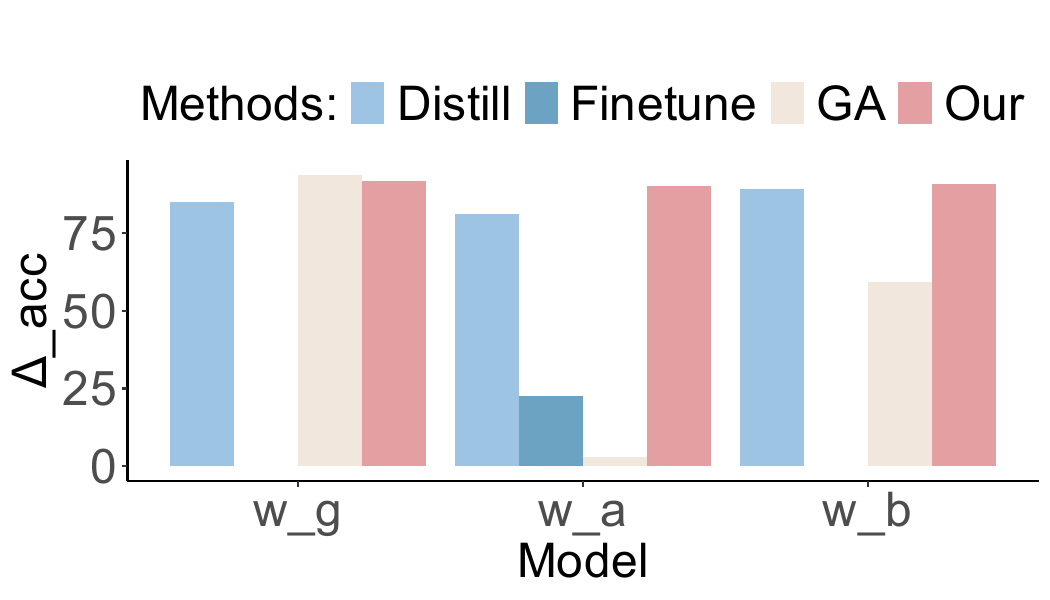}
    }
   \subfigure[IL in ResNet18 of \#$C_f=10$]{
        \centering
        \includegraphics[width=0.23\textwidth]{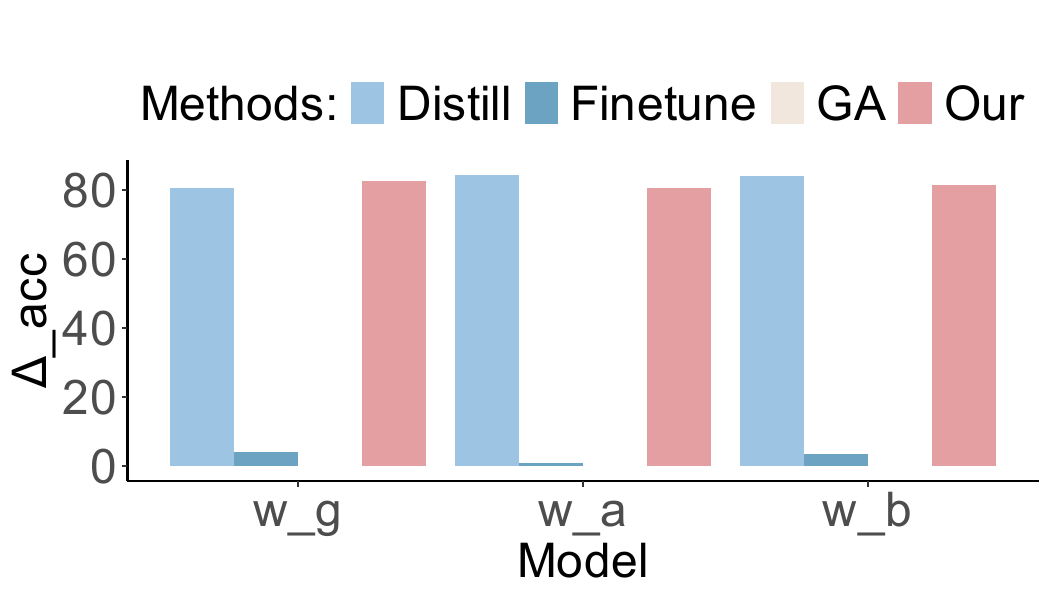}
    }
     \subfigure[IL in AlexNet of \#$C_f=1$]{
        \centering
        \includegraphics[width=0.23\textwidth]{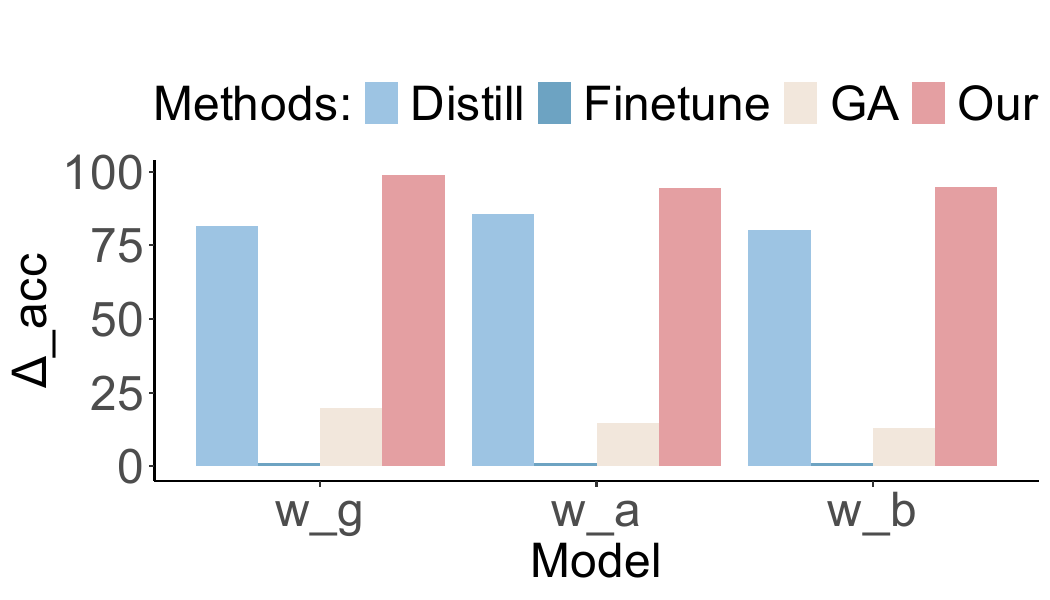}
    }
        \subfigure[IL in AlexNet of \#$C_f=10$]{
        \centering
        \includegraphics[width=0.23\textwidth]{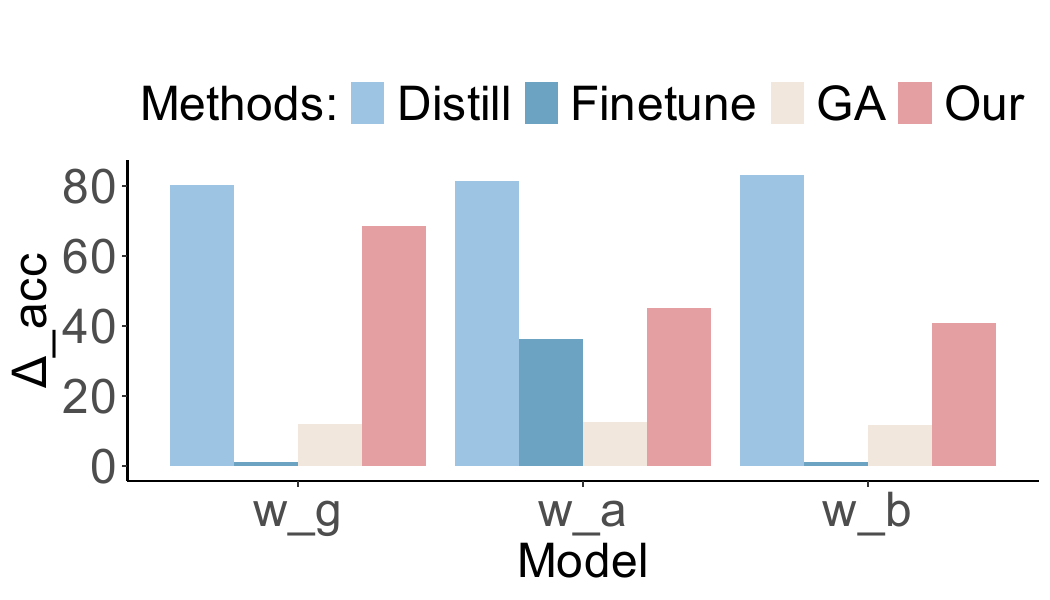}
    }

    \subfigure[TL in ResNet18 of \#$C_f=1$]{
        \centering
        \includegraphics[width=0.23\textwidth]{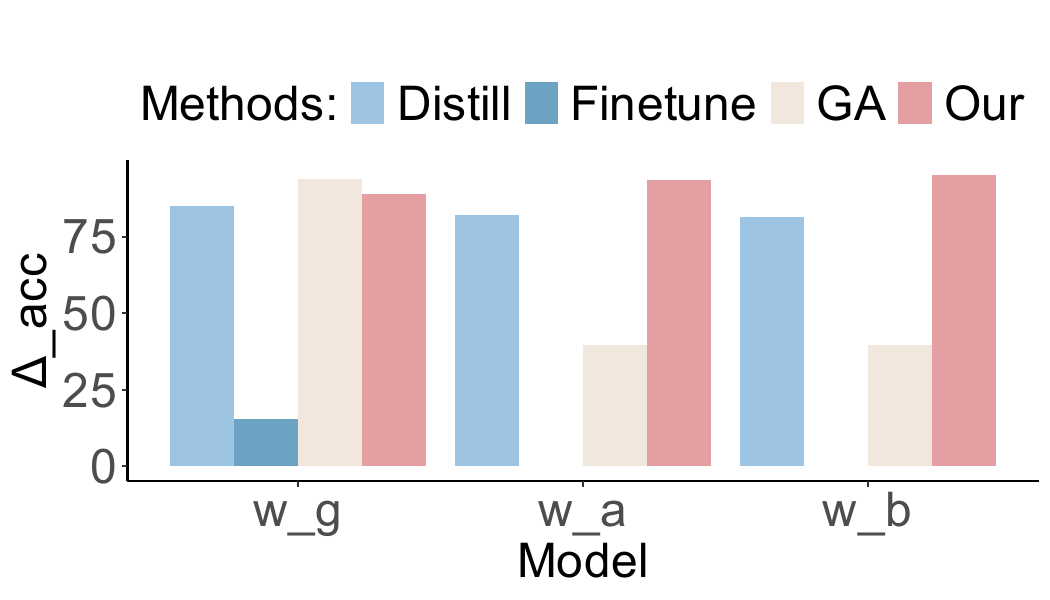}
    }
   \subfigure[TL in ResNet18 of \#$C_f=10$]{
        \centering
        \includegraphics[width=0.23\textwidth]{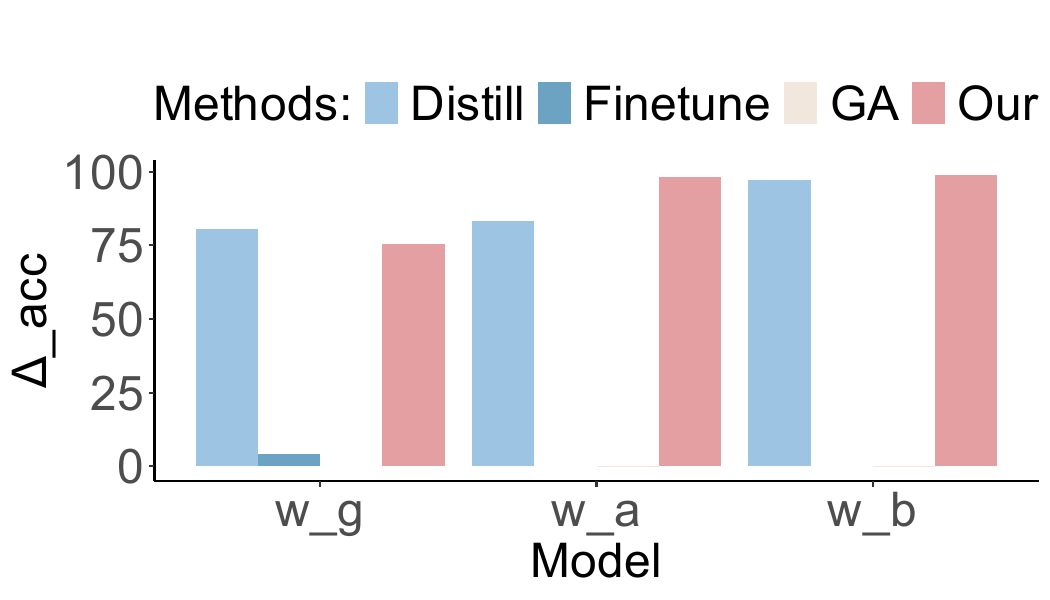}
    }
     \subfigure[TL in AlexNet of \#$C_f=1$]{
        \centering
        \includegraphics[width=0.23\textwidth]{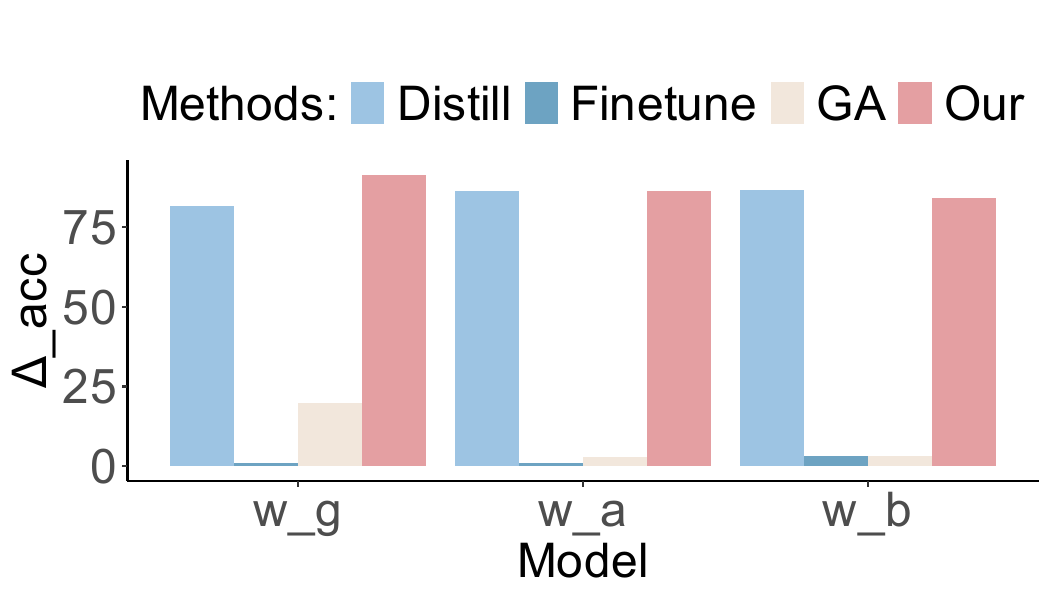}
    }
        \subfigure[TL in AlexNet of \#$C_f=10$]{
        \centering
        \includegraphics[width=0.23\textwidth]{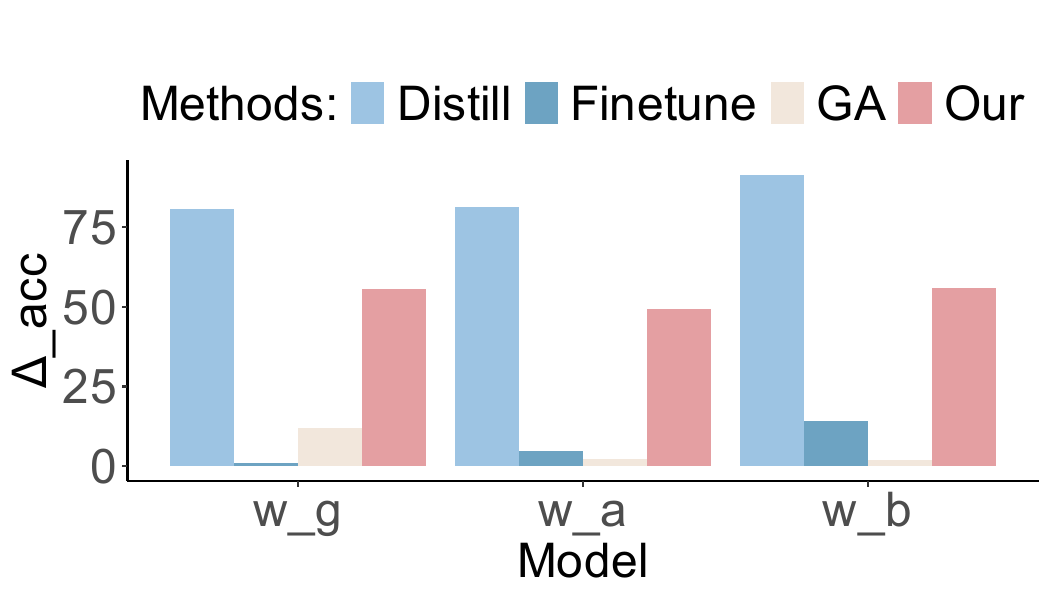}
    }
      
  \caption{Baseline Unlearning Accuracy of IL and TL Frameworks on CIFAR100.}

  \label{fig:baseline}
  \vspace{-1em}
\end{figure*}

\subsection{Framework Setup and Benchmarks}

     

\noindent\textbf{Targeted Frameworks.} We detail setups for four learning frameworks with DAG, all adopting the structure shown in Fig.~\ref{fig:total_Experi_Scenarios_2}. These frameworks vary in their dataset and label distribution conditions.

\begin{packeditemize}
    \item \textbf{Federated Unlearning.} 
    (\textcolor{light_yellow}{Yellow} in Fig.~\ref{fig:total_Experi_Scenarios_2}) The CIFAR100 and TinyImageNet datasets are each randomly divided into five parts. When the model $w_g$ contains labels that need to be unlearned, the models $w_a$ and $w_b$ that inherit from it need to perform the unlearning operation.
    
    \item \textbf{Distributed Data-Parallel  Unlearning.} 
    (\textcolor{light_pink}{Red} in Fig.~\ref{fig:total_Experi_Scenarios_2}) The CIFAR100 and TinyImageNet datasets are each randomly divided into three parts, each executing a sub-task in parallel. When the model $w_g$ contains labels that need to be unlearned, sub-tasks $w_a$ and $w_b$ perform the unlearning operation in parallel.

    \item \textbf{Incremental Unlearning.} 
    (\textcolor{light_blue}{Blue} in Fig.~\ref{fig:total_Experi_Scenarios_2}) In the CIFAR100 and TinyImageNet datasets, the initial models are trained on labels 0-90 and 0-190, respectively, with each inheriting model adding two more classes. Models $w_g$, $w_a$, and $w_b$ are selected for experimental demonstration.

     \item \textbf{Transfer Unlearning.} 
     (\textcolor{light_pink}{Red} in Fig.~\ref{fig:total_Experi_Scenarios_2}) Since TL involves fine-tuning the last layer, adjustments are focused on this layer. In experiments, a model $w_g$ is trained on CIFAR100 labels 0-90, then transferred to models $w_a$ and $w_b$ trained on labels 0-92 and 0-98, respectively. Similarly, in TinyImageNet, model $w_g$ is trained on labels 0-190 and transferred to models $w_a$ and $w_b$ trained on labels 0-192 and 0-198.
    
\end{packeditemize}

\begin{table*}
\centering
\setlength{\extrarowheight}{0pt}
\addtolength{\extrarowheight}{\aboverulesep}
\addtolength{\extrarowheight}{\belowrulesep}
\setlength{\aboverulesep}{0pt}
\setlength{\belowrulesep}{0pt}
\caption{Distributed Data-Parallel Unlearning Performance on CIFAR100}
\label{table:distributed_unlearn}
\resizebox{0.78\textwidth}{!}{
\begin{tabular}{ccc|ccc|ccc|ccc|cccccc} 
\toprule
\multirow{3}{*}{\textbf{Model}}     & \multirow{3}{*}{\#$C_f$} & \multirow{3}{*}{\textbf{Metrics}}                  & \multicolumn{3}{c|}{\multirow{2}{*}{Original~(\%)}}                                                                   & \multicolumn{3}{c|}{\multirow{2}{*}{Re-training~(\%)}}                                                             & \multicolumn{3}{c|}{\multirow{2}{*}{FIUn~(\%)}}                                                                                                  & \multicolumn{6}{c}{\textbf{Cumulative Unlearning Time (s)}}                                                                                                                                                                                                                         \\ 
\cmidrule{13-18}
                                    &                          &                                                    & \multicolumn{3}{c|}{}                                                                                                 & \multicolumn{3}{c|}{}                                                                                              & \multicolumn{3}{c|}{}                                                                                                                            & \multicolumn{3}{c|}{Re-training}                                             & \multicolumn{3}{c}{FIUn}                                                                                                                                                                             \\ 
\cmidrule{4-18}
                                    &                          &                                                    & $w_g$                                 & $w_a$                                 & $w_b$                                 & $w_g$                                & $w_a$                                & $w_b$                                & $w_g$                                          & $w_a$                                          & $w_b$                                          & $w_g$                  & $w_a$                  & \multicolumn{1}{c|}{$w_b$} & $w_g$                                                           & $w_a$                                                           & $w_b$                                                            \\ 
\midrule
\multirow{6}{*}{\rotcell{AlexNet}}  & \multirow{2}{*}{1}       & $AD_r\uparrow$                                     & 98.01                                 & 99.96                                 & 99.96                                 & 99.96                                & 99.96                                & 99.96                                & {\cellcolor[rgb]{1,0.992,0.859}}\textbf{97.23} & {\cellcolor[rgb]{1,0.992,0.859}}\textbf{98.85} & {\cellcolor[rgb]{1,0.992,0.859}}\textbf{97.66} & \multirow{2}{*}{30.13} & \multirow{2}{*}{31.26} & \multirow{2}{*}{31.14}     & {\cellcolor[rgb]{1,0.992,0.859}}                                & {\cellcolor[rgb]{1,0.992,0.859}}                                & {\cellcolor[rgb]{1,0.992,0.859}}                                 \\
                                    &                          & {\cellcolor[rgb]{0.922,0.922,1}}${AD}_f\downarrow$ & {\cellcolor[rgb]{0.922,0.922,1}}99.99 & {\cellcolor[rgb]{0.922,0.922,1}}99.96 & {\cellcolor[rgb]{0.922,0.922,1}}99.96 & {\cellcolor[rgb]{0.922,0.922,1}}0.00 & {\cellcolor[rgb]{0.922,0.922,1}}0.00 & {\cellcolor[rgb]{0.922,0.922,1}}0.00 & {\cellcolor[rgb]{1,0.992,0.859}}\textbf{0.00}  & {\cellcolor[rgb]{1,0.992,0.859}}\textbf{0.00}  & {\cellcolor[rgb]{1,0.992,0.859}}\textbf{0.00}  &                        &                        &                            & \multirow{-2}{*}{{\cellcolor[rgb]{1,0.992,0.859}}\textbf{0.23}} & \multirow{-2}{*}{{\cellcolor[rgb]{1,0.992,0.859}}\textbf{0.14}} & \multirow{-2}{*}{{\cellcolor[rgb]{1,0.992,0.859}}\textbf{0.09}}  \\ 
\hhline{~-----------------}
                                    & \multirow{2}{*}{2}       & $AD_r\uparrow$~                                    & 99.96                                 & 99.96                                 & 99.96                                 & 99.96                                & 99.96                                & 99.96                                & {\cellcolor[rgb]{1,0.992,0.859}}\textbf{96.14} & {\cellcolor[rgb]{1,0.992,0.859}}\textbf{97.07} & {\cellcolor[rgb]{1,0.992,0.859}}\textbf{96.71} & \multirow{2}{*}{26.43} & \multirow{2}{*}{26.47} & \multirow{2}{*}{26.41}     & {\cellcolor[rgb]{1,0.992,0.859}}                                & {\cellcolor[rgb]{1,0.992,0.859}}                                & {\cellcolor[rgb]{1,0.992,0.859}}                                 \\
                                    &                          & {\cellcolor[rgb]{0.922,0.922,1}}${AD}_f\downarrow$ & {\cellcolor[rgb]{0.922,0.922,1}}99.96 & {\cellcolor[rgb]{0.922,0.922,1}}99.96 & {\cellcolor[rgb]{0.922,0.922,1}}99.96 & {\cellcolor[rgb]{0.922,0.922,1}}0.00 & {\cellcolor[rgb]{0.922,0.922,1}}0.00 & {\cellcolor[rgb]{0.922,0.922,1}}0.00 & {\cellcolor[rgb]{1,0.992,0.859}}\textbf{0.00}  & {\cellcolor[rgb]{1,0.992,0.859}}\textbf{0.00}  & {\cellcolor[rgb]{1,0.992,0.859}}\textbf{0.00}  &                        &                        &                            & \multirow{-2}{*}{{\cellcolor[rgb]{1,0.992,0.859}}\textbf{0.27}} & \multirow{-2}{*}{{\cellcolor[rgb]{1,0.992,0.859}}\textbf{0.16}} & \multirow{-2}{*}{{\cellcolor[rgb]{1,0.992,0.859}}\textbf{0.11}}  \\ 
\hhline{~-----------------}
                                    & \multirow{2}{*}{4}       & $AD_r\uparrow$                                     & 98.01                                 & 99.96                                 & 99.96                                 & 99.96                                & 99.96                                & 99.96                                & {\cellcolor[rgb]{1,0.992,0.859}}\textbf{94.58} & {\cellcolor[rgb]{1,0.992,0.859}}\textbf{90.41} & {\cellcolor[rgb]{1,0.992,0.859}}\textbf{90.97} & \multirow{2}{*}{24.33} & \multirow{2}{*}{24.36} & \multirow{2}{*}{24.37}     & {\cellcolor[rgb]{1,0.992,0.859}}                                & {\cellcolor[rgb]{1,0.992,0.859}}                                & {\cellcolor[rgb]{1,0.992,0.859}}                                 \\
                                    &                          & {\cellcolor[rgb]{0.922,0.922,1}}${AD}_f\downarrow$ & {\cellcolor[rgb]{0.922,0.922,1}}99.09 & {\cellcolor[rgb]{0.922,0.922,1}}99.96 & {\cellcolor[rgb]{0.922,0.922,1}}99.96 & {\cellcolor[rgb]{0.922,0.922,1}}0.00 & {\cellcolor[rgb]{0.922,0.922,1}}0.00 & {\cellcolor[rgb]{0.922,0.922,1}}0.00 & {\cellcolor[rgb]{1,0.992,0.859}}\textbf{0.00}  & {\cellcolor[rgb]{1,0.992,0.859}}\textbf{0.00}  & {\cellcolor[rgb]{1,0.992,0.859}}\textbf{0.00}  &                        &                        &                            & \multirow{-2}{*}{{\cellcolor[rgb]{1,0.992,0.859}}\textbf{0.23}} & \multirow{-2}{*}{{\cellcolor[rgb]{1,0.992,0.859}}\textbf{0.14}} & \multirow{-2}{*}{{\cellcolor[rgb]{1,0.992,0.859}}\textbf{0.10}}  \\ 
\midrule
\multirow{6}{*}{\rotcell{ResNet18}} & \multirow{2}{*}{1}       & $AD_r\uparrow$                                     & 99.99                                 & 99.99                                 & 99.99                                 & 99.99                                & 99.99                                & 99.99                                & {\cellcolor[rgb]{1,0.992,0.859}}\textbf{91.87} & {\cellcolor[rgb]{1,0.992,0.859}}\textbf{94.67} & {\cellcolor[rgb]{1,0.992,0.859}}\textbf{93.67} & \multirow{2}{*}{32.53} & \multirow{2}{*}{31.95} & \multirow{2}{*}{32.18}     & {\cellcolor[rgb]{1,0.992,0.859}}                                & {\cellcolor[rgb]{1,0.992,0.859}}                                & {\cellcolor[rgb]{1,0.992,0.859}}                                 \\
                                    &                          & {\cellcolor[rgb]{0.922,0.922,1}}${AD}_f\downarrow$ & {\cellcolor[rgb]{0.922,0.922,1}}99.99 & {\cellcolor[rgb]{0.922,0.922,1}}99.99 & {\cellcolor[rgb]{0.922,0.922,1}}99.99 & {\cellcolor[rgb]{0.922,0.922,1}}0.00 & {\cellcolor[rgb]{0.922,0.922,1}}0.00 & {\cellcolor[rgb]{0.922,0.922,1}}0.00 & {\cellcolor[rgb]{1,0.992,0.859}}\textbf{0.00}  & {\cellcolor[rgb]{1,0.992,0.859}}\textbf{0.00}  & {\cellcolor[rgb]{1,0.992,0.859}}\textbf{0.00}  &                        &                        &                            & \multirow{-2}{*}{{\cellcolor[rgb]{1,0.992,0.859}}\textbf{0.64}} & \multirow{-2}{*}{{\cellcolor[rgb]{1,0.992,0.859}}\textbf{0.34}} & \multirow{-2}{*}{{\cellcolor[rgb]{1,0.992,0.859}}\textbf{0.30}}  \\ 
\hhline{~-----------------}
                                    & \multirow{2}{*}{2}       & $AD_r\uparrow$                                     & 99.99                                 & 99.99                                 & 99.99                                 & 99.99                                & 99.99                                & 99.99                                & {\cellcolor[rgb]{1,0.992,0.859}}\textbf{89.94} & {\cellcolor[rgb]{1,0.992,0.859}}\textbf{94.58} & {\cellcolor[rgb]{1,0.992,0.859}}\textbf{92.40} & \multirow{2}{*}{28.28} & \multirow{2}{*}{27.22} & \multirow{2}{*}{28.47}     & {\cellcolor[rgb]{1,0.992,0.859}}                                & {\cellcolor[rgb]{1,0.992,0.859}}                                & {\cellcolor[rgb]{1,0.992,0.859}}                                 \\
                                    &                          & {\cellcolor[rgb]{0.922,0.922,1}}${AD}_f\downarrow$ & {\cellcolor[rgb]{0.922,0.922,1}}99.99 & {\cellcolor[rgb]{0.922,0.922,1}}99.99 & {\cellcolor[rgb]{0.922,0.922,1}}99.99 & {\cellcolor[rgb]{0.922,0.922,1}}0.00 & {\cellcolor[rgb]{0.922,0.922,1}}0.00 & {\cellcolor[rgb]{0.922,0.922,1}}0.00 & {\cellcolor[rgb]{1,0.992,0.859}}\textbf{0.00}  & {\cellcolor[rgb]{1,0.992,0.859}}\textbf{0.00}  & {\cellcolor[rgb]{1,0.992,0.859}}\textbf{0.00}  &                        &                        &                            & \multirow{-2}{*}{{\cellcolor[rgb]{1,0.992,0.859}}\textbf{0.66}} & \multirow{-2}{*}{{\cellcolor[rgb]{1,0.992,0.859}}\textbf{0.34}} & \multirow{-2}{*}{{\cellcolor[rgb]{1,0.992,0.859}}\textbf{0.30}}  \\ 
\hhline{~-----------------}
                                    & \multirow{2}{*}{4}       & $AD_r\uparrow$                                     & 99.96                                 & 89.20                                 & 88.52                                 & 99.96                                & 99.96                                & 86.44                                & {\cellcolor[rgb]{1,0.992,0.859}}\textbf{93.95} & {\cellcolor[rgb]{1,0.992,0.859}}\textbf{89.26} & {\cellcolor[rgb]{1,0.992,0.859}}\textbf{91.77} & \multirow{2}{*}{25.37} & \multirow{2}{*}{24.97} & \multirow{2}{*}{25.13}     & {\cellcolor[rgb]{1,0.992,0.859}}                                & {\cellcolor[rgb]{1,0.992,0.859}}                                & {\cellcolor[rgb]{1,0.992,0.859}}                                 \\
                                    &                          & {\cellcolor[rgb]{0.922,0.922,1}}${AD}_f\downarrow$ & {\cellcolor[rgb]{0.922,0.922,1}}99.99 & {\cellcolor[rgb]{0.922,0.922,1}}99.99 & {\cellcolor[rgb]{0.922,0.922,1}}99.99 & {\cellcolor[rgb]{0.922,0.922,1}}0.00 & {\cellcolor[rgb]{0.922,0.922,1}}0.00 & {\cellcolor[rgb]{0.922,0.922,1}}0.00 & {\cellcolor[rgb]{1,0.992,0.859}}\textbf{0.00}  & {\cellcolor[rgb]{1,0.992,0.859}}\textbf{0.00}  & {\cellcolor[rgb]{1,0.992,0.859}}\textbf{0.00}  &                        &                        &                            & \multirow{-2}{*}{{\cellcolor[rgb]{1,0.992,0.859}}\textbf{0.66}} & \multirow{-2}{*}{{\cellcolor[rgb]{1,0.992,0.859}}\textbf{0.35}} & \multirow{-2}{*}{{\cellcolor[rgb]{1,0.992,0.859}}\textbf{0.31}}  \\
\bottomrule
\end{tabular}
}
\end{table*}

\noindent\textbf{Benchmark.} 
The experiment specifically focuses on class-level unlearning tasks for clarity of presentation. We re-train the models that include unlearned labels and subsequent models influenced by these labels due to inheritance. We remove the data with unlearned labels from each model's dataset. The remaining data is then used to train each model individually. The models are trained sequentially according to their inheritance relationships.

\noindent\textbf{Hyperparameters.}
Table~\ref{table: hyperparameter} summarizes the hyperparameters for various models and our FIUn method.
All hyperparameters remain fixed during experiments to ensure fair comparisons across models.



\subsection{Label Category Settings}


The unlearned label categories, denoted as $\#C_{f}$, may vary across different models and data distributions. For instance, model $w_a$ aims to unlearn labels 2, 3, and 4, while model $w_b$ intends to unlearn labels 2, 3, and 6. There is an overlap between the unlearned label categories of model $w_a$ and model $w_b$, specifically labels 2 and 3. We refer to the degree of overlapping among these unlearned label categories as $\#\cap_{f}$.
The unlearning process is applied to individual and combined labels. For CIFAR100 and Yahoo! Answers, we select combinations of 1, 2, and 4 categories, while for TinyImageNet, we select 1, 4, and 6 categories. The overlap of unlearned labels is divided into 60\%, 40\%, and 20\%.

\vspace{-0.2em}
\section{Evaluation}\label{sec:experimental results}

We use an NVIDIA 4090 GPU to evaluate FIUn’s impact on the accuracy of unlearned and retained labels across four frameworks: FL, DDPL, IL, and TL. Accuracy and time consumption during unlearning are assessed by comparing FIUn with existing methods, followed by an analysis of re-training, the most effective (but inefficient) unlearning benchmark, focusing on inheritance depth and parameter layers impact on unlearning effectiveness.

\subsection{General Label Unlearning Analysis}

We consider existing emerging unlearning methods for comparison: 1) fine-tuning, which adjusts the original model using the remaining dataset~\cite{odusami2021analysis}, 2) gradient ascent (GA), employing negative gradients for unlearning~\cite{ga}, and 3) distill, which distills knowledge from the original model into a student model using the remaining data~\cite{distil1Unl}.


Table~\ref{table:fed_baseline} reports unlearning speed and Fig.\ref{fig:baseline1} shows unlearning accuracy $\bigtriangleup_{acc}$; Fig.\ref{fig:baseline} compares $\bigtriangleup_{acc}$ under IL and TL. Our method improves unlearning speed by up to 99\% over alternatives and maintains a consistent accuracy advantage: for single-label and up to 10 classes, GA and fine-tuning finish in about 2 seconds, distillation and retraining are slower, while our approach averages under 1 second. In accuracy, fine-tuning and GA remain suboptimal, and our method matches or surpasses distillation across most architectures (with a slight dip on AlexNet). Overall, it delivers the fastest cumulative unlearning across IMNs. Building on this, we next compare with retraining, which is widely regarded as the gold standard for unlearning and retention accuracy~\cite{sisa}; see Tables~\ref{table:transfer_unlearn1} and~\ref{table:continue_unlearn_TinyImageNet}, with additional results in \textbf{Appendix~\ref{appendix_b1}}.


\begin{table*}[!ht]
\centering
\setlength{\extrarowheight}{0pt}
\addtolength{\extrarowheight}{\aboverulesep}
\addtolength{\extrarowheight}{\belowrulesep}
\setlength{\aboverulesep}{0pt}
\setlength{\belowrulesep}{0pt}
\caption{Federated Unlearning Performance on TinyImageNet}
\label{table:fedrated_TinyImageNet}
\resizebox{0.8\textwidth}{!}{
\begin{tabular}{ccc|ccc|ccc|ccc|cccccc} 
\toprule
\multirow{3}{*}{\textbf{Model}}        & \multirow{3}{*}{\#$C_f$} & \multirow{3}{*}{\textbf{Metrics}}                  & \multicolumn{3}{c|}{\multirow{2}{*}{Original~(\%)}}                                                                   & \multicolumn{3}{c|}{\multirow{2}{*}{Re-training~(\%)}}                                                             & \multicolumn{3}{c|}{\multirow{2}{*}{FIUn~(\%)}}                                                                                                  & \multicolumn{6}{c}{\textbf{Cumulative Unlearning Time (s)}}                                                                                                                                                                                                                             \\ 
\cmidrule{13-18}
                                       &                          &                                                    & \multicolumn{3}{c|}{}                                                                                                 & \multicolumn{3}{c|}{}                                                                                              & \multicolumn{3}{c|}{}                                                                                                                            & \multicolumn{3}{c|}{Re-training}                                                 & \multicolumn{3}{c}{FIUn}                                                                                                                                                                             \\ 
\cmidrule{4-18}
                                       &                          &                                                    & $w_g$                                 & $w_a$                                 & $w_b$                                 & $w_g$                                & $w_a$                                & $w_b$                                & $w_g$                                          & $w_a$                                          & $w_b$                                          & $w_g$                    & $w_a$                    & \multicolumn{1}{c|}{$w_b$} & $w_g$                                                           & $w_a$                                                           & $w_b$                                                            \\ 
\midrule
\multirow{6}{*}{\rotcell{DenseNet161}} & \multirow{2}{*}{1}       & $AD_r\uparrow$                                     & 99.99                                 & 99.99                                 & 99.81                                 & 99.99                                & 99.99                                & 98.80                                & {\cellcolor[rgb]{1,0.992,0.859}}\textbf{88.03} & {\cellcolor[rgb]{1,0.992,0.859}}\textbf{77.64} & {\cellcolor[rgb]{1,0.992,0.859}}\textbf{76.02} & \multirow{2}{*}{999.41}  & \multirow{2}{*}{1996.43}  & \multirow{2}{*}{2997.64}    & {\cellcolor[rgb]{1,0.992,0.859}}                                & {\cellcolor[rgb]{1,0.992,0.859}}                                & {\cellcolor[rgb]{1,0.992,0.859}}                                 \\
                                       &                          & {\cellcolor[rgb]{0.922,0.922,1}}${AD}_f\downarrow$ & {\cellcolor[rgb]{0.922,0.922,1}}99.99 & {\cellcolor[rgb]{0.922,0.922,1}}99.99 & {\cellcolor[rgb]{0.922,0.922,1}}99.66 & {\cellcolor[rgb]{0.922,0.922,1}}0.00 & {\cellcolor[rgb]{0.922,0.922,1}}0.00 & {\cellcolor[rgb]{0.922,0.922,1}}0.00 & {\cellcolor[rgb]{1,0.992,0.859}}\textbf{0.00}  & {\cellcolor[rgb]{1,0.992,0.859}}\textbf{0.00}  & {\cellcolor[rgb]{1,0.992,0.859}}\textbf{0.00}  &                          &                          &                            & \multirow{-2}{*}{{\cellcolor[rgb]{1,0.992,0.859}}\textbf{2.51}} & \multirow{-2}{*}{{\cellcolor[rgb]{1,0.992,0.859}}\textbf{4.59}} & \multirow{-2}{*}{{\cellcolor[rgb]{1,0.992,0.859}}\textbf{4.95}}  \\ 
\hhline{~-----------------}
                                       & \multirow{2}{*}{4}       & $AD_r\uparrow$~                                    & 99.99                                 & 99.99                                 & 99.80                                 & 99.99                                & 99.99                                & 99.51                                & {\cellcolor[rgb]{1,0.992,0.859}}\textbf{83.01} & {\cellcolor[rgb]{1,0.992,0.859}}\textbf{71.43} & {\cellcolor[rgb]{1,0.992,0.859}}\textbf{71} & \multirow{2}{*}{1004.64} & \multirow{2}{*}{2006.31} & \multirow{2}{*}{3006.14}   & {\cellcolor[rgb]{1,0.992,0.859}}                                & {\cellcolor[rgb]{1,0.992,0.859}}                                & {\cellcolor[rgb]{1,0.992,0.859}}                                 \\
                                       &                          & {\cellcolor[rgb]{0.922,0.922,1}}${AD}_f\downarrow$ & {\cellcolor[rgb]{0.922,0.922,1}}99.99 & {\cellcolor[rgb]{0.922,0.922,1}}99.99 & {\cellcolor[rgb]{0.922,0.922,1}}99.49 & {\cellcolor[rgb]{0.922,0.922,1}}0.00 & {\cellcolor[rgb]{0.922,0.922,1}}0.00 & {\cellcolor[rgb]{0.922,0.922,1}}0.00 & {\cellcolor[rgb]{1,0.992,0.859}}\textbf{0.00}  & {\cellcolor[rgb]{1,0.992,0.859}}\textbf{0.00}  & {\cellcolor[rgb]{1,0.992,0.859}}\textbf{0.00}  &                          &                          &                            & \multirow{-2}{*}{{\cellcolor[rgb]{1,0.992,0.859}}\textbf{3.16}} & \multirow{-2}{*}{{\cellcolor[rgb]{1,0.992,0.859}}\textbf{4.69}} & \multirow{-2}{*}{{\cellcolor[rgb]{1,0.992,0.859}}\textbf{4.89}}  \\ 
\hhline{~-----------------}
                                       & \multirow{2}{*}{6}       & $AD_r\uparrow$                                     & 99.99                                 & 99.99                                 & 99.96                                 & 99.99                                & 99.99                                & 99.16                                & {\cellcolor[rgb]{1,0.992,0.859}}\textbf{80.00} & {\cellcolor[rgb]{1,0.992,0.859}}\textbf{67.67} & {\cellcolor[rgb]{1,0.992,0.859}}\textbf{66.58} & \multirow{2}{*}{1019.43} & \multirow{2}{*}{2020.16} & \multirow{2}{*}{3018.64}   & {\cellcolor[rgb]{1,0.992,0.859}}                                & {\cellcolor[rgb]{1,0.992,0.859}}                                & {\cellcolor[rgb]{1,0.992,0.859}}                                 \\
                                       &                          & {\cellcolor[rgb]{0.922,0.922,1}}${AD}_f\downarrow$ & {\cellcolor[rgb]{0.922,0.922,1}}99.99 & {\cellcolor[rgb]{0.922,0.922,1}}99.99 & {\cellcolor[rgb]{0.922,0.922,1}}99.89 & {\cellcolor[rgb]{0.922,0.922,1}}0.00 & {\cellcolor[rgb]{0.922,0.922,1}}0.00 & {\cellcolor[rgb]{0.922,0.922,1}}0.00 & {\cellcolor[rgb]{1,0.992,0.859}}\textbf{0.00}  & {\cellcolor[rgb]{1,0.992,0.859}}\textbf{0.00}  & {\cellcolor[rgb]{1,0.992,0.859}}\textbf{0.00}  &                          &                          &                            & \multirow{-2}{*}{{\cellcolor[rgb]{1,0.992,0.859}}\textbf{3.10}} & \multirow{-2}{*}{{\cellcolor[rgb]{1,0.992,0.859}}\textbf{4.96}} & \multirow{-2}{*}{{\cellcolor[rgb]{1,0.992,0.859}}\textbf{4.75}}  \\ 
\midrule
\multirow{6}{*}{\rotcell{ResNet18}}    & \multirow{2}{*}{1}       & $AD_r\uparrow$                                     & 99.99                                 & 99.99                                 & 99.99                                 & 99.99                                & 99.99                                & 99.99                                & {\cellcolor[rgb]{1,0.992,0.859}}\textbf{99.73} & {\cellcolor[rgb]{1,0.992,0.859}}\textbf{99.97} & {\cellcolor[rgb]{1,0.992,0.859}}\textbf{99.46} & \multirow{2}{*}{112.95}   & \multirow{2}{*}{216.39}   & \multirow{2}{*}{314.29}     & {\cellcolor[rgb]{1,0.992,0.859}}                                & {\cellcolor[rgb]{1,0.992,0.859}}                                & {\cellcolor[rgb]{1,0.992,0.859}}                                 \\
                                       &                          & {\cellcolor[rgb]{0.922,0.922,1}}${AD}_f\downarrow$ & {\cellcolor[rgb]{0.922,0.922,1}}99.99 & {\cellcolor[rgb]{0.922,0.922,1}}99.99 & {\cellcolor[rgb]{0.922,0.922,1}}99.99 & {\cellcolor[rgb]{0.922,0.922,1}}0.00 & {\cellcolor[rgb]{0.922,0.922,1}}0.00 & {\cellcolor[rgb]{0.922,0.922,1}}0.00 & {\cellcolor[rgb]{1,0.992,0.859}}\textbf{0.00}  & {\cellcolor[rgb]{1,0.992,0.859}}\textbf{0.00}  & {\cellcolor[rgb]{1,0.992,0.859}}\textbf{0.00}  &                          &                          &                            & \multirow{-2}{*}{{\cellcolor[rgb]{1,0.992,0.859}}\textbf{1.35}} & \multirow{-2}{*}{{\cellcolor[rgb]{1,0.992,0.859}}\textbf{3.68}} & \multirow{-2}{*}{{\cellcolor[rgb]{1,0.992,0.859}}\textbf{3.42}}  \\ 
\hhline{~-----------------}
                                       & \multirow{2}{*}{4}       & $AD_r\uparrow$                                     & 99.99                                 & 99.99                                 & 99.99                                 & 99.99                                & 99.99                                & 99.99                                & {\cellcolor[rgb]{1,0.992,0.859}}\textbf{99.17} & {\cellcolor[rgb]{1,0.992,0.859}}\textbf{99.98} & {\cellcolor[rgb]{1,0.992,0.859}}\textbf{99.98} & \multirow{2}{*}{109.89}   & \multirow{2}{*}{209.31}   & \multirow{2}{*}{310.42}     & {\cellcolor[rgb]{1,0.992,0.859}}                                & {\cellcolor[rgb]{1,0.992,0.859}}                                & {\cellcolor[rgb]{1,0.992,0.859}}                                 \\
                                       &                          & {\cellcolor[rgb]{0.922,0.922,1}}${AD}_f\downarrow$ & {\cellcolor[rgb]{0.922,0.922,1}}99.99 & {\cellcolor[rgb]{0.922,0.922,1}}99.99 & {\cellcolor[rgb]{0.922,0.922,1}}99.99 & {\cellcolor[rgb]{0.922,0.922,1}}0.00 & {\cellcolor[rgb]{0.922,0.922,1}}0.00 & {\cellcolor[rgb]{0.922,0.922,1}}0.00 & {\cellcolor[rgb]{1,0.992,0.859}}\textbf{0.00}  & {\cellcolor[rgb]{1,0.992,0.859}}\textbf{0.00}  & {\cellcolor[rgb]{1,0.992,0.859}}\textbf{0.00}  &                          &                          &                            & \multirow{-2}{*}{{\cellcolor[rgb]{1,0.992,0.859}}\textbf{1.63}} & \multirow{-2}{*}{{\cellcolor[rgb]{1,0.992,0.859}}\textbf{3.26}} & \multirow{-2}{*}{{\cellcolor[rgb]{1,0.992,0.859}}\textbf{3.57}}  \\ 
\hhline{~-----------------}
                                       & \multirow{2}{*}{6}       & $AD_r\uparrow$                                     & 99.99                                 & 99.99                                 & 99.99                                 & 99.96                                & 99.99                                & 99.99                                & {\cellcolor[rgb]{1,0.992,0.859}}\textbf{99.54} & {\cellcolor[rgb]{1,0.992,0.859}}\textbf{99.98} & {\cellcolor[rgb]{1,0.992,0.859}}\textbf{99.98} & \multirow{2}{*}{107.39}   & \multirow{2}{*}{208.12}   & \multirow{2}{*}{305.32}     & {\cellcolor[rgb]{1,0.992,0.859}}                                & {\cellcolor[rgb]{1,0.992,0.859}}                                & {\cellcolor[rgb]{1,0.992,0.859}}                                 \\
                                       &                          & {\cellcolor[rgb]{0.922,0.922,1}}${AD}_f\downarrow$ & {\cellcolor[rgb]{0.922,0.922,1}}99.99 & {\cellcolor[rgb]{0.922,0.922,1}}99.99 & {\cellcolor[rgb]{0.922,0.922,1}}99.99 & {\cellcolor[rgb]{0.922,0.922,1}}0.00 & {\cellcolor[rgb]{0.922,0.922,1}}0.00 & {\cellcolor[rgb]{0.922,0.922,1}}0.00 & {\cellcolor[rgb]{1,0.992,0.859}}\textbf{0.00}  & {\cellcolor[rgb]{1,0.992,0.859}}\textbf{0.00}  & {\cellcolor[rgb]{1,0.992,0.859}}\textbf{0.00}  &                          &                          &                            & \multirow{-2}{*}{{\cellcolor[rgb]{1,0.992,0.859}}\textbf{1.46}} & \multirow{-2}{*}{{\cellcolor[rgb]{1,0.992,0.859}}\textbf{3.37}} & \multirow{-2}{*}{{\cellcolor[rgb]{1,0.992,0.859}}\textbf{3.51}}  \\
\bottomrule
\end{tabular}
}
\end{table*}

\begin{table*}
\centering
\setlength{\extrarowheight}{0pt}
\addtolength{\extrarowheight}{\aboverulesep}
\addtolength{\extrarowheight}{\belowrulesep}
\setlength{\aboverulesep}{0pt}
\setlength{\belowrulesep}{0pt}
\caption{Incremental Unlearning Performance on TinyImageNet}
\label{table:continue_unlearn_TinyImageNet}
\resizebox{0.78\textwidth}{!}{
\begin{tabular}{c|c|c|ccccccccc|cccccc} 
\hline
\multirow{3}{*}{Model}                 & \multirow{3}{*}{\#$C_f$} & \multirow{3}{*}{Metrics}                                                & \multicolumn{3}{c|}{\multirow{2}{*}{Original (\%)}}                                                                   & \multicolumn{3}{c|}{\multirow{2}{*}{Re-training (\%)}}                                                             & \multicolumn{3}{c|}{\multirow{2}{*}{FIUn~(\%)}}                                                                                      & \multicolumn{6}{c}{Cumulative Unlearning Time (s)}                                                                                                                                                                                                                              \\ 
\cline{13-18}
                                       &                          &                                                                         & \multicolumn{3}{c|}{}                                                                                                 & \multicolumn{3}{c|}{}                                                                                              & \multicolumn{3}{c|}{}                                                                                                                & \multicolumn{3}{c|}{Re-training}                                                     & \multicolumn{3}{c}{FIUn}                                                                                                                                                                 \\ 
\cline{4-18}
                                       &                          &                                                                         & \multicolumn{1}{c|}{$w_g$}            & \multicolumn{1}{c|}{$w_a$}            & \multicolumn{1}{c|}{$w_b$}            & \multicolumn{1}{c|}{$w_g$}           & \multicolumn{1}{c|}{$w_a$}           & \multicolumn{1}{c|}{$w_b$}           & \multicolumn{1}{c|}{$w_g$}                 & \multicolumn{1}{c|}{$w_a$}                 & $w_b$                                      & \multicolumn{1}{c|}{$w_g$} & \multicolumn{1}{c|}{$w_a$} & \multicolumn{1}{c|}{$w_b$} & \multicolumn{1}{c|}{$w_g$}                                  & \multicolumn{1}{c|}{$w_a$}                                  & $w_b$                                                        \\ 
\hline
\multirow{6}{*}{\rotcell{DenseNet161}} & \multirow{2}{*}{1}       & $AD_r\uparrow$                                                          & 99.99                                 & 99.99                                 & 99.81                                 & 99.99                                & 99.99                                & 98.80                                & {\cellcolor[rgb]{1,1,0.878}}\textbf{89.76} & {\cellcolor[rgb]{1,1,0.878}}\textbf{78.50} & {\cellcolor[rgb]{1,1,0.878}}\textbf{73.87} & \multirow{2}{*}{991.43}    & \multirow{2}{*}{1995.36}   & \multirow{2}{*}{3989.47}   & {\cellcolor[rgb]{1,1,0.878}}                                & {\cellcolor[rgb]{1,1,0.878}}                                & {\cellcolor[rgb]{1,1,0.878}}                                 \\
                                       &                          & \multicolumn{1}{l|}{{\cellcolor[rgb]{0.933,0.929,1}}${AD}_f\downarrow$} & {\cellcolor[rgb]{0.933,0.929,1}}99.99 & {\cellcolor[rgb]{0.933,0.929,1}}99.99 & {\cellcolor[rgb]{0.933,0.929,1}}99.66 & {\cellcolor[rgb]{0.933,0.929,1}}0.00 & {\cellcolor[rgb]{0.933,0.929,1}}0.00 & {\cellcolor[rgb]{0.933,0.929,1}}0.00 & {\cellcolor[rgb]{1,1,0.878}}\textbf{0.00}  & {\cellcolor[rgb]{1,1,0.878}}\textbf{0.00}  & {\cellcolor[rgb]{1,1,0.878}}\textbf{0.00}  &                            &                            &                            & \multirow{-2}{*}{{\cellcolor[rgb]{1,1,0.878}}\textbf{2.75}} & \multirow{-2}{*}{{\cellcolor[rgb]{1,1,0.878}}\textbf{4.85}} & \multirow{-2}{*}{{\cellcolor[rgb]{1,1,0.878}}\textbf{4.43}}  \\ 
\hhline{~-----------------}
                                       & \multirow{2}{*}{4}       & $AD_r\uparrow$~                                                         & 99.99                                 & 99.99                                 & 99.80                                 & 99.99                                & 99.99                                & 99.51                                & {\cellcolor[rgb]{1,1,0.878}}\textbf{83.98} & {\cellcolor[rgb]{1,1,0.878}}\textbf{65.87} & {\cellcolor[rgb]{1,1,0.878}}\textbf{59.11} & \multirow{2}{*}{1006.32}   & \multirow{2}{*}{2004.75}   & \multirow{2}{*}{4009.41}   & {\cellcolor[rgb]{1,1,0.878}}                                & {\cellcolor[rgb]{1,1,0.878}}                                & {\cellcolor[rgb]{1,1,0.878}}                                 \\
                                       &                          & \multicolumn{1}{l|}{{\cellcolor[rgb]{0.933,0.929,1}}${AD}_f\downarrow$} & {\cellcolor[rgb]{0.933,0.929,1}}99.99 & {\cellcolor[rgb]{0.933,0.929,1}}99.99 & {\cellcolor[rgb]{0.933,0.929,1}}99.49 & {\cellcolor[rgb]{0.933,0.929,1}}0.00 & {\cellcolor[rgb]{0.933,0.929,1}}0.00 & {\cellcolor[rgb]{0.933,0.929,1}}0.00 & {\cellcolor[rgb]{1,1,0.878}}\textbf{0.00}  & {\cellcolor[rgb]{1,1,0.878}}\textbf{0.00}  & {\cellcolor[rgb]{1,1,0.878}}\textbf{0.00}  &                            &                            &                            & \multirow{-2}{*}{{\cellcolor[rgb]{1,1,0.878}}\textbf{2.64}} & \multirow{-2}{*}{{\cellcolor[rgb]{1,1,0.878}}\textbf{4.43}} & \multirow{-2}{*}{{\cellcolor[rgb]{1,1,0.878}}\textbf{4.64}}  \\ 
\hhline{~-----------------}
                                       & \multirow{2}{*}{6}       & $AD_r\uparrow$                                                          & 99.99                                 & 99.99                                 & 99.96                                 & 99.99                                & 99.99                                & 99.16                                & {\cellcolor[rgb]{1,1,0.878}}\textbf{86.81} & {\cellcolor[rgb]{1,1,0.878}}\textbf{61.02} & {\cellcolor[rgb]{1,1,0.878}}\textbf{55.04} & \multirow{2}{*}{1019.43}   & \multirow{2}{*}{2019.50}   & \multirow{2}{*}{4019.46}   & {\cellcolor[rgb]{1,1,0.878}}                                & {\cellcolor[rgb]{1,1,0.878}}                                & {\cellcolor[rgb]{1,1,0.878}}                                 \\
                                       &                          & \multicolumn{1}{l|}{{\cellcolor[rgb]{0.933,0.933,1}}${AD}_f\downarrow$} & {\cellcolor[rgb]{0.933,0.933,1}}99.99 & {\cellcolor[rgb]{0.933,0.933,1}}99.99 & {\cellcolor[rgb]{0.933,0.933,1}}99.89 & {\cellcolor[rgb]{0.933,0.933,1}}0.00 & {\cellcolor[rgb]{0.933,0.933,1}}0.00 & {\cellcolor[rgb]{0.933,0.933,1}}0.00 & {\cellcolor[rgb]{1,1,0.878}}\textbf{0.00}  & {\cellcolor[rgb]{1,1,0.878}}\textbf{0.00}  & {\cellcolor[rgb]{1,1,0.878}}\textbf{0.00}  &                            &                            &                            & \multirow{-2}{*}{{\cellcolor[rgb]{1,1,0.878}}\textbf{2.36}} & \multirow{-2}{*}{{\cellcolor[rgb]{1,1,0.878}}\textbf{4.47}} & \multirow{-2}{*}{{\cellcolor[rgb]{1,1,0.878}}\textbf{4.74}}  \\ 
\hline
\multirow{6}{*}{\rotcell{ResNet18}}    & \multirow{2}{*}{1}       & $AD_r\uparrow$                                                          & 99.99                                 & 99.99                                 & 99.99                                 & 99.99                                & 99.99                                & 99.99                                & {\cellcolor[rgb]{1,1,0.878}}\textbf{98.66} & {\cellcolor[rgb]{1,1,0.878}}\textbf{96.33} & {\cellcolor[rgb]{1,1,0.878}}\textbf{96.07} & \multirow{2}{*}{120.32}    & \multirow{2}{*}{221.75}    & \multirow{2}{*}{439.85}    & {\cellcolor[rgb]{1,1,0.878}}                                & {\cellcolor[rgb]{1,1,0.878}}                                & {\cellcolor[rgb]{1,1,0.878}}                                 \\
                                       &                          & \multicolumn{1}{l|}{{\cellcolor[rgb]{0.933,0.933,1}}${AD}_f\downarrow$} & {\cellcolor[rgb]{0.933,0.933,1}}99.99 & {\cellcolor[rgb]{0.933,0.933,1}}99.99 & {\cellcolor[rgb]{0.933,0.933,1}}99.99 & {\cellcolor[rgb]{0.933,0.933,1}}0.00 & {\cellcolor[rgb]{0.933,0.933,1}}0.00 & {\cellcolor[rgb]{0.933,0.933,1}}0.00 & {\cellcolor[rgb]{1,1,0.878}}\textbf{0.00}  & {\cellcolor[rgb]{1,1,0.878}}\textbf{0.00}  & {\cellcolor[rgb]{1,1,0.878}}\textbf{0.00}  &                            &                            &                            & \multirow{-2}{*}{{\cellcolor[rgb]{1,1,0.878}}\textbf{1.37}} & \multirow{-2}{*}{{\cellcolor[rgb]{1,1,0.878}}\textbf{3.75}} & \multirow{-2}{*}{{\cellcolor[rgb]{1,1,0.878}}\textbf{3.43}}  \\ 
\hhline{~-----------------}
                                       & \multirow{2}{*}{4}       & $AD_r\uparrow$                                                          & 99.99                                 & 99.99                                 & 99.99                                 & 99.99                                & 99.99                                & 99.99                                & {\cellcolor[rgb]{1,1,0.878}}\textbf{98.84} & {\cellcolor[rgb]{1,1,0.878}}\textbf{96.34} & {\cellcolor[rgb]{1,1,0.878}}\textbf{96.70} & \multirow{2}{*}{115.32}    & \multirow{2}{*}{217.33}    & \multirow{2}{*}{435.17}    & {\cellcolor[rgb]{1,1,0.878}}                                & {\cellcolor[rgb]{1,1,0.878}}                                & {\cellcolor[rgb]{1,1,0.878}}                                 \\
                                       &                          & \multicolumn{1}{l|}{{\cellcolor[rgb]{0.933,0.933,1}}${AD}_f\downarrow$} & {\cellcolor[rgb]{0.933,0.933,1}}99.99 & {\cellcolor[rgb]{0.933,0.933,1}}99.99 & {\cellcolor[rgb]{0.933,0.933,1}}99.99 & {\cellcolor[rgb]{0.933,0.933,1}}0.00 & {\cellcolor[rgb]{0.933,0.933,1}}0.00 & {\cellcolor[rgb]{0.933,0.933,1}}0.00 & {\cellcolor[rgb]{1,1,0.878}}\textbf{0.00}  & {\cellcolor[rgb]{1,1,0.878}}\textbf{0.00}  & {\cellcolor[rgb]{1,1,0.878}}\textbf{0.00}  &                            &                            &                            & \multirow{-2}{*}{{\cellcolor[rgb]{1,1,0.878}}\textbf{1.41}} & \multirow{-2}{*}{{\cellcolor[rgb]{1,1,0.878}}\textbf{3.34}} & \multirow{-2}{*}{{\cellcolor[rgb]{1,1,0.878}}\textbf{3.64}}  \\ 
\hhline{~-----------------}
                                       & \multirow{2}{*}{6}       & $AD_r\uparrow$                                                          & 99.99                                 & 99.99                                 & 99.99                                 & 99.96                                & 99.99                                & 99.99                                & {\cellcolor[rgb]{1,1,0.878}}\textbf{98.82} & {\cellcolor[rgb]{1,1,0.878}}\textbf{99.99} & {\cellcolor[rgb]{1,1,0.878}}\textbf{98.59} & \multirow{2}{*}{110.32}    & \multirow{2}{*}{211.35}    & \multirow{2}{*}{431.37}    & {\cellcolor[rgb]{1,1,0.878}}                                & {\cellcolor[rgb]{1,1,0.878}}                                & {\cellcolor[rgb]{1,1,0.878}}                                 \\
                                       &                          & \multicolumn{1}{l|}{{\cellcolor[rgb]{0.933,0.933,1}}${AD}_f\downarrow$} & {\cellcolor[rgb]{0.933,0.933,1}}99.99 & {\cellcolor[rgb]{0.933,0.933,1}}99.99 & {\cellcolor[rgb]{0.933,0.933,1}}99.99 & {\cellcolor[rgb]{0.933,0.933,1}}0.00 & {\cellcolor[rgb]{0.933,0.933,1}}0.00 & {\cellcolor[rgb]{0.933,0.933,1}}0.00 & {\cellcolor[rgb]{1,1,0.878}}\textbf{0.00}  & {\cellcolor[rgb]{1,1,0.878}}\textbf{0.00}  & {\cellcolor[rgb]{1,1,0.878}}\textbf{0.00}  &                            &                            &                            & \multirow{-2}{*}{{\cellcolor[rgb]{1,1,0.878}}\textbf{1.47}} & \multirow{-2}{*}{{\cellcolor[rgb]{1,1,0.878}}\textbf{3.46}} & \multirow{-2}{*}{{\cellcolor[rgb]{1,1,0.878}}\textbf{3.75}}  \\
\hline
\end{tabular}}
\vspace{-0.5em}
\end{table*}

\begin{figure*}[!ht]
 \centering
  \subfigure[FL in CIFAR100 ResNet18]{
         \centering
         \includegraphics[width=0.23\textwidth]{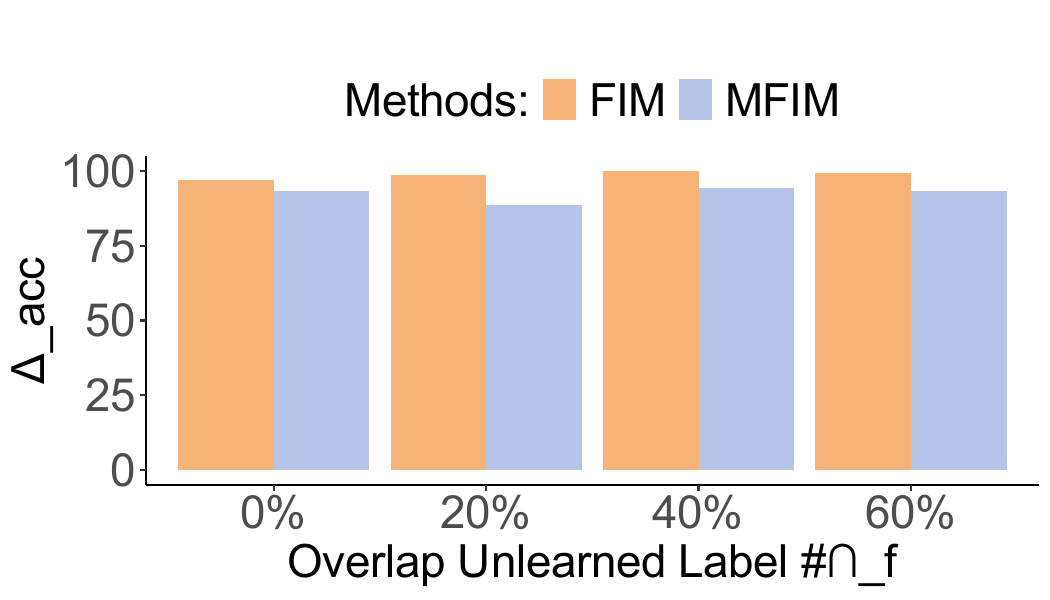}
         \label{fig:MultiFLResNetCifar}
     }
     \subfigure[FL CIFAR100 by AlexNet]{
         \centering\includegraphics[width=0.23\textwidth]{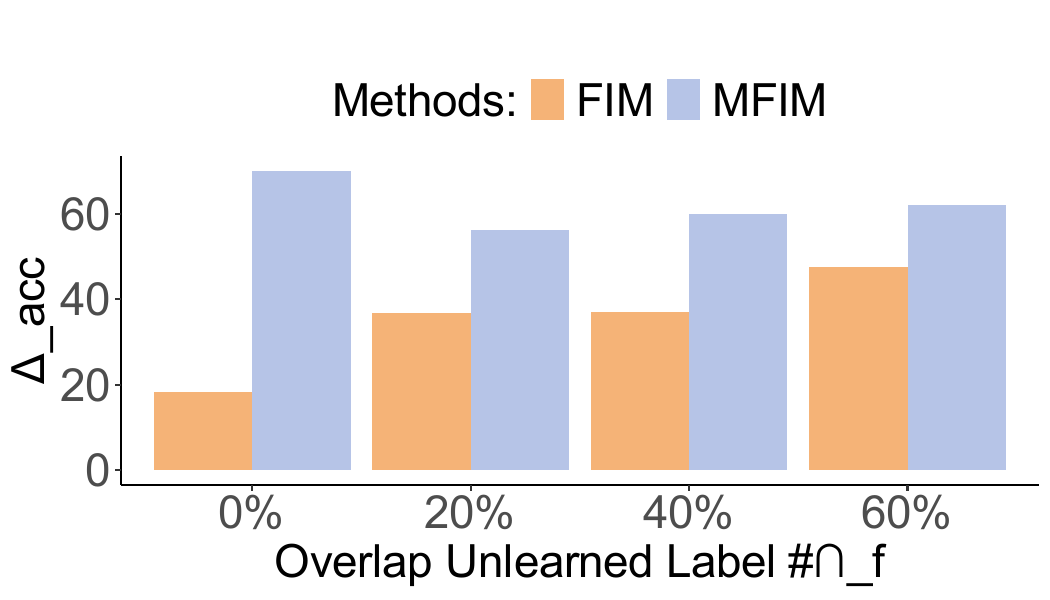}
         \label{fig:multicifaralex}
     }
       \subfigure[FL TinyImageNet by ResNet18]{
         \centering
         \includegraphics[width=0.23\textwidth]{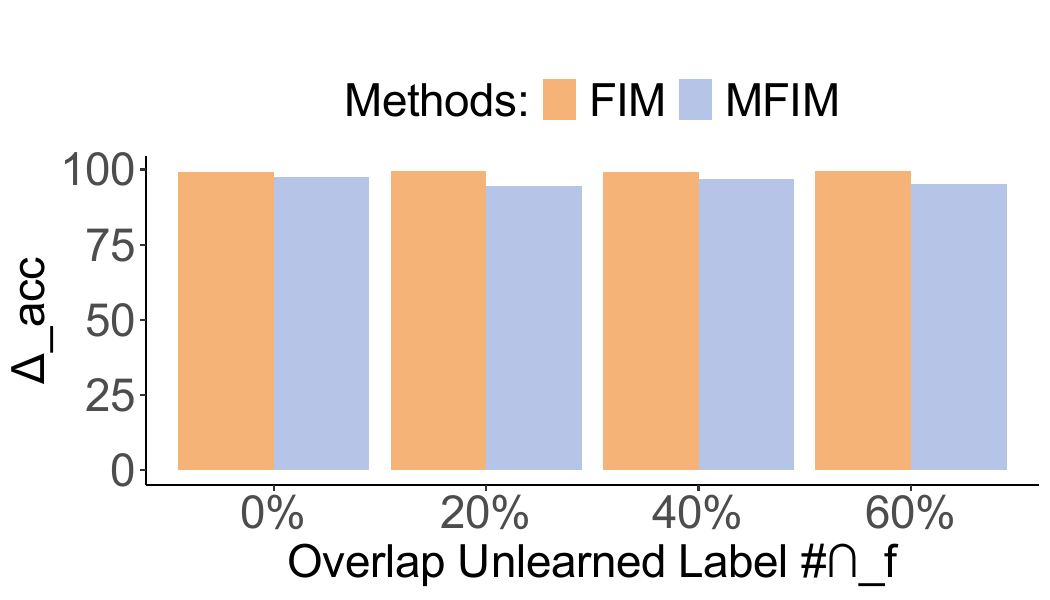}
        \label{fig:multiFedImage}
     }
    \subfigure[FL TinyImageNet by DenseNet161]{
        \centering
        \includegraphics[width=0.23\textwidth]{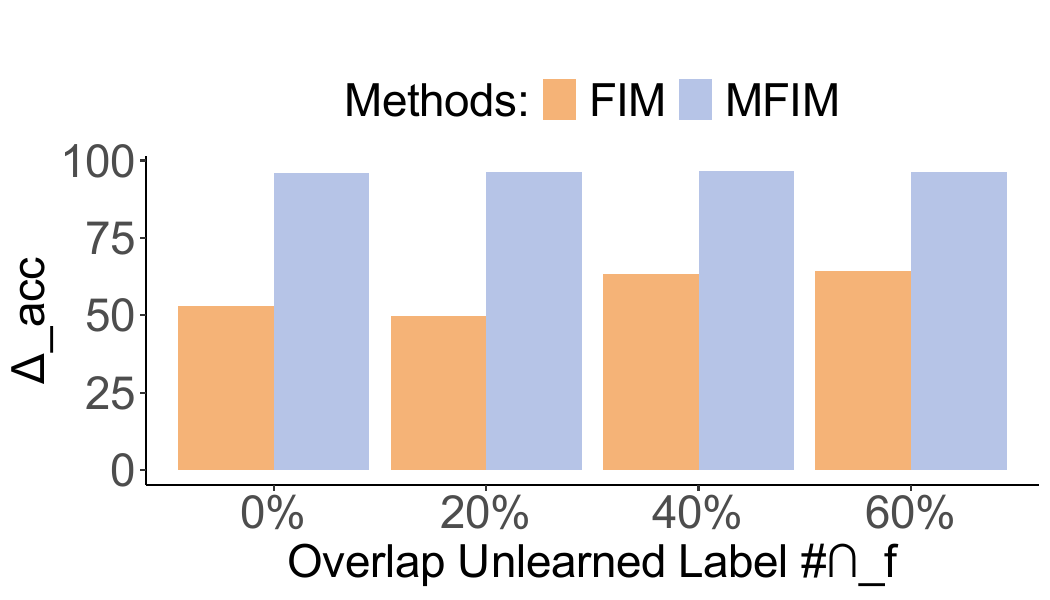}
       
        \label{fig:multiFLCifardense}
    }
  \caption{Overlap \#$\cap_f$ analysis, \#$C_f$=15. MFIM stands for Merging FIM function.}
  \label{fig:overlap_analysis}
  \vspace{-1em}
\end{figure*}



 \smallskip
\textit{1) Single-label unlearning.} For all frameworks and model types in CIFAR100, experimental results show that the $AD_f$ metric reaches 0, strongly proving that our method effectively unlearns single-class labels. Meanwhile, the average $AD_r$ metric is as high as 94.53\%, further confirming our method's efficiency in retaining labels. For all frameworks and model types in TinyImageNet, the $AD_f$ metric also reaches 0, and the average $AD_r$ metric reaches 79.49\%. This validates the effectiveness of our method.

\textit{2) Multi-label unlearning.} 
For the CIFAR100 dataset, with \#$C_f = 2, 4$, the average $AD_f$ metric across all frameworks and model types is 1.93\%, while the average $AD_r$ metric reaches 86.01\%. For the TinyImageNet dataset, the average $AD_f$ metric is 0.21\%, and the average $AD_r$ metric is 83.54\%. These results further demonstrate the effectiveness of our method.
On the other hand, for both the IL and TL frameworks, as the number of unlearning label classes increases, the $AD_f$ metric for all models approaches 0. However, the $AD_r$ performance of the AlexNet and DenseNet161 models shows a downward trend, while the $AD_r$ performance of the ResNet18 model remains stable.

\begin{table*}
\centering
\setlength{\extrarowheight}{0pt}
\addtolength{\extrarowheight}{\aboverulesep}
\addtolength{\extrarowheight}{\belowrulesep}
\setlength{\aboverulesep}{0pt}
\setlength{\belowrulesep}{0pt}
\caption{Distributed Data-Parallel Unlearning Performance on CIFAR100 with Multiple Label Distribution}
\label{table:dis_cifar100_multi}
\resizebox{0.8\textwidth}{!}{
\begin{tabular}{c|c|c|ccccccccc|cccccc} 
\hline
\multirow{3}{*}{Model}                 & \multirow{3}{*}{$\#\cap_{f}$} & \multirow{3}{*}{Metrics}                                                & \multicolumn{3}{c|}{\multirow{2}{*}{Original (\%)}}                                                                   & \multicolumn{3}{c|}{\multirow{2}{*}{Re-training (\%)}}                                                             & \multicolumn{3}{c|}{\multirow{2}{*}{FIUn~(\%)}}                                                                                                        & \multicolumn{6}{c}{Cumulative Unlearning Time (s)}                                                                                                                                                                                                                                                      \\ 
\cmidrule{13-18}
                                       &                                                                   &                                                  & \multicolumn{3}{c|}{}                                                                                                 & \multicolumn{3}{c|}{}                                                                                              & \multicolumn{3}{c|}{}                                                                                                                            & \multicolumn{3}{c|}{Re-training}                                                 & \multicolumn{3}{c}{FIUn}                                                                                                                                                                             \\ 
\cmidrule{4-18}
                                       &                                                                   &                                                  & $w_g$                                 & $w_a$                                 & $w_b$                                 & $w_g$                                & $w_a$                                & $w_b$                                & $w_g$                                          & $w_a$                                          & $w_b$                                          & $w_g$                    & $w_a$                    & \multicolumn{1}{c|}{$w_b$} & $w_g$                                                           & $w_a$                                                           & $w_b$                                                            \\ 
\midrule
\multirow{6}{*}{\rotcell{DenseNet161}} & \multirow{2}{*}{60\%}                                             & $AD_r\uparrow$                                   & 99.99                                 & 99.99                                 & 99.81                                 & 98.80                                & 99.99                                & 99.99                                & {\cellcolor[rgb]{1,0.992,0.859}}\textbf{95.18} & {\cellcolor[rgb]{1,0.992,0.859}}\textbf{97.07} & {\cellcolor[rgb]{1,0.992,0.859}}\textbf{99.99} & \multirow{2}{*}{999.41}  & \multirow{2}{*}{996.43}  & \multirow{2}{*}{997.64}    & {\cellcolor[rgb]{1,0.992,0.859}}                                & {\cellcolor[rgb]{1,0.992,0.859}}                                & {\cellcolor[rgb]{1,0.992,0.859}}                                 \\
                                       &                                                                   & {\cellcolor[rgb]{0.922,0.922,1}}$AD_f\downarrow$ & {\cellcolor[rgb]{0.922,0.922,1}}99.99 & {\cellcolor[rgb]{0.922,0.922,1}}99.99 & {\cellcolor[rgb]{0.922,0.922,1}}99.66 & {\cellcolor[rgb]{0.922,0.922,1}}0.00 & {\cellcolor[rgb]{0.922,0.922,1}}0.00 & {\cellcolor[rgb]{0.922,0.922,1}}0.00 & {\cellcolor[rgb]{1,0.992,0.859}}\textbf{0.00}  & {\cellcolor[rgb]{1,0.992,0.859}}\textbf{0.00}  & {\cellcolor[rgb]{1,0.992,0.859}}\textbf{0.00}  &                          &                          &                            & \multirow{-2}{*}{{\cellcolor[rgb]{1,0.992,0.859}}\textbf{0.15}} & \multirow{-2}{*}{{\cellcolor[rgb]{1,0.992,0.859}}\textbf{0.08}} & \multirow{-2}{*}{{\cellcolor[rgb]{1,0.992,0.859}}\textbf{0.09}}  \\ 
\hhline{~-----------------}
                                       & \multirow{2}{*}{40\%}                                             & $AD_r\uparrow$~                                  & 99.99                                 & 99.99                                 & 99.80                                 & 99.51                                & 99.99                                & 99.99                                & {\cellcolor[rgb]{1,0.992,0.859}}\textbf{90.53} & {\cellcolor[rgb]{1,0.992,0.859}}\textbf{97.07} & {\cellcolor[rgb]{1,0.992,0.859}}\textbf{98.82} & \multirow{2}{*}{1004.64} & \multirow{2}{*}{1006.31} & \multirow{2}{*}{1006.14}   & {\cellcolor[rgb]{1,0.992,0.859}}                                & {\cellcolor[rgb]{1,0.992,0.859}}                                & {\cellcolor[rgb]{1,0.992,0.859}}                                 \\
                                       &                                                                   & {\cellcolor[rgb]{0.922,0.922,1}}$AD_f\downarrow$ & {\cellcolor[rgb]{0.922,0.922,1}}99.99 & {\cellcolor[rgb]{0.922,0.922,1}}99.99 & {\cellcolor[rgb]{0.922,0.922,1}}99.49 & {\cellcolor[rgb]{0.922,0.922,1}}0.00 & {\cellcolor[rgb]{0.922,0.922,1}}0.00 & {\cellcolor[rgb]{0.922,0.922,1}}0.00 & {\cellcolor[rgb]{1,0.992,0.859}}\textbf{0.00}  & {\cellcolor[rgb]{1,0.992,0.859}}\textbf{0.00}  & {\cellcolor[rgb]{1,0.992,0.859}}\textbf{0.00}  &                          &                          &                            & \multirow{-2}{*}{{\cellcolor[rgb]{1,0.992,0.859}}\textbf{0.19}} & \multirow{-2}{*}{{\cellcolor[rgb]{1,0.992,0.859}}\textbf{0.09}} & \multirow{-2}{*}{{\cellcolor[rgb]{1,0.992,0.859}}\textbf{0.11}}  \\ 
\hhline{~-----------------}
                                       & \multirow{2}{*}{20\%}                                             & $AD_r\uparrow$                                   & 99.99                                 & 99.99                                 & 99.96                                 & 99.16                                & 99.99                                & 99.99                                & {\cellcolor[rgb]{1,0.992,0.859}}\textbf{95.29} & {\cellcolor[rgb]{1,0.992,0.859}}\textbf{97.07} & {\cellcolor[rgb]{1,0.992,0.859}}\textbf{99.89} & \multirow{2}{*}{1019.43} & \multirow{2}{*}{1020.16} & \multirow{2}{*}{1018.64}   & {\cellcolor[rgb]{1,0.992,0.859}}                                & {\cellcolor[rgb]{1,0.992,0.859}}                                & {\cellcolor[rgb]{1,0.992,0.859}}                                 \\
                                       &                                                                   & {\cellcolor[rgb]{0.922,0.922,1}}$AD_f\downarrow$ & {\cellcolor[rgb]{0.922,0.922,1}}99.99 & {\cellcolor[rgb]{0.922,0.922,1}}99.99 & {\cellcolor[rgb]{0.922,0.922,1}}99.89 & {\cellcolor[rgb]{0.922,0.922,1}}0.00 & {\cellcolor[rgb]{0.922,0.922,1}}0.00 & {\cellcolor[rgb]{0.922,0.922,1}}0.00 & {\cellcolor[rgb]{1,0.992,0.859}}\textbf{0.00}  & {\cellcolor[rgb]{1,0.992,0.859}}\textbf{0.00}  & {\cellcolor[rgb]{1,0.992,0.859}}\textbf{0.00}  &                          &                          &                            & \multirow{-2}{*}{{\cellcolor[rgb]{1,0.992,0.859}}\textbf{0.18}} & \multirow{-2}{*}{{\cellcolor[rgb]{1,0.992,0.859}}\textbf{0.07}} & \multirow{-2}{*}{{\cellcolor[rgb]{1,0.992,0.859}}\textbf{0.10}}  \\ 
\midrule
\multirow{6}{*}{\rotcell{ResNet18}}    & \multirow{2}{*}{60\%}                                             & $AD_r\uparrow$                                   & 99.99                                 & 99.99                                 & 99.99                                 & 99.99                                & 99.99                                & 99.99                                & {\cellcolor[rgb]{1,0.992,0.859}}\textbf{97.62} & {\cellcolor[rgb]{1,0.992,0.859}}\textbf{88.94} & {\cellcolor[rgb]{1,0.992,0.859}}\textbf{87.87} & \multirow{2}{*}{30.36}   & \multirow{2}{*}{33.18}   & \multirow{2}{*}{37.31}     & {\cellcolor[rgb]{1,0.992,0.859}}                                & {\cellcolor[rgb]{1,0.992,0.859}}                                & {\cellcolor[rgb]{1,0.992,0.859}}                                 \\
                                       &                                                                   & {\cellcolor[rgb]{0.922,0.922,1}}$AD_f\downarrow$ & {\cellcolor[rgb]{0.922,0.922,1}}99.99 & {\cellcolor[rgb]{0.922,0.922,1}}99.99 & {\cellcolor[rgb]{0.922,0.922,1}}99.99 & {\cellcolor[rgb]{0.922,0.922,1}}0.00 & {\cellcolor[rgb]{0.922,0.922,1}}0.00 & {\cellcolor[rgb]{0.922,0.922,1}}0.00 & {\cellcolor[rgb]{1,0.992,0.859}}\textbf{0.00}  & {\cellcolor[rgb]{1,0.992,0.859}}\textbf{0.00}  & {\cellcolor[rgb]{1,0.992,0.859}}\textbf{0.00}  &                          &                          &                            & \multirow{-2}{*}{{\cellcolor[rgb]{1,0.992,0.859}}\textbf{0.60}} & \multirow{-2}{*}{{\cellcolor[rgb]{1,0.992,0.859}}\textbf{0.32}} & \multirow{-2}{*}{{\cellcolor[rgb]{1,0.992,0.859}}\textbf{0.29}}  \\ 
\hhline{~-----------------}
                                       & \multirow{2}{*}{40\%}                                             & $AD_r\uparrow$                                   & 99.99                                 & 99.99                                 & 99.99                                 & 99.99                                & 99.99                                & 99.99                                & {\cellcolor[rgb]{1,0.992,0.859}}\textbf{96.50} & {\cellcolor[rgb]{1,0.992,0.859}}\textbf{89.35} & {\cellcolor[rgb]{1,0.992,0.859}}\textbf{87.89} & \multirow{2}{*}{28.30}   & \multirow{2}{*}{29.13}   & \multirow{2}{*}{52.33}     & {\cellcolor[rgb]{1,0.992,0.859}}                                & {\cellcolor[rgb]{1,0.992,0.859}}                                & {\cellcolor[rgb]{1,0.992,0.859}}                                 \\
                                       &                                                                   & {\cellcolor[rgb]{0.922,0.922,1}}$AD_f\downarrow$ & {\cellcolor[rgb]{0.922,0.922,1}}99.99 & {\cellcolor[rgb]{0.922,0.922,1}}99.99 & {\cellcolor[rgb]{0.922,0.922,1}}99.99 & {\cellcolor[rgb]{0.922,0.922,1}}0.00 & {\cellcolor[rgb]{0.922,0.922,1}}0.00 & {\cellcolor[rgb]{0.922,0.922,1}}0.00 & {\cellcolor[rgb]{1,0.992,0.859}}\textbf{0.00}  & {\cellcolor[rgb]{1,0.992,0.859}}\textbf{0.00}  & {\cellcolor[rgb]{1,0.992,0.859}}\textbf{0.00}  &                          &                          &                            & \multirow{-2}{*}{{\cellcolor[rgb]{1,0.992,0.859}}\textbf{0.59}} & \multirow{-2}{*}{{\cellcolor[rgb]{1,0.992,0.859}}\textbf{0.34}} & \multirow{-2}{*}{{\cellcolor[rgb]{1,0.992,0.859}}\textbf{0.30}}  \\ 
\hhline{~-----------------}
                                       & \multirow{2}{*}{20\%}                                             & $AD_r\uparrow$                                   & 99.99                                 & 99.99                                 & 99.99                                 & 99.96                                & 99.99                                & 99.99                                & {\cellcolor[rgb]{1,0.992,0.859}}\textbf{95.31} & {\cellcolor[rgb]{1,0.992,0.859}}\textbf{88.17} & {\cellcolor[rgb]{1,0.992,0.859}}\textbf{93.06} & \multirow{2}{*}{26.39}   & \multirow{2}{*}{23.37}   & \multirow{2}{*}{56.61}     & {\cellcolor[rgb]{1,0.992,0.859}}                                & {\cellcolor[rgb]{1,0.992,0.859}}                                & {\cellcolor[rgb]{1,0.992,0.859}}                                 \\
                                       &                                                                   & {\cellcolor[rgb]{0.922,0.922,1}}$AD_f\downarrow$ & {\cellcolor[rgb]{0.922,0.922,1}}99.99 & {\cellcolor[rgb]{0.922,0.922,1}}99.99 & {\cellcolor[rgb]{0.922,0.922,1}}99.99 & {\cellcolor[rgb]{0.922,0.922,1}}0.00 & {\cellcolor[rgb]{0.922,0.922,1}}0.00 & {\cellcolor[rgb]{0.922,0.922,1}}0.00 & {\cellcolor[rgb]{1,0.992,0.859}}\textbf{0.00}  & {\cellcolor[rgb]{1,0.992,0.859}}\textbf{0.00}  & {\cellcolor[rgb]{1,0.992,0.859}}\textbf{0.00}  &                          &                          &                            & \multirow{-2}{*}{{\cellcolor[rgb]{1,0.992,0.859}}\textbf{0.58}} & \multirow{-2}{*}{{\cellcolor[rgb]{1,0.992,0.859}}\textbf{0.28}} & \multirow{-2}{*}{{\cellcolor[rgb]{1,0.992,0.859}}\textbf{0.30}}  \\
\bottomrule
\end{tabular}}
\vspace{-1em}
\end{table*}

\begin{table*}
\centering
\setlength{\extrarowheight}{0pt}
\addtolength{\extrarowheight}{\aboverulesep}
\addtolength{\extrarowheight}{\belowrulesep}
\setlength{\aboverulesep}{0pt}
\setlength{\belowrulesep}{0pt}
\caption{Federated Unlearning Performance on TinyImageNet with Multiple Label Distribution}
\label{table:fedrated_tinyimage_multi}
\resizebox{0.8\textwidth}{!}{
\begin{tabular}{c|c|c|ccccccccc|cccccc} 
\hline
\multirow{3}{*}{Model}                 & \multirow{3}{*}{$\#\cap_{f}$} & \multirow{3}{*}{Metrics}                                                & \multicolumn{3}{c|}{\multirow{2}{*}{Original (\%)}}                                                                   & \multicolumn{3}{c|}{\multirow{2}{*}{Re-training (\%)}}                                                             & \multicolumn{3}{c|}{\multirow{2}{*}{FIUn~(\%)}}                                                                                                        & \multicolumn{6}{c}{Cumulative Unlearning Time (s)}                                                                                                                                                                                                                                                \\ 
\cline{13-18}
                                       &                               &                                                                         & \multicolumn{3}{c|}{}                                                                                                 & \multicolumn{3}{c|}{}                                                                                              & \multicolumn{3}{c|}{}                                                                                                                                  & \multicolumn{3}{c|}{Re-training}                                                     & \multicolumn{3}{c}{FIUn}                                                                                                                                                                                   \\ 
\cline{4-18}
                                       &                               &                                                                         & \multicolumn{1}{c|}{$w_g$}            & \multicolumn{1}{c|}{$w_a$}            & \multicolumn{1}{c|}{$w_b$}            & \multicolumn{1}{c|}{$w_g$}           & \multicolumn{1}{c|}{$w_a$}           & \multicolumn{1}{c|}{$w_b$}           & \multicolumn{1}{c|}{$w_g$}                          & \multicolumn{1}{c|}{$w_a$}                 & $w_b$                                               & \multicolumn{1}{c|}{$w_g$} & \multicolumn{1}{c|}{$w_a$} & \multicolumn{1}{c|}{$w_b$} & \multicolumn{1}{c|}{$w_g$}                                           & \multicolumn{1}{c|}{$w_a$}                                  & $w_b$                                                                 \\ 
\hline
\multirow{6}{*}{\rotcell{DenseNet161}} & \multirow{2}{*}{60\%}         & $AD_r\uparrow$                                                          & 99.99                                 & 99.99                                 & 99.81                                 & 98.80                                & 99.99                                & 99.99                                & {\cellcolor[rgb]{1,1,0.878}}\textbf{\textbf{96.39}} & {\cellcolor[rgb]{1,1,0.878}}\textbf{84.75} & {\cellcolor[rgb]{1,1,0.878}}\textbf{\textbf{77.89}} & \multirow{2}{*}{3000.66}   & \multirow{2}{*}{1998.46}   & \multirow{2}{*}{994.32}    & {\cellcolor[rgb]{1,1,0.878}}                                         & {\cellcolor[rgb]{1,1,0.878}}                                & {\cellcolor[rgb]{1,1,0.878}}                                          \\
                                       &                               & \multicolumn{1}{l|}{{\cellcolor[rgb]{0.933,0.933,1}}${AD}_f\downarrow$} & {\cellcolor[rgb]{0.933,0.933,1}}99.99 & {\cellcolor[rgb]{0.933,0.933,1}}99.99 & {\cellcolor[rgb]{0.933,0.933,1}}99.66 & {\cellcolor[rgb]{0.933,0.933,1}}0.00 & {\cellcolor[rgb]{0.933,0.933,1}}0.00 & {\cellcolor[rgb]{0.933,0.933,1}}0.00 & {\cellcolor[rgb]{1,1,0.878}}\textbf{0.00}           & {\cellcolor[rgb]{1,1,0.878}}\textbf{0.00}  & {\cellcolor[rgb]{1,1,0.878}}\textbf{0.00}           &                            &                            &                            & \multirow{-2}{*}{{\cellcolor[rgb]{1,1,0.878}}\textbf{\textbf{4.69}}} & \multirow{-2}{*}{{\cellcolor[rgb]{1,1,0.878}}\textbf{2.39}} & \multirow{-2}{*}{{\cellcolor[rgb]{1,1,0.878}}\textbf{\textbf{2.30}}}  \\ 
\hhline{~-----------------}
                                       & \multirow{2}{*}{40\%}         & $AD_r\uparrow$~                                                         & 99.50                                 & 99.99                                 & 99.80                                 & 99.51                                & 99.99                                & 99.99                                & {\cellcolor[rgb]{1,1,0.878}}\textbf{\textbf{96.64}} & {\cellcolor[rgb]{1,1,0.878}}\textbf{85.40} & {\cellcolor[rgb]{1,1,0.878}}\textbf{\textbf{85.00}} & \multirow{2}{*}{3009.53}   & \multirow{2}{*}{1010.45}   & \multirow{2}{*}{2011.14}   & {\cellcolor[rgb]{1,1,0.878}}                                         & {\cellcolor[rgb]{1,1,0.878}}                                & {\cellcolor[rgb]{1,1,0.878}}                                          \\
                                       &                               & \multicolumn{1}{l|}{{\cellcolor[rgb]{0.933,0.933,1}}${AD}_f\downarrow$} & {\cellcolor[rgb]{0.933,0.933,1}}99.99 & {\cellcolor[rgb]{0.933,0.933,1}}99.99 & {\cellcolor[rgb]{0.933,0.933,1}}99.49 & {\cellcolor[rgb]{0.933,0.933,1}}0.00 & {\cellcolor[rgb]{0.933,0.933,1}}0.00 & {\cellcolor[rgb]{0.933,0.933,1}}0.00 & {\cellcolor[rgb]{1,1,0.878}}\textbf{0.00}           & {\cellcolor[rgb]{1,1,0.878}}\textbf{0.00}  & {\cellcolor[rgb]{1,1,0.878}}\textbf{0.00}           &                            &                            &                            & \multirow{-2}{*}{{\cellcolor[rgb]{1,1,0.878}}\textbf{\textbf{5.12}}} & \multirow{-2}{*}{{\cellcolor[rgb]{1,1,0.878}}\textbf{2.69}} & \multirow{-2}{*}{{\cellcolor[rgb]{1,1,0.878}}\textbf{\textbf{2.14}}}  \\ 
\hhline{~-----------------}
                                       & \multirow{2}{*}{20\%}         & $AD_r\uparrow$                                                          & 99.99                                 & 99.99                                 & 99.96                                 & 99.16                                & 99.99                                & 99.99                                & {\cellcolor[rgb]{1,1,0.878}}\textbf{\textbf{96.22}} & {\cellcolor[rgb]{1,1,0.878}}\textbf{81.49} & {\cellcolor[rgb]{1,1,0.878}}\textbf{\textbf{78.83}} & \multirow{2}{*}{3022.64}   & \multirow{2}{*}{2020.67}   & \multirow{2}{*}{1021.34}   & {\cellcolor[rgb]{1,1,0.878}}                                         & {\cellcolor[rgb]{1,1,0.878}}                                & {\cellcolor[rgb]{1,1,0.878}}                                          \\
                                       &                               & \multicolumn{1}{l|}{{\cellcolor[rgb]{0.933,0.933,1}}${AD}_f\downarrow$} & {\cellcolor[rgb]{0.933,0.933,1}}99.99 & {\cellcolor[rgb]{0.933,0.933,1}}99.99 & {\cellcolor[rgb]{0.933,0.933,1}}99.89 & {\cellcolor[rgb]{0.933,0.933,1}}0.00 & {\cellcolor[rgb]{0.933,0.933,1}}0.00 & {\cellcolor[rgb]{0.933,0.933,1}}0.00 & {\cellcolor[rgb]{1,1,0.878}}\textbf{0.00}           & {\cellcolor[rgb]{1,1,0.878}}\textbf{0.00}  & {\cellcolor[rgb]{1,1,0.878}}\textbf{0.00}           &                            &                            &                            & \multirow{-2}{*}{{\cellcolor[rgb]{1,1,0.878}}\textbf{\textbf{4.98}}} & \multirow{-2}{*}{{\cellcolor[rgb]{1,1,0.878}}\textbf{2.64}} & \multirow{-2}{*}{{\cellcolor[rgb]{1,1,0.878}}\textbf{\textbf{2.59}}}  \\ 
\hline
\multirow{6}{*}{\rotcell{ResNet18}}    & \multirow{2}{*}{60\%}         & $AD_r\uparrow$                                                          & 99.99                                 & 99.99                                 & 99.99                                 & 99.99                                & 99.99                                & 99.99                                & {\cellcolor[rgb]{1,1,0.878}}\textbf{\textbf{95.23}} & {\cellcolor[rgb]{1,1,0.878}}\textbf{98.54} & {\cellcolor[rgb]{1,1,0.878}}\textbf{\textbf{98.44}} & \multirow{2}{*}{349.32}    & \multirow{2}{*}{246.43}    & \multirow{2}{*}{123.74}    & {\cellcolor[rgb]{1,1,0.878}}                                         & {\cellcolor[rgb]{1,1,0.878}}                                & {\cellcolor[rgb]{1,1,0.878}}                                          \\
                                       &                               & \multicolumn{1}{l|}{{\cellcolor[rgb]{0.933,0.933,1}}${AD}_f\downarrow$} & {\cellcolor[rgb]{0.933,0.933,1}}99.99 & {\cellcolor[rgb]{0.933,0.933,1}}99.99 & {\cellcolor[rgb]{0.933,0.933,1}}99.99 & {\cellcolor[rgb]{0.933,0.933,1}}0.00 & {\cellcolor[rgb]{0.933,0.933,1}}0.00 & {\cellcolor[rgb]{0.933,0.933,1}}0.00 & {\cellcolor[rgb]{1,1,0.878}}\textbf{0.00}           & {\cellcolor[rgb]{1,1,0.878}}\textbf{0.00}  & {\cellcolor[rgb]{1,1,0.878}}\textbf{0.00}           &                            &                            &                            & \multirow{-2}{*}{{\cellcolor[rgb]{1,1,0.878}}\textbf{\textbf{3.16}}} & \multirow{-2}{*}{{\cellcolor[rgb]{1,1,0.878}}\textbf{1.65}} & \multirow{-2}{*}{{\cellcolor[rgb]{1,1,0.878}}\textbf{\textbf{1.68}}}  \\ 
\hhline{~-----------------}
                                       & \multirow{2}{*}{40\%}         & $AD_r\uparrow$                                                          & 99.99                                 & 99.99                                 & 99.99                                 & 99.99                                & 99.99                                & 99.99                                & {\cellcolor[rgb]{1,1,0.878}}\textbf{\textbf{96.98}} & {\cellcolor[rgb]{1,1,0.878}}\textbf{99.44} & {\cellcolor[rgb]{1,1,0.878}}\textbf{\textbf{99.49}} & \multirow{2}{*}{355.32}    & \multirow{2}{*}{247.64}    & \multirow{2}{*}{128.43}    & {\cellcolor[rgb]{1,1,0.878}}                                         & {\cellcolor[rgb]{1,1,0.878}}                                & {\cellcolor[rgb]{1,1,0.878}}                                          \\
                                       &                               & \multicolumn{1}{l|}{{\cellcolor[rgb]{0.933,0.933,1}}${AD}_f\downarrow$} & {\cellcolor[rgb]{0.933,0.933,1}}99.99 & {\cellcolor[rgb]{0.933,0.933,1}}99.99 & {\cellcolor[rgb]{0.933,0.933,1}}99.99 & {\cellcolor[rgb]{0.933,0.933,1}}0.00 & {\cellcolor[rgb]{0.933,0.933,1}}0.00 & {\cellcolor[rgb]{0.933,0.933,1}}0.00 & {\cellcolor[rgb]{1,1,0.878}}\textbf{0.00}           & {\cellcolor[rgb]{1,1,0.878}}\textbf{0.00}  & {\cellcolor[rgb]{1,1,0.878}}\textbf{0.00}           &                            &                            &                            & \multirow{-2}{*}{{\cellcolor[rgb]{1,1,0.878}}\textbf{\textbf{2.97}}} & \multirow{-2}{*}{{\cellcolor[rgb]{1,1,0.878}}\textbf{1.46}} & \multirow{-2}{*}{{\cellcolor[rgb]{1,1,0.878}}\textbf{\textbf{1.59}}}  \\ 
\hhline{~-----------------}
                                       & \multirow{2}{*}{20\%}         & $AD_r\uparrow$                                                          & 99.99                                 & 99.99                                 & 99.99                                 & 99.96                                & 99.99                                & 99.99                                & {\cellcolor[rgb]{1,1,0.878}}\textbf{\textbf{94.43}} & {\cellcolor[rgb]{1,1,0.878}}\textbf{98.86} & {\cellcolor[rgb]{1,1,0.878}}\textbf{\textbf{99.38}} & \multirow{2}{*}{354.83}    & \multirow{2}{*}{246.43}    & \multirow{2}{*}{129.35}    & {\cellcolor[rgb]{1,1,0.878}}                                         & {\cellcolor[rgb]{1,1,0.878}}                                & {\cellcolor[rgb]{1,1,0.878}}                                          \\
                                       &                               & \multicolumn{1}{l|}{{\cellcolor[rgb]{0.933,0.933,1}}${AD}_f\downarrow$} & {\cellcolor[rgb]{0.933,0.933,1}}99.99 & {\cellcolor[rgb]{0.933,0.933,1}}99.99 & {\cellcolor[rgb]{0.933,0.933,1}}99.99 & {\cellcolor[rgb]{0.933,0.933,1}}0.00 & {\cellcolor[rgb]{0.933,0.933,1}}0.00 & {\cellcolor[rgb]{0.933,0.933,1}}0.00 & {\cellcolor[rgb]{1,1,0.878}}\textbf{0.00}           & {\cellcolor[rgb]{1,1,0.878}}\textbf{0.00}  & {\cellcolor[rgb]{1,1,0.878}}\textbf{0.00}           &                            &                            &                            & \multirow{-2}{*}{{\cellcolor[rgb]{1,1,0.878}}\textbf{\textbf{2.89}}} & \multirow{-2}{*}{{\cellcolor[rgb]{1,1,0.878}}\textbf{1.54}} & \multirow{-2}{*}{{\cellcolor[rgb]{1,1,0.878}}\textbf{\textbf{1.38}}}  \\
\hline
\end{tabular}}
\vspace{-1em}
\end{table*}

Further observation reveals that the number of parameters in the last layer of DenseNet161 is 4.3 times that of ResNet18, and the number of parameters in the last layer of AlexNet is 7.9 times that of ResNet18. These results indicate that as the number of model parameters and the classes of unlearned labels increases, the $AD_r$ performance may decline. 

\vspace{-0.4em}
\begin{center}
\fbox{%
\begin{minipage}{0.96\linewidth}
\textbf{Takeaway--faster and better.}
FIUn outperforms others in unlearning speed by 99\%, averaging under one second per model, while effectively removing targeted labels and preserving retained accuracy, matching the effectiveness of the re-training method, which is considered the most effective.
\end{minipage}
}
\end{center}
\vspace{-1em}

\begin{figure*}[h]
 \centering
  \subfigure[FL, TinyImageNet, ResNet18]{
         \centering
         \includegraphics[width=0.22\textwidth]{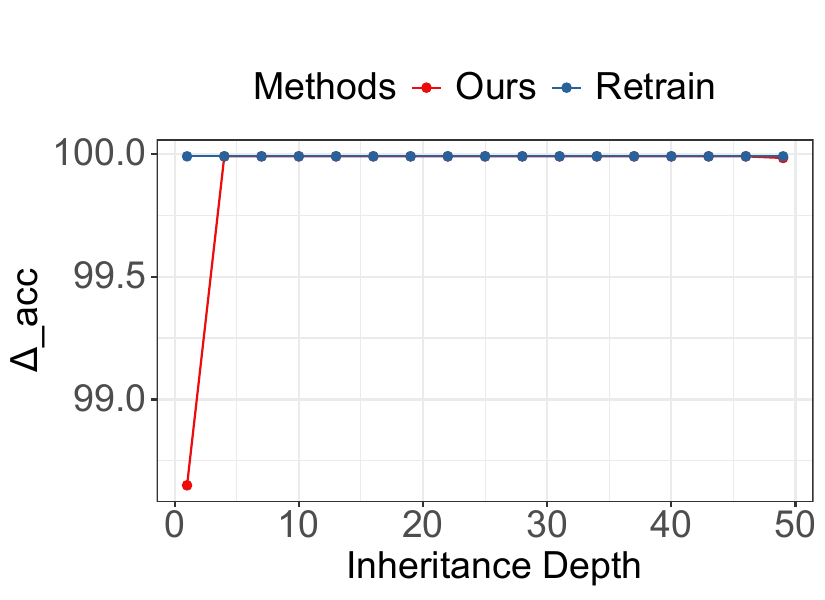}
         \label{fig:FLImageRes_2}
     }
     \subfigure[FL, CIFAR100, ResNet18]{
         \centering\includegraphics[width=0.22\textwidth]{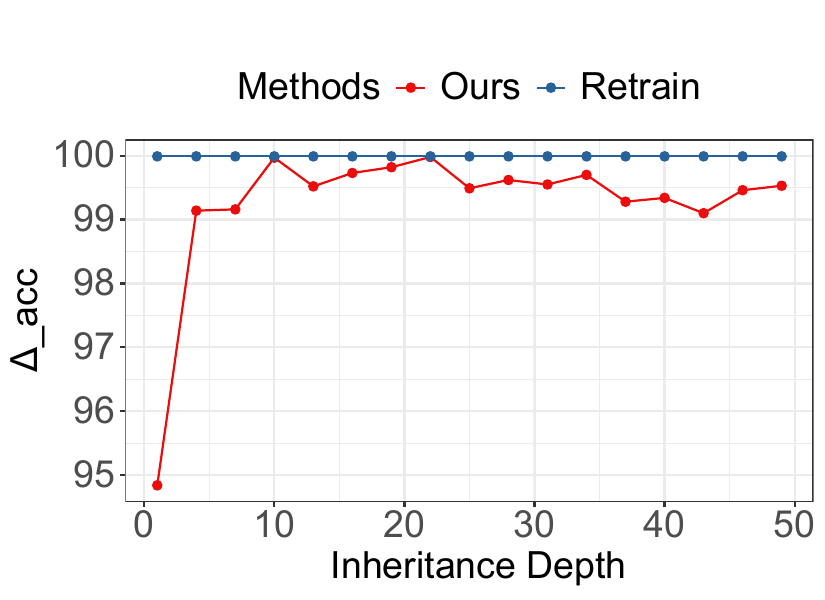}
         \label{fig:flcifarres_2}
     }
       \subfigure[FL, TinyImageNet, DenseNet161]{
         \centering
         \includegraphics[width=0.22\textwidth]{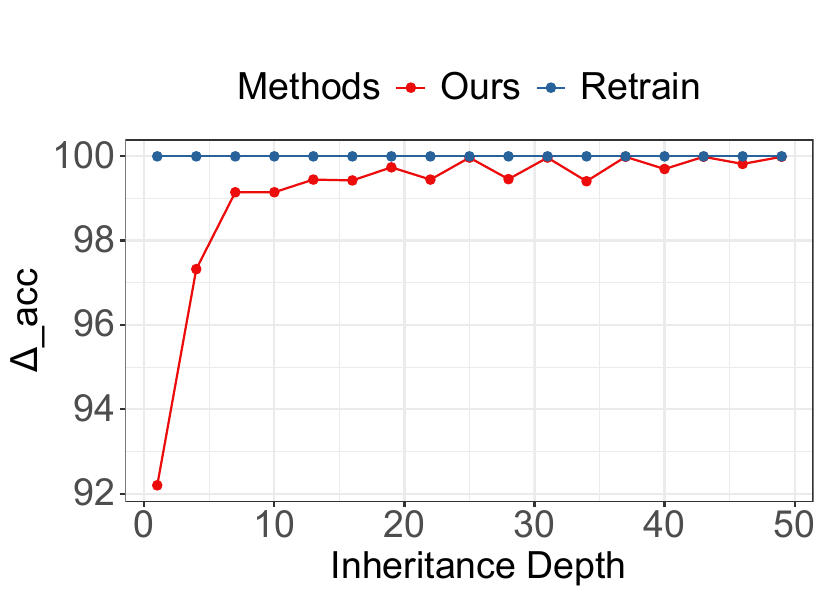}
         \label{fig:FLImageDense}
     }
    \subfigure[FL, CIFAR100, DenseNet161]{
        \centering
        \includegraphics[width=0.22\textwidth]{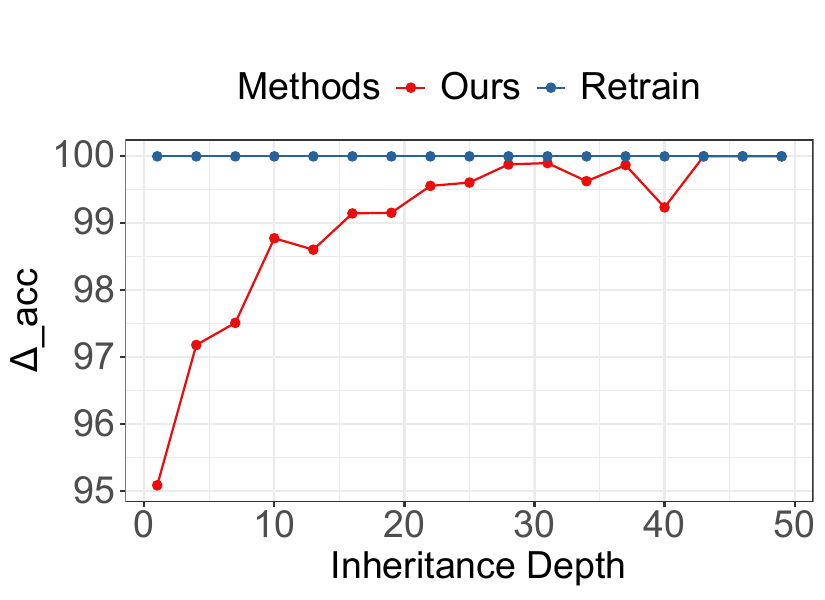}
        
        \label{fig:FLCifarDense}
    }
 \subfigure[IL, TinyImageNet, ResNet18]{
         \centering
         \includegraphics[width=0.22\textwidth]{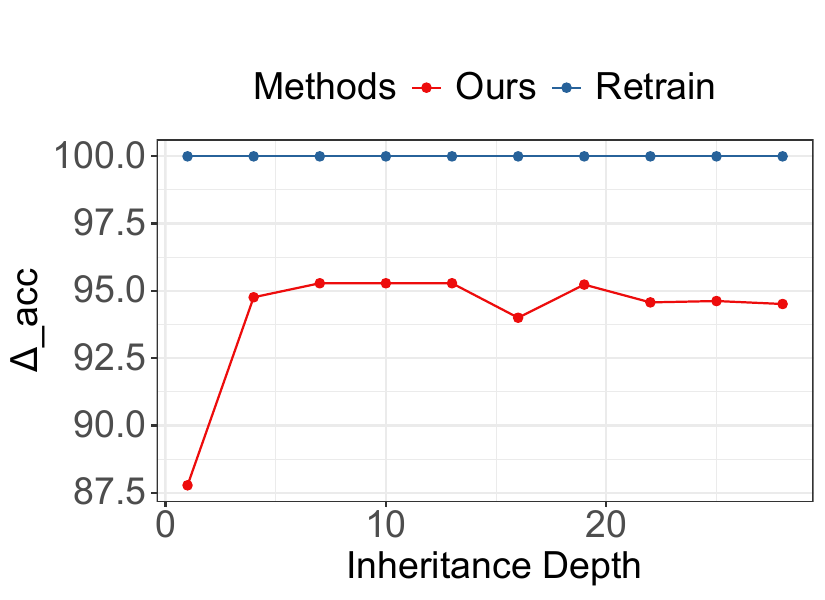}
         \label{fig:incImageRes}
     }
     \subfigure[IL, CIFAR100, ResNet18]{
         \centering\includegraphics[width=0.22\textwidth]{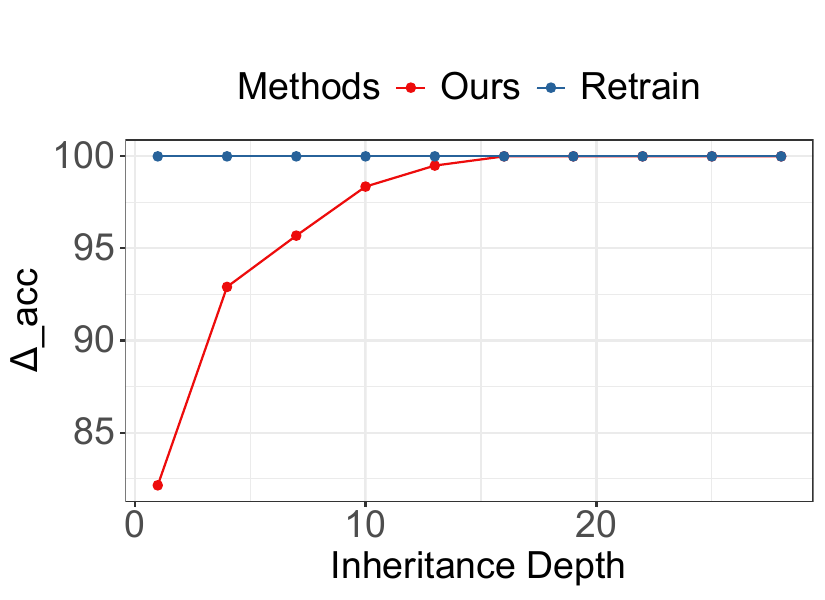}
         \label{fig:inccifarres}
     }
       \subfigure[IL, TinyImageNet, DenseNet161]{
         \centering
         \includegraphics[width=0.22\textwidth]{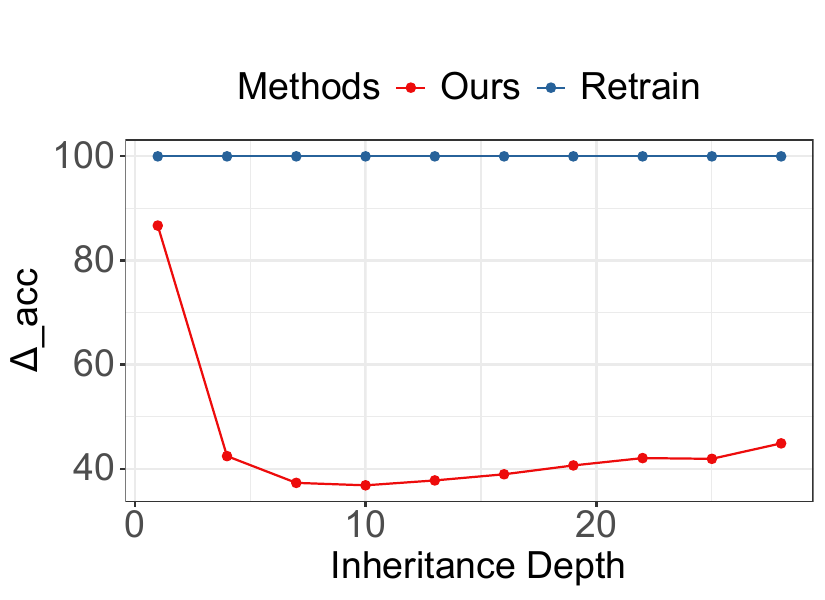}
         \label{fig:incImageDense}
     }
    \subfigure[IL, CIFAR100, DenseNet161]{
        \centering
        \includegraphics[width=0.22\textwidth]{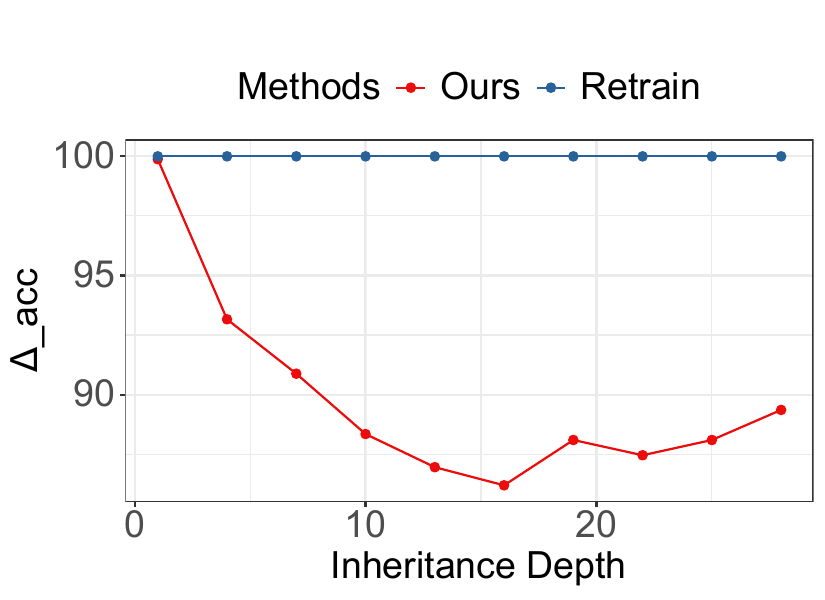}
        \label{fig:incCifarDense}
    }

  \caption{Inheritance depth analysis \#$C_f=1$.}
  \label{fig:Reference_deepth}
  \vspace{-1em}
\end{figure*}

\begin{figure*}[!ht]
 \centering
  \subfigure[FL, TinyImageNet, ResNet18]{
         \centering
         \includegraphics[width=0.22\textwidth]{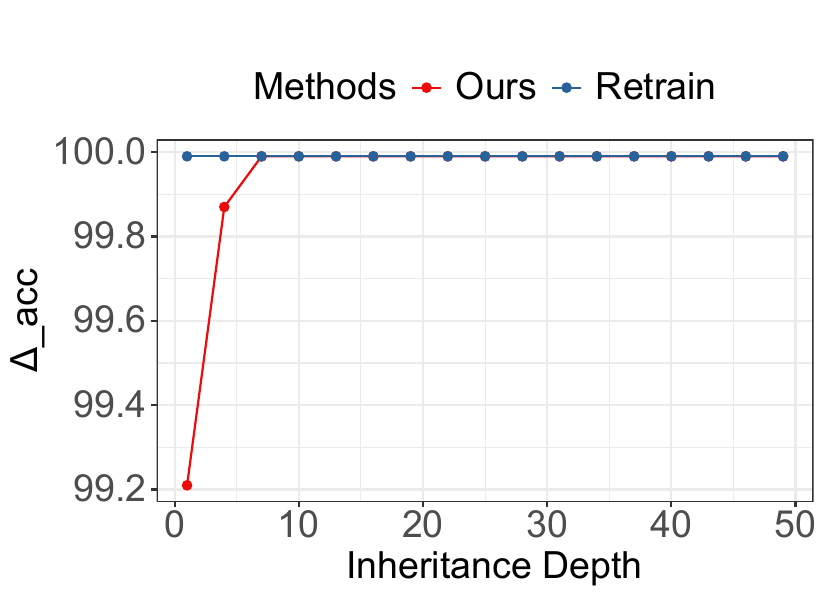}
         \label{fig:FLImageRes-4}
     }
     \subfigure[FL, CIFAR100, ResNet18]{
         \centering\includegraphics[width=0.22\textwidth]{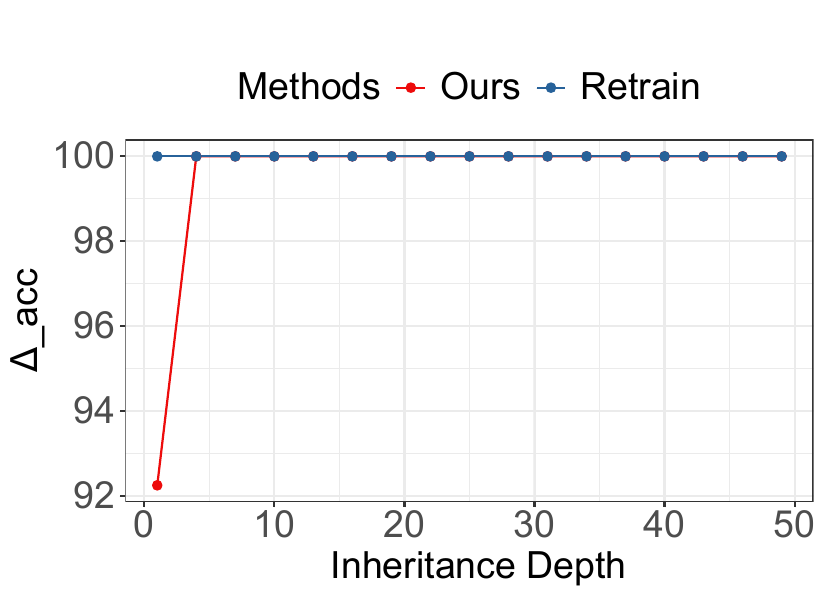}
         \label{fig:flcifarres_4}
     }
       \subfigure[FL, TinyImageNet, DenseNet161]{
         \centering
         \includegraphics[width=0.22\textwidth]{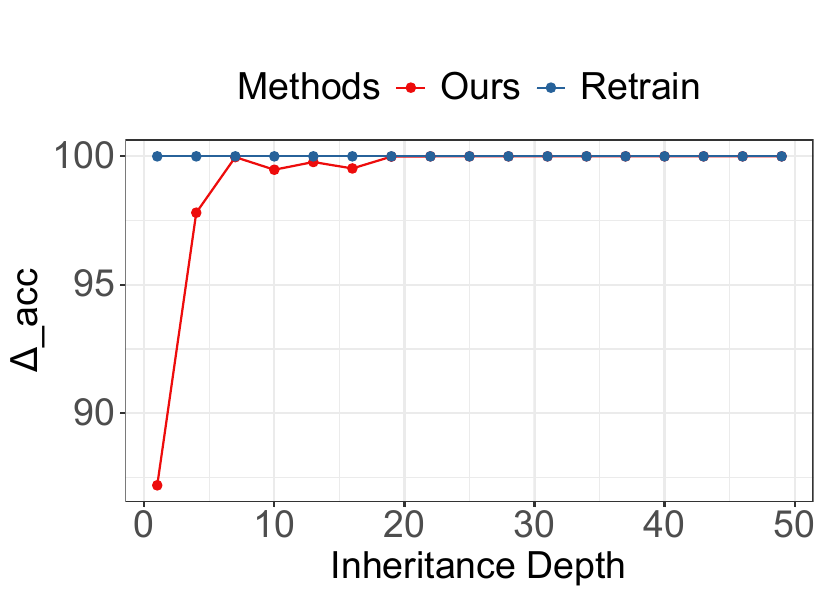}
         \label{fig:FLImageDense_4}
     }
    \subfigure[FL, CIFAR100, DenseNet161]{
        \centering
        \includegraphics[width=0.22\textwidth]{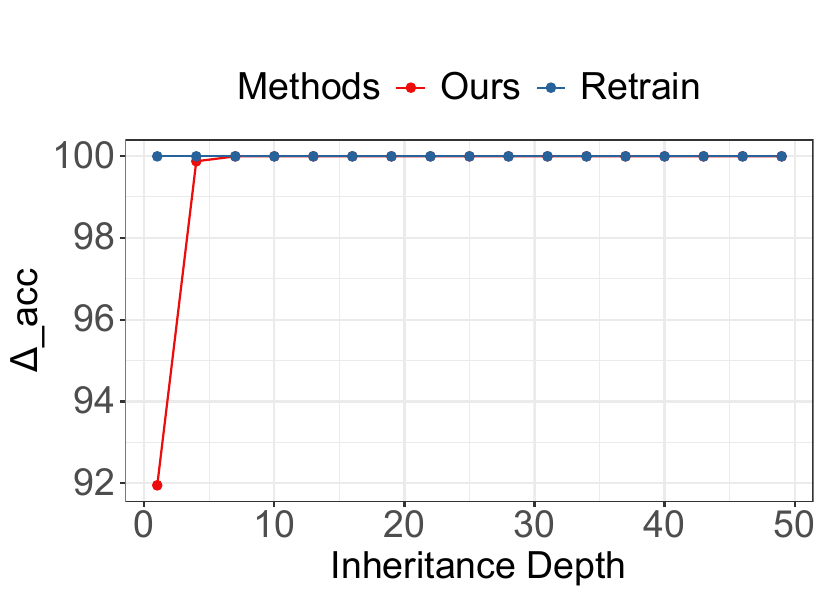}
      
        \label{fig:FLCifarDense_4}
    }
    \subfigure[IL, TinyImageNet, ResNet18]{
    \centering
        \includegraphics[width=0.22\textwidth]{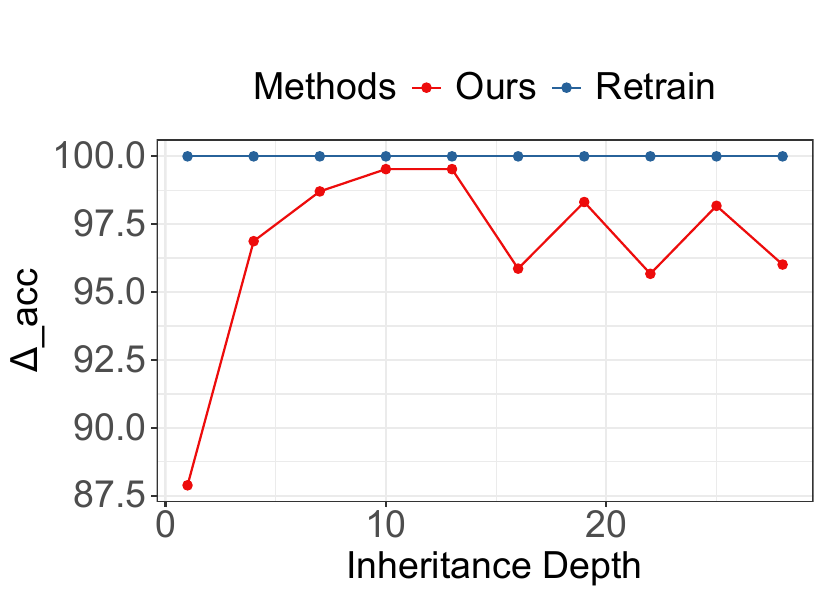}
      
        \label{fig:ILTinyRes_4}
    }
    \subfigure[IL, CIFAR100, ResNet18]{
        \centering
        \includegraphics[width=0.22\textwidth]{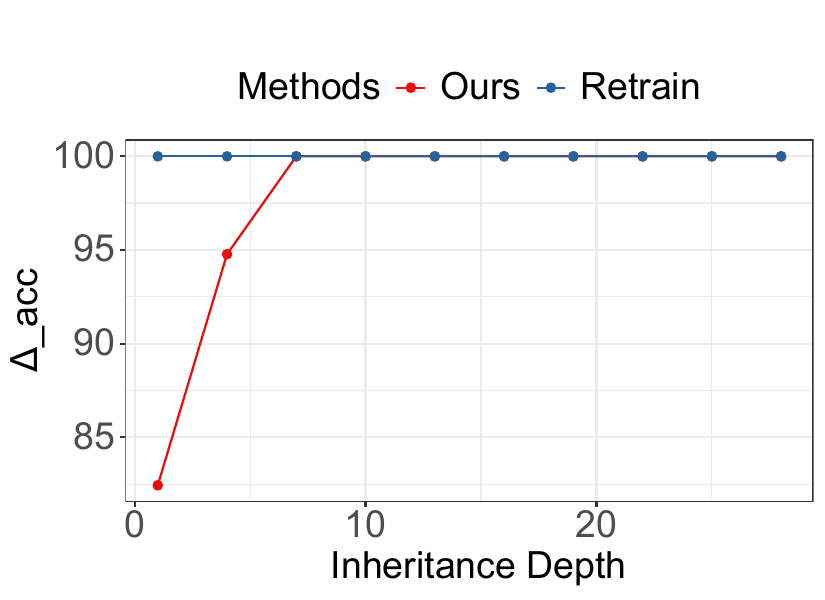}
      
        \label{fig:ILCifarRes_4}
    }
   \subfigure[IL, TinyImageNet, DenseNet161]{
        \centering
        \includegraphics[width=0.22\textwidth]{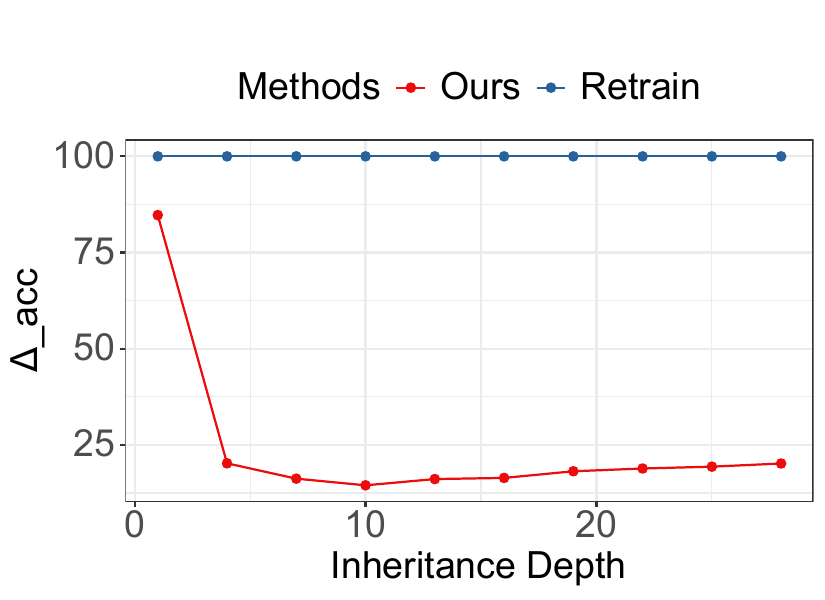}
      
        \label{fig:ILTinyDense_4}
    }
  \subfigure[IL, CIFAR100, DenseNet161]{
        \centering
        \includegraphics[width=0.22\textwidth]{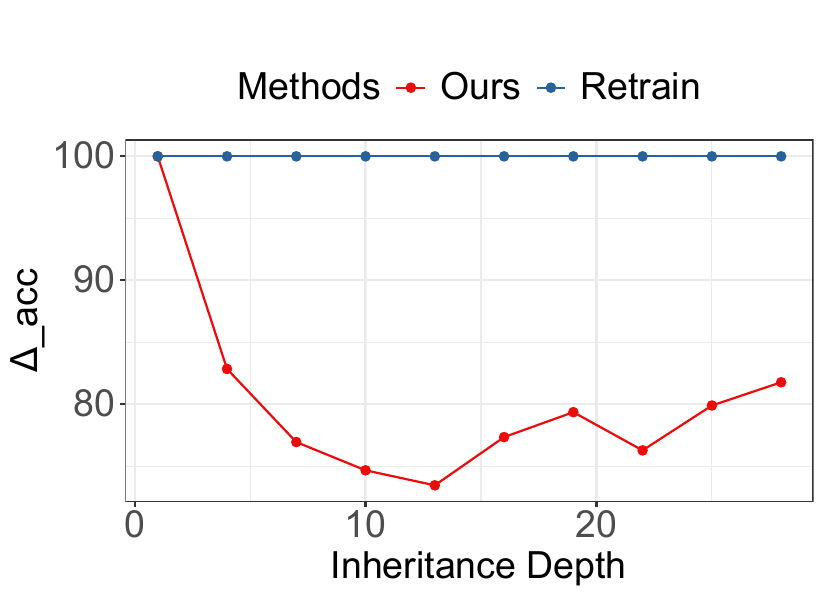}
      
        \label{fig:ILCifarDense_4}
    }

  \caption{Inheritance depth analysis \#$C_f=4$.}
  \label{fig:Reference_deepth1_1}
  \vspace{-1em}
\end{figure*}

\begin{figure*}[!ht]
 \centering
  \subfigure[FL, AlexNet, CIFAR100]{
         \centering
         \includegraphics[width=0.22\textwidth]{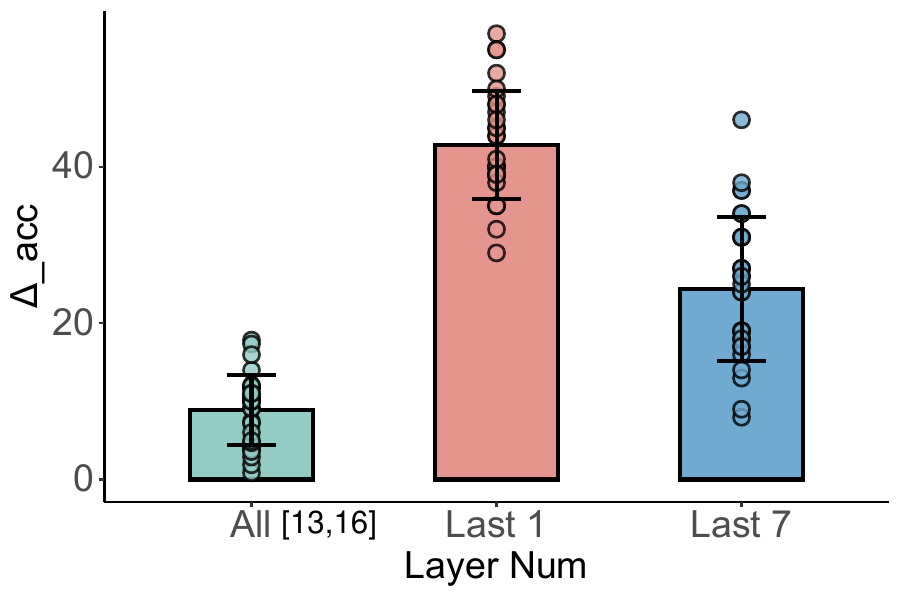}
         \label{fig:LayerFLAlexCifar}
     }
     \subfigure[FL, CIFAR100, ResNet18]{
         \centering\includegraphics[width=0.22\textwidth]{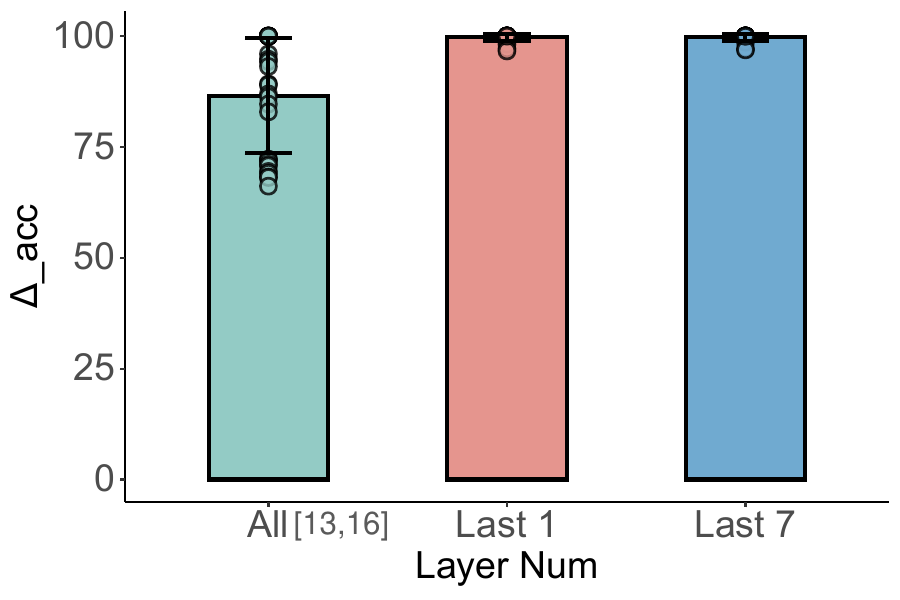}
         \label{fig:LayerFLResnet}
     }
       \subfigure[IL, CIFAR100, ResNet18]{
         \centering
         \includegraphics[width=0.22\textwidth]{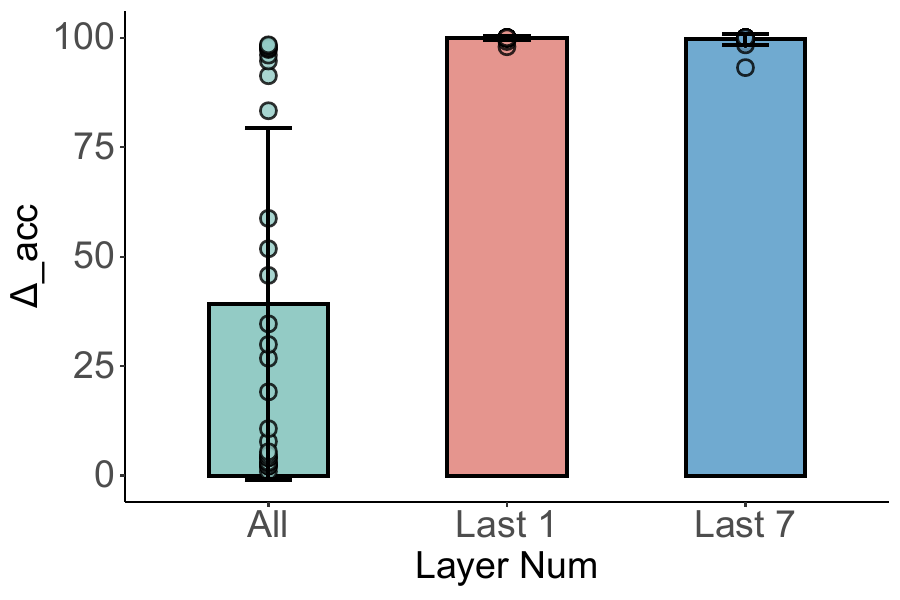}
         \label{fig:LayerILResnet}
     }
    \subfigure[IL,CIFAR100, AlexNet]{
        \centering
        \includegraphics[width=0.22\textwidth]{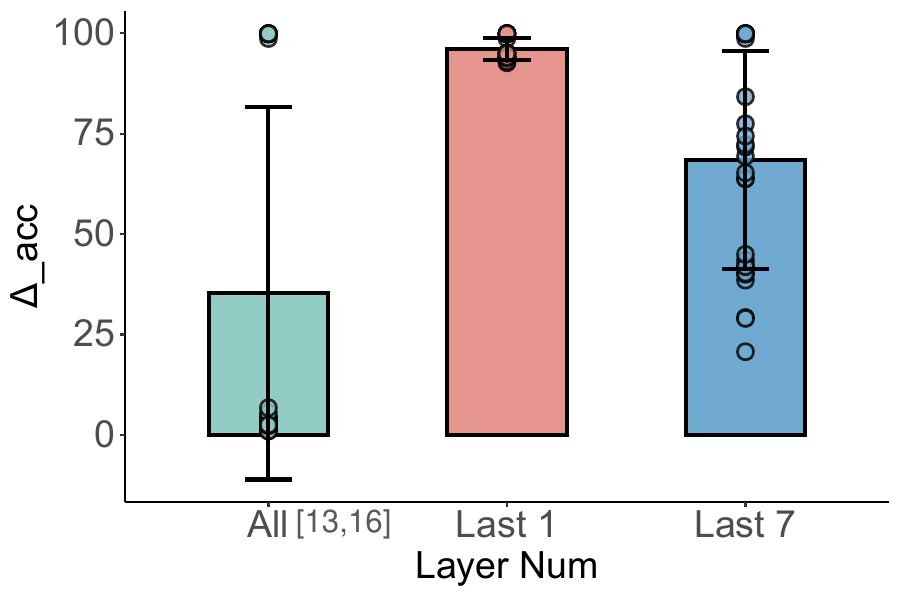}
     
        \label{fig:LayerILCifaralex}
    }
 \subfigure[FL, TinyImageNet, DenseNet161]{
         \centering
         \includegraphics[width=0.22\textwidth]{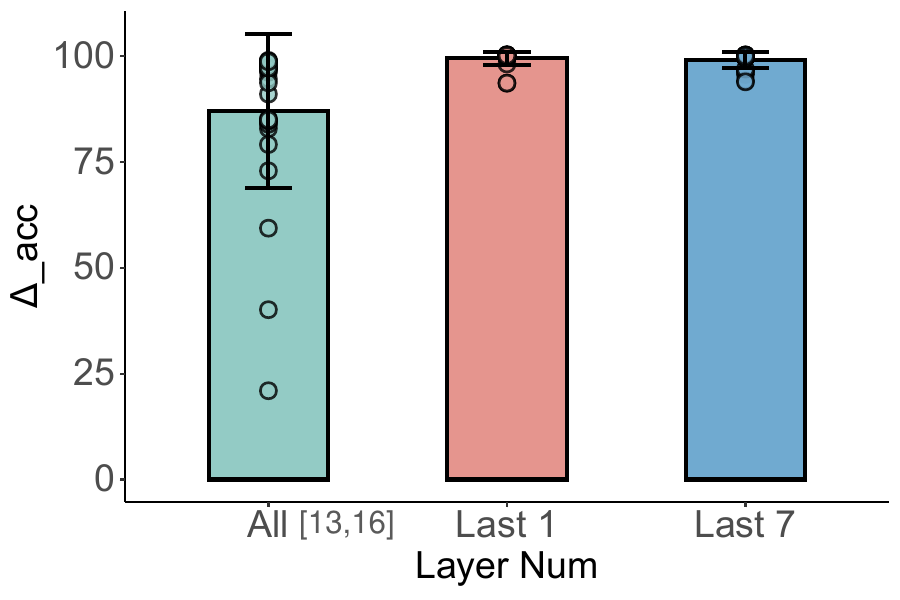}
         \label{fig:layerFLDense}
     }
     \subfigure[FL, TinyImageNet, ResNet18]{
         \centering\includegraphics[width=0.22\textwidth]{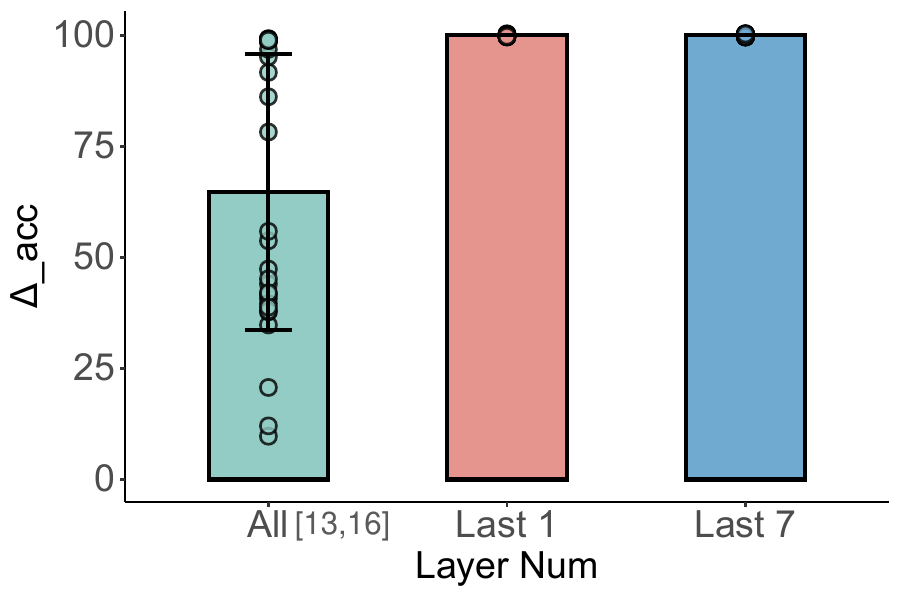}
         \label{fig:LayerFLImage}
     }
       \subfigure[IL, TinyImageNet, ResNet18]{
         \centering
         \includegraphics[width=0.22\textwidth]{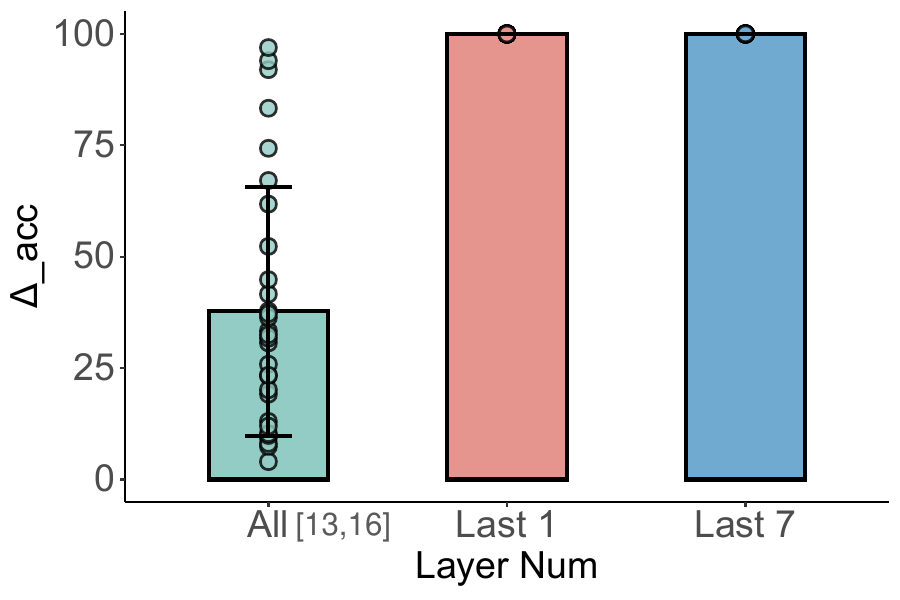}
        \label{fig:layerILImageRes}
     }
    \subfigure[IL, TinyImageNet, DenseNet161]{
        \centering
        \includegraphics[width=0.22\textwidth]{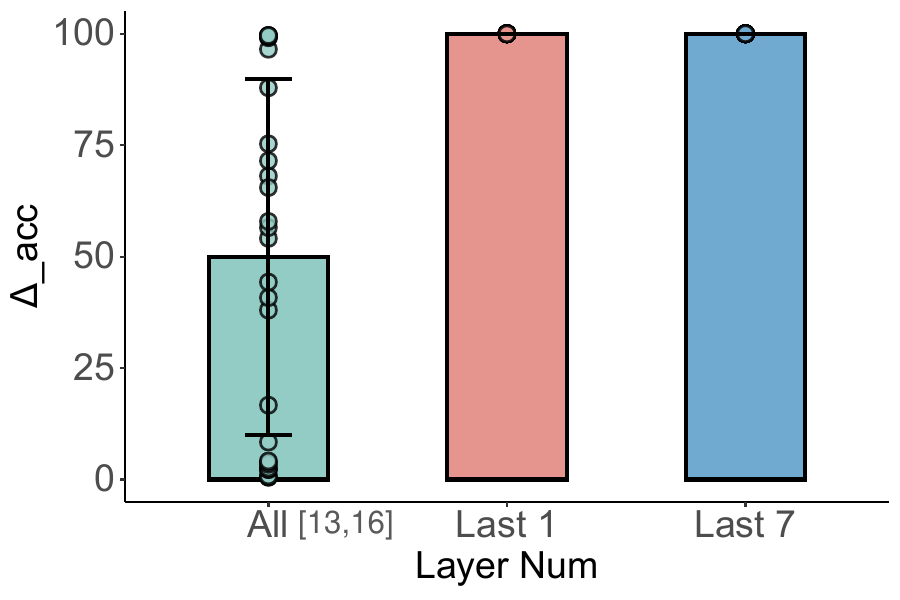}
       \label{fig:LayerILImageDense}
    }
  \caption{Unlearning layer analysis \#$C_f$=15 for all layers \cite{foster2024fast,golatkar2020eternal}, last 1 layer and last 7 layers.}
  \label{fig:unlearning_layer}
\vspace{-1em}
\end{figure*}

\begin{table*}
\centering
\setlength{\extrarowheight}{0pt}
\addtolength{\extrarowheight}{\aboverulesep}
\addtolength{\extrarowheight}{\belowrulesep}
\setlength{\aboverulesep}{0pt}
\setlength{\belowrulesep}{0pt}
\caption{Large-scale Federated Unlearning Performance of Language and Graph Models on Yahoo! Answers}
\label{table:Yahoo}
\resizebox{0.8\textwidth}{!}{
\begin{tabular}{ccc|ccc|ccc|ccc|cccccc} 
\toprule
\multirow{3}{*}{\textbf{Model}}            & \multirow{3}{*}{\#$C_f$} & \multirow{3}{*}{\textbf{Metrics}}                & \multicolumn{3}{c|}{\multirow{2}{*}{Original~(\%)}}                                                                   & \multicolumn{3}{c|}{\multirow{2}{*}{Re-training~(\%)}}                                                             & \multicolumn{3}{c|}{\multirow{2}{*}{FIUn~(\%)}}                                                                                                                                                                                                                                                                                               & \multicolumn{6}{c}{\textbf{Cumulative Unlearning Time (s)}}                                                                                                                                                                                                                           \\ 
\cmidrule{13-18}
                                           &                          &                                                  & \multicolumn{3}{c|}{}                                                                                                 & \multicolumn{3}{c|}{}                                                                                              & \multicolumn{3}{c|}{}                                                                                                                                                                                                                                                                                                                         & \multicolumn{3}{c|}{Re-training}                                               & \multicolumn{3}{c}{FIUn}                                                                                                                                                                             \\ 
\cmidrule{4-18}
                                           &                          &                                                  & $w_g$                                 & $w_a$                                 & $w_b$                                 & $w_g$                                & $w_a$                                & $w_b$                                & $w_g$                                                                                                         & $w_a$                                                                                                         & $w_b$                                                                                                         & $w_g$                   & $w_a$                   & \multicolumn{1}{c|}{$w_b$} & $w_g$                                                           & $w_a$                                                           & $w_b$                                                            \\ 
\midrule
\multirow{6}{*}{\rotcell{Bert-Base-Cased}} & \multirow{2}{*}{1}       & $AD_r\uparrow$                                   & 99.99                                 & 99.99                                 & 99.99                                 & 99.99                                & 99.99                                & 99.99                                & {\cellcolor[rgb]{1,0.992,0.859}}\textbf{99.99}                                                                & {\cellcolor[rgb]{1,0.992,0.859}}\textbf{\textbf{99.99}}                                                       & {\cellcolor[rgb]{1,0.992,0.859}}\textbf{\textbf{99.99}}                                                       & \multirow{2}{*}{105.63} & \multirow{2}{*}{215.14} & \multirow{2}{*}{213.75}    & {\cellcolor[rgb]{1,0.992,0.859}}                                & {\cellcolor[rgb]{1,0.992,0.859}}                                & {\cellcolor[rgb]{1,0.992,0.859}}                                 \\
                                           &                          & {\cellcolor[rgb]{0.922,0.922,1}}$AD_f\downarrow$ & {\cellcolor[rgb]{0.922,0.922,1}}99.99 & {\cellcolor[rgb]{0.922,0.922,1}}99.99 & {\cellcolor[rgb]{0.922,0.922,1}}99.99 & {\cellcolor[rgb]{0.922,0.922,1}}0.00 & {\cellcolor[rgb]{0.922,0.922,1}}0.00 & {\cellcolor[rgb]{0.922,0.922,1}}0.00 & {\cellcolor[rgb]{1,0.992,0.859}}\textbf{0.99}                                                                 & {\cellcolor[rgb]{1,0.992,0.859}}\textbf{0.00}                                                                 & {\cellcolor[rgb]{1,0.992,0.859}}\textbf{0.00}                                                                 &                         &                         &                            & \multirow{-2}{*}{{\cellcolor[rgb]{1,0.992,0.859}}\textbf{1.56}} & \multirow{-2}{*}{{\cellcolor[rgb]{1,0.992,0.859}}\textbf{2.84}} & \multirow{-2}{*}{{\cellcolor[rgb]{1,0.992,0.859}}\textbf{2.79}}  \\ 
\hhline{~-----------------}
                                           & \multirow{2}{*}{2}       & $AD_r\uparrow$~                                  & 99.99                                 & 99.99                                 & 99.99                                 & 99.99                                & 99.99                                & 99.99                                & {\cellcolor[rgb]{1,0.992,0.859}}\textbf{\textbf{99.99}}                                                       & {\cellcolor[rgb]{1,0.992,0.859}}\textbf{\textbf{99.99}}                                                       & {\cellcolor[rgb]{1,0.992,0.859}}\textbf{\textbf{99.99}}                                                       & \multirow{2}{*}{93.15}  & \multirow{2}{*}{186.43} & \multirow{2}{*}{190.32}    & {\cellcolor[rgb]{1,0.992,0.859}}                                & {\cellcolor[rgb]{1,0.992,0.859}}                                & {\cellcolor[rgb]{1,0.992,0.859}}                                 \\
                                           &                          & {\cellcolor[rgb]{0.922,0.922,1}}$AD_f\downarrow$ & {\cellcolor[rgb]{0.922,0.922,1}}99.99 & {\cellcolor[rgb]{0.922,0.922,1}}99.99 & {\cellcolor[rgb]{0.922,0.922,1}}99.99 & {\cellcolor[rgb]{0.922,0.922,1}}0.00 & {\cellcolor[rgb]{0.922,0.922,1}}0.00 & {\cellcolor[rgb]{0.922,0.922,1}}0.00 & {\cellcolor[rgb]{1,0.992,0.859}}\textbf{15.13}                                                                & {\cellcolor[rgb]{1,0.992,0.859}}\textbf{0.00}                                                                 & {\cellcolor[rgb]{1,0.992,0.859}}\textbf{10.21}                                                                &                         &                         &                            & \multirow{-2}{*}{{\cellcolor[rgb]{1,0.992,0.859}}\textbf{1.51}} & \multirow{-2}{*}{{\cellcolor[rgb]{1,0.992,0.859}}\textbf{2.75}} & \multirow{-2}{*}{{\cellcolor[rgb]{1,0.992,0.859}}\textbf{2.81}}  \\ 
\hhline{~-----------------}
                                           & \multirow{2}{*}{4}       & $AD_r\uparrow$                                   & 99.99                                 & 99.99                                 & 99.99                                 & 99.99                                & 99.99                                & 99.99                                & {\cellcolor[rgb]{1,0.992,0.859}}\textbf{\textbf{\textbf{\textbf{\textbf{\textbf{\textbf{\textbf{99.99}}}}}}}} & {\cellcolor[rgb]{1,0.992,0.859}}\textbf{\textbf{\textbf{\textbf{\textbf{\textbf{\textbf{\textbf{99.99}}}}}}}} & {\cellcolor[rgb]{1,0.992,0.859}}\textbf{\textbf{\textbf{\textbf{\textbf{\textbf{\textbf{\textbf{99.99}}}}}}}} & \multirow{2}{*}{70.31}  & \multirow{2}{*}{143.53} & \multirow{2}{*}{145.17}    & {\cellcolor[rgb]{1,0.992,0.859}}                                & {\cellcolor[rgb]{1,0.992,0.859}}                                & {\cellcolor[rgb]{1,0.992,0.859}}                                 \\
                                           &                          & {\cellcolor[rgb]{0.922,0.922,1}}$AD_f\downarrow$ & {\cellcolor[rgb]{0.922,0.922,1}}99.99 & {\cellcolor[rgb]{0.922,0.922,1}}99.99 & {\cellcolor[rgb]{0.922,0.922,1}}99.99 & {\cellcolor[rgb]{0.922,0.922,1}}0.00 & {\cellcolor[rgb]{0.922,0.922,1}}0.00 & {\cellcolor[rgb]{0.922,0.922,1}}0.00 & {\cellcolor[rgb]{1,0.992,0.859}}\textbf{26.73}                                                                & {\cellcolor[rgb]{1,0.992,0.859}}\textbf{0.00}                                                                 & {\cellcolor[rgb]{1,0.992,0.859}}\textbf{6.41}                                                                 &                         &                         &                            & \multirow{-2}{*}{{\cellcolor[rgb]{1,0.992,0.859}}\textbf{1.90}} & \multirow{-2}{*}{{\cellcolor[rgb]{1,0.992,0.859}}\textbf{2.91}} & \multirow{-2}{*}{{\cellcolor[rgb]{1,0.992,0.859}}\textbf{3.14}}  \\ 
\midrule
\multirow{6}{*}{\rotcell{GCNN}}            & \multirow{2}{*}{1}       & $AD_r\uparrow$                                   & 99.56                                 & 99.93                                 & 96.43                                 & 99.69                                & 99.99                                & 97.64                                & {\cellcolor[rgb]{1,0.992,0.859}}\textbf{91.54}                                                                & {\cellcolor[rgb]{1,0.992,0.859}}\textbf{68.69}                                                                & {\cellcolor[rgb]{1,0.992,0.859}}\textbf{95.67}                                                                & \multirow{2}{*}{33.95}  & \multirow{2}{*}{67.31}  & \multirow{2}{*}{101.64}    & {\cellcolor[rgb]{1,0.992,0.859}}                                & {\cellcolor[rgb]{1,0.992,0.859}}                                & {\cellcolor[rgb]{1,0.992,0.859}}                                 \\
                                           &                          & {\cellcolor[rgb]{0.922,0.922,1}}$AD_f\downarrow$ & {\cellcolor[rgb]{0.922,0.922,1}}98.81 & {\cellcolor[rgb]{0.922,0.922,1}}99.99 & {\cellcolor[rgb]{0.922,0.922,1}}94.64 & {\cellcolor[rgb]{0.922,0.922,1}}0.00 & {\cellcolor[rgb]{0.922,0.922,1}}0.00 & {\cellcolor[rgb]{0.922,0.922,1}}0.00 & {\cellcolor[rgb]{1,0.992,0.859}}\textbf{0.00}                                                                 & {\cellcolor[rgb]{1,0.992,0.859}}\textbf{0.00}                                                                 & {\cellcolor[rgb]{1,0.992,0.859}}\textbf{9.52}                                                                 &                         &                         &                            & \multirow{-2}{*}{{\cellcolor[rgb]{1,0.992,0.859}}\textbf{0.42}} & \multirow{-2}{*}{{\cellcolor[rgb]{1,0.992,0.859}}\textbf{0.77}} & \multirow{-2}{*}{{\cellcolor[rgb]{1,0.992,0.859}}\textbf{0.76}}  \\ 
\hhline{~-----------------}
                                           & \multirow{2}{*}{2}       & $AD_r\uparrow$                                   & 99.56                                 & 99.93                                 & 96.43                                 & 99.99                                & 99.99                                & 98.31                                & {\cellcolor[rgb]{1,0.992,0.859}}\textbf{63.89}                                                                & {\cellcolor[rgb]{1,0.992,0.859}}\textbf{64.29}                                                                & {\cellcolor[rgb]{1,0.992,0.859}}\textbf{71.64}                                                                & \multirow{2}{*}{28.60}  & \multirow{2}{*}{57.14}  & \multirow{2}{*}{86.41}     & {\cellcolor[rgb]{1,0.992,0.859}}                                & {\cellcolor[rgb]{1,0.992,0.859}}                                & {\cellcolor[rgb]{1,0.992,0.859}}                                 \\
                                           &                          & {\cellcolor[rgb]{0.922,0.922,1}}$AD_f\downarrow$ & {\cellcolor[rgb]{0.922,0.922,1}}99.11 & {\cellcolor[rgb]{0.922,0.922,1}}99.99 & {\cellcolor[rgb]{0.922,0.922,1}}95.60 & {\cellcolor[rgb]{0.922,0.922,1}}0.00 & {\cellcolor[rgb]{0.922,0.922,1}}0.00 & {\cellcolor[rgb]{0.922,0.922,1}}0.00 & {\cellcolor[rgb]{1,0.992,0.859}}\textbf{0.00}                                                                 & {\cellcolor[rgb]{1,0.992,0.859}}\textbf{0.00}                                                                 & {\cellcolor[rgb]{1,0.992,0.859}}\textbf{0.00}                                                                 &                         &                         &                            & \multirow{-2}{*}{{\cellcolor[rgb]{1,0.992,0.859}}\textbf{0.44}} & \multirow{-2}{*}{{\cellcolor[rgb]{1,0.992,0.859}}\textbf{0.87}} & \multirow{-2}{*}{{\cellcolor[rgb]{1,0.992,0.859}}\textbf{0.86}}  \\ 
\hhline{~-----------------}
                                           & \multirow{2}{*}{4}       & $AD_r\uparrow$                                   & 99.56                                 & 99.93                                 & 96.43                                 & 99.99                                & 99.99                                & 98.34                                & {\cellcolor[rgb]{1,0.992,0.859}}\textbf{93.17}                                                                & {\cellcolor[rgb]{1,0.992,0.859}}\textbf{63.07}                                                                & {\cellcolor[rgb]{1,0.992,0.859}}\textbf{97.41}                                                                & \multirow{2}{*}{21.63}  & \multirow{2}{*}{42.60}  & \multirow{2}{*}{64.31}     & {\cellcolor[rgb]{1,0.992,0.859}}                                & {\cellcolor[rgb]{1,0.992,0.859}}                                & {\cellcolor[rgb]{1,0.992,0.859}}                                 \\
                                           &                          & {\cellcolor[rgb]{0.922,0.922,1}}$AD_f\downarrow$ & {\cellcolor[rgb]{0.922,0.922,1}}99.07 & {\cellcolor[rgb]{0.922,0.922,1}}99.99 & {\cellcolor[rgb]{0.922,0.922,1}}95.67 & {\cellcolor[rgb]{0.922,0.922,1}}0.00 & {\cellcolor[rgb]{0.922,0.922,1}}0.00 & {\cellcolor[rgb]{0.922,0.922,1}}0.00 & {\cellcolor[rgb]{1,0.992,0.859}}\textbf{0.00}                                                                 & {\cellcolor[rgb]{1,0.992,0.859}}\textbf{0.00}                                                                 & {\cellcolor[rgb]{1,0.992,0.859}}\textbf{25.37}                                                                &                         &                         &                            & \multirow{-2}{*}{{\cellcolor[rgb]{1,0.992,0.859}}\textbf{0.34}} & \multirow{-2}{*}{{\cellcolor[rgb]{1,0.992,0.859}}\textbf{0.79}} & \multirow{-2}{*}{{\cellcolor[rgb]{1,0.992,0.859}}\textbf{0.84}}  \\
\bottomrule
\end{tabular}
}
\vspace{-1em}
\end{table*}

\begin{figure*}[!ht]
 \centering
  \subfigure[Throughput, FL, TinyImageNet, ResNet18]{
         \centering
         \includegraphics[width=0.22\textwidth]{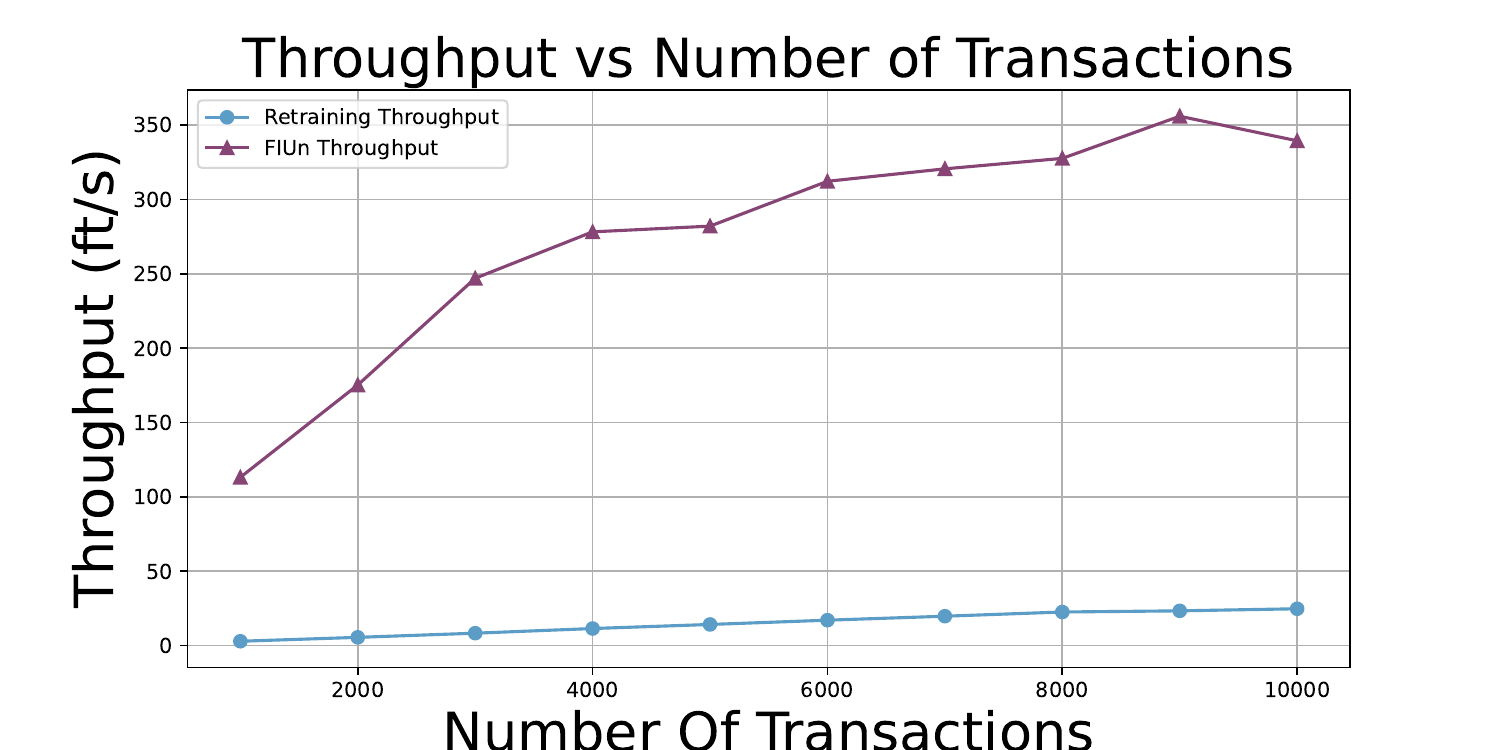}
         \label{fig:throuput}
     }
     \subfigure[Latency, FL, TinyImageNet, ResNet18]{
         \centering\includegraphics[width=0.22\textwidth]{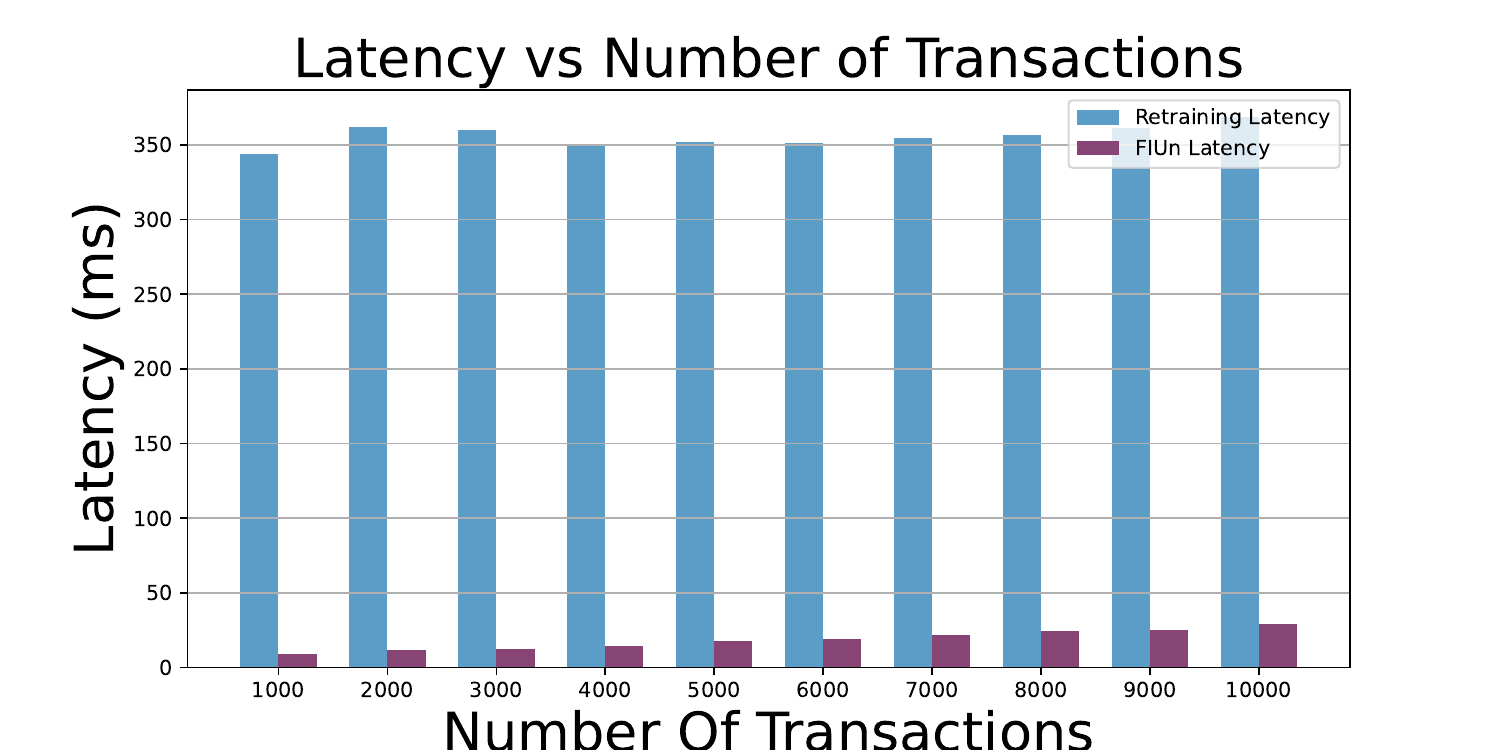}
         \label{fig:latency}
     }
       \subfigure[Throughput, FL, TinyImageNet, DenseNet161]{
         \centering
         \includegraphics[width=0.22\textwidth]{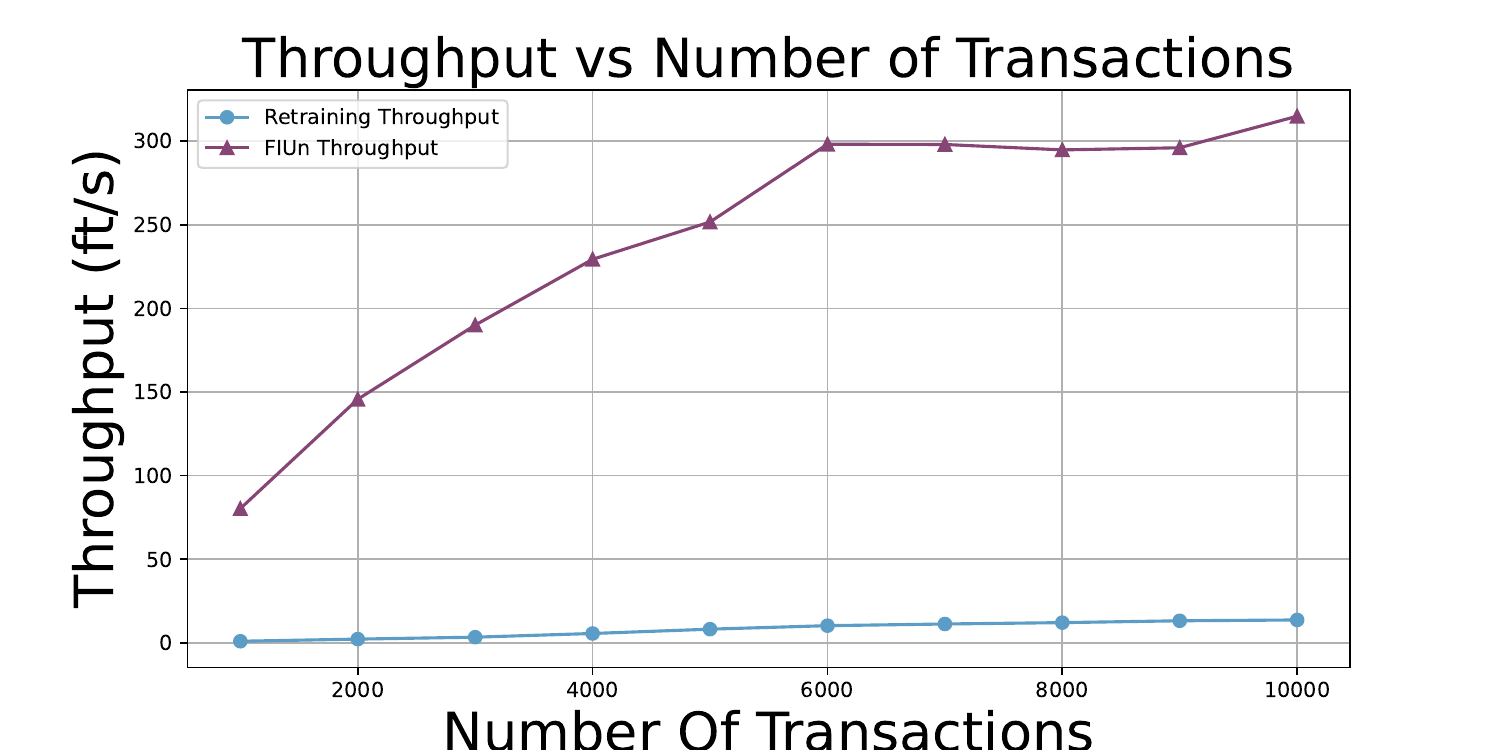}
         \label{fig:throughput_dense}
     }
    \subfigure[Latency, FL, TinyImageNet, DenseNet161]{
        \centering
        \includegraphics[width=0.22\textwidth]{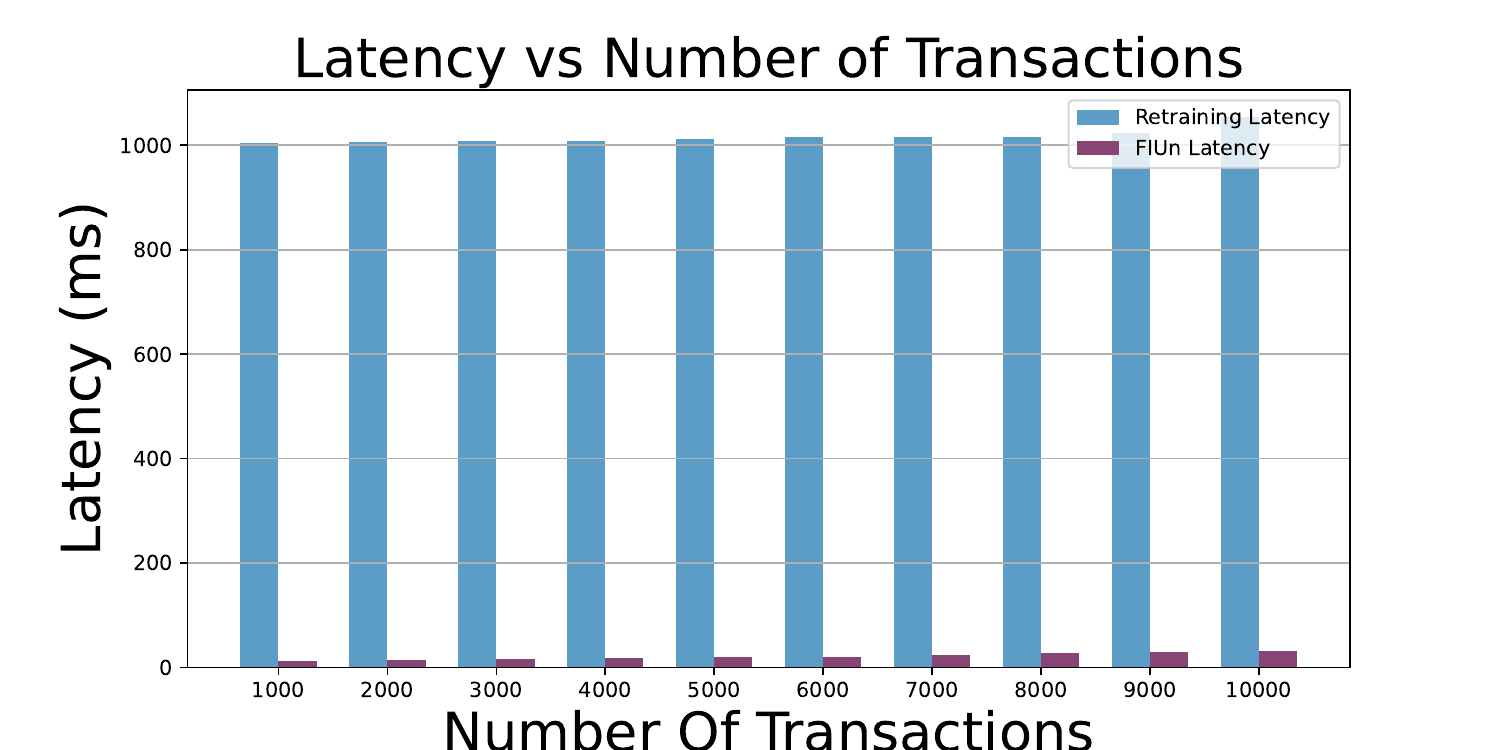}
      
        \label{fig:latency_dense}
    }
    \subfigure[Throughput, FL, Yahoo, Bert]{
    \centering
        \includegraphics[width=0.22\textwidth]{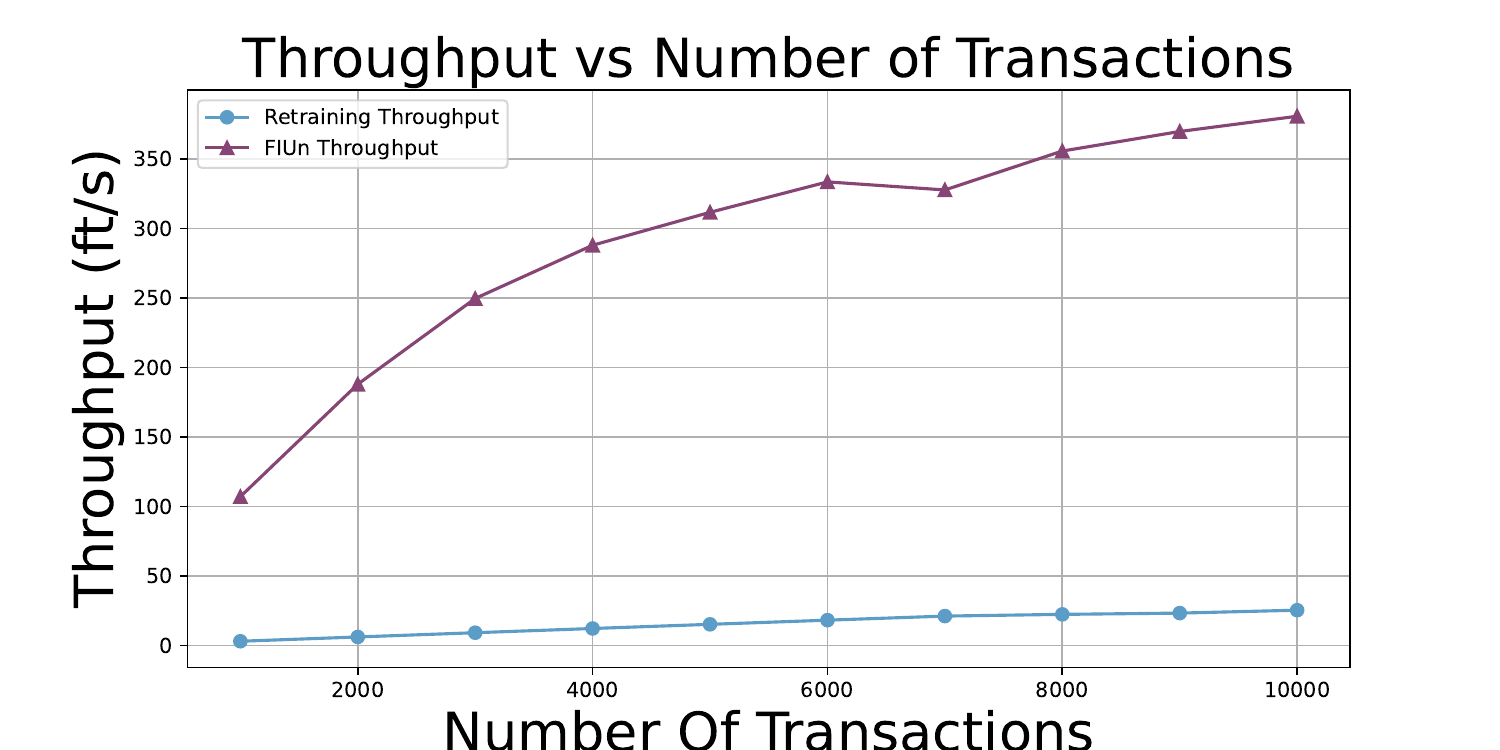}
      
        \label{fig:throughput_bert}
    }
    \subfigure[Latency, FL, Yahoo, Bert]{
        \centering
        \includegraphics[width=0.22\textwidth]{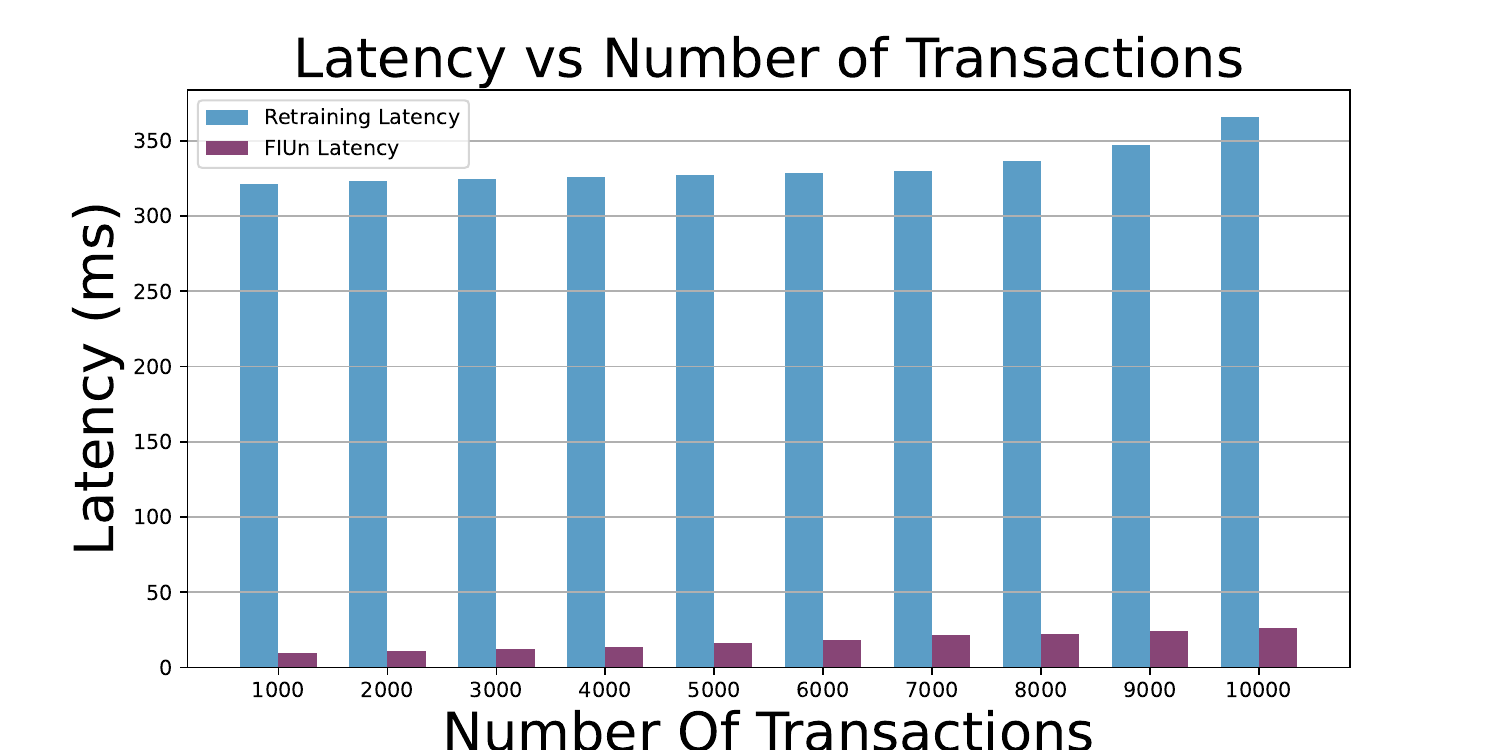}
      
        \label{fig:latency_bert}
    }
   \subfigure[Throughput, FL, Yahoo, GCNN]{
        \centering
        \includegraphics[width=0.22\textwidth]{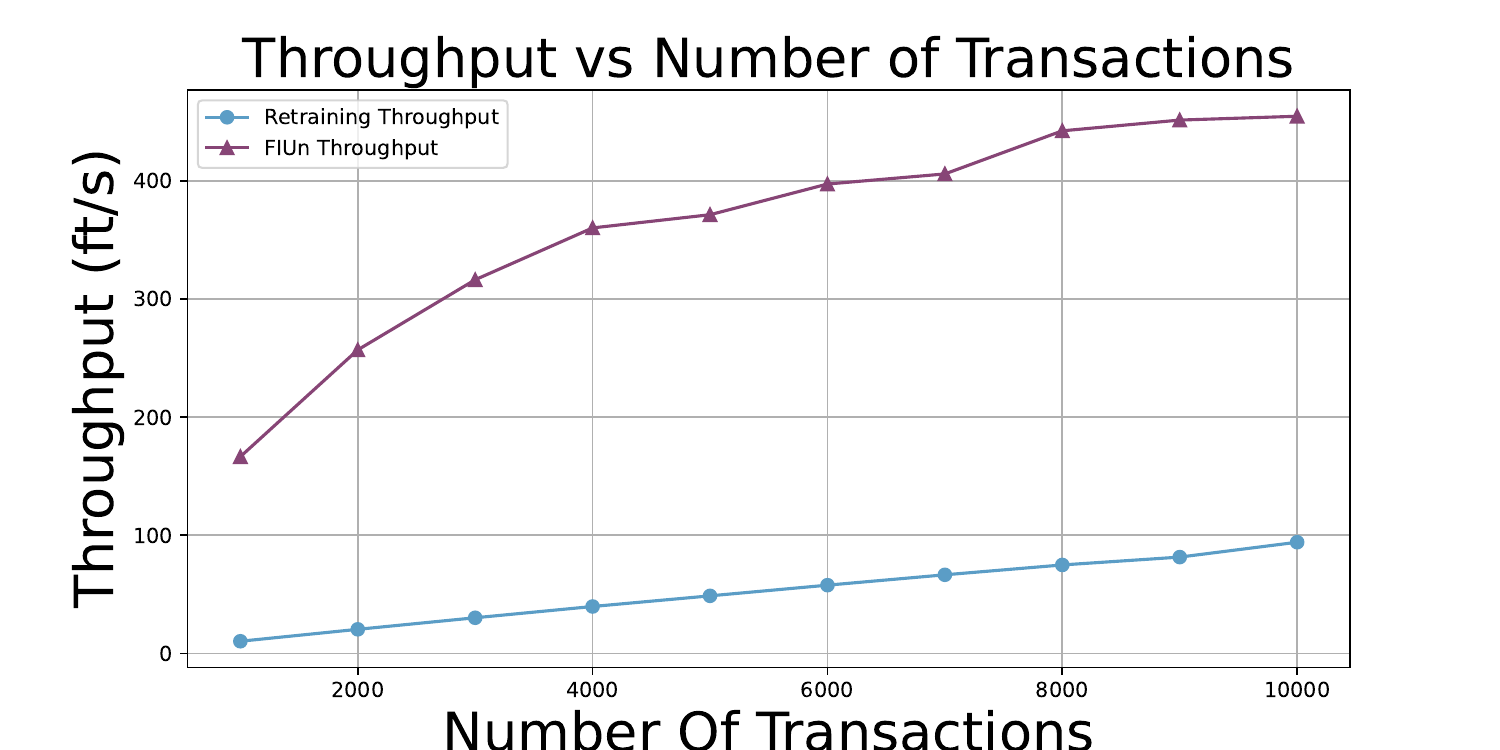}
      
        \label{fig:throughput_gcnn}
    }
  \subfigure[Latency, FL, Yahoo, GCNN]{
        \centering
        \includegraphics[width=0.22\textwidth]{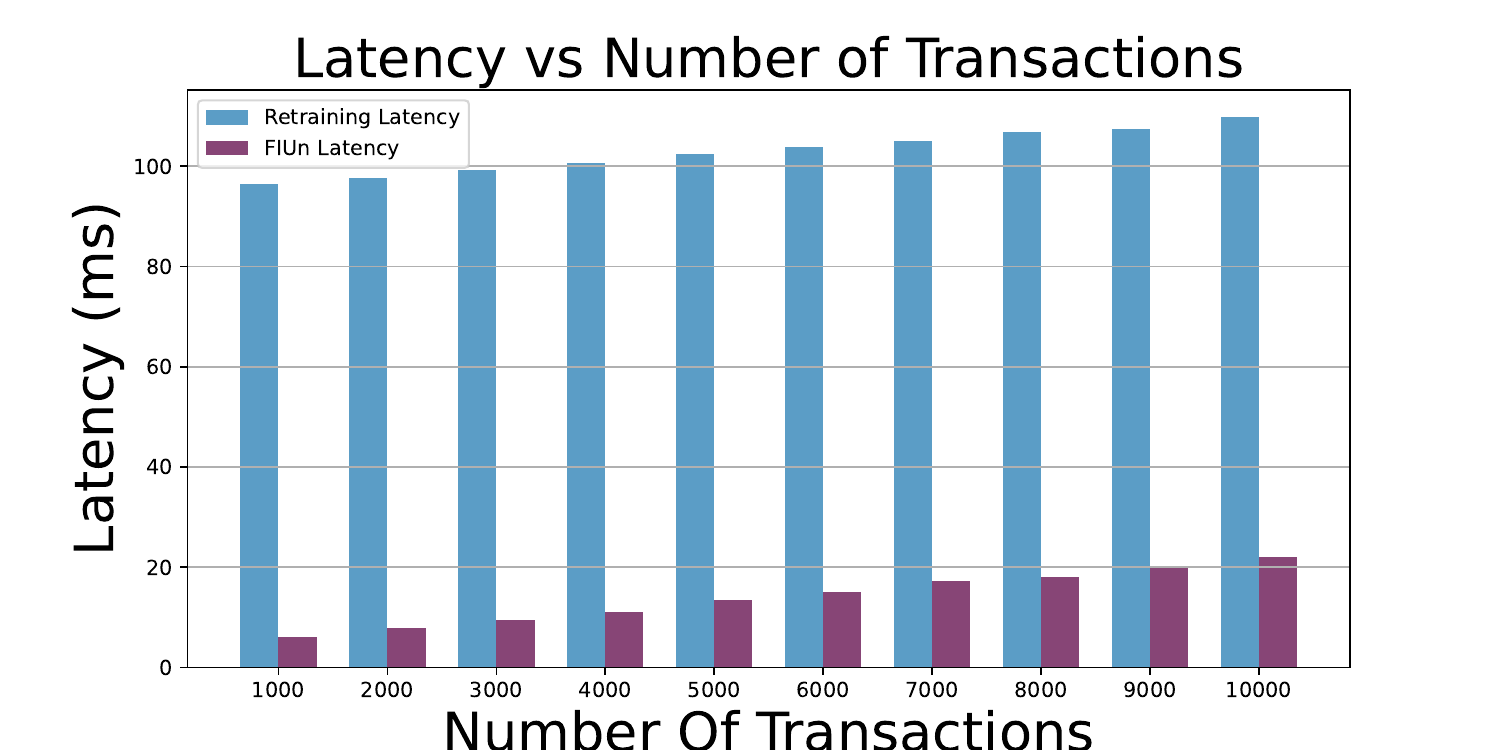}
      
        \label{fig:latency_gcnn}
    }

  \caption{Scalability results of the FIUn method in FL. Settings: unlearning requests from 1,000 to 10,000 (step 1,000), 8,000 concurrent requests, DAG depth 3, models include TinyImageNet (ResNet18, DenseNet161) and Yahoo (Bert, GCNN). Line charts show throughput vs. transactions; bar charts show latency.}
  \label{fig:scalibility}
  \vspace{-1em}
\end{figure*}

\subsection{Merged Label Unlearning Analysis}
This experiment assesses the applicability of the MFIM function to models with various distributions of unlearned labels. 
We analyze the unlearning effectiveness of various models across CIFAR100 DDPL. As demonstrated in Table~\ref{table:dis_cifar100_multi},
the average $AD_r$ is 89.72\%, and the average $AD_f$ is 5.62\%, confirming the unlearning effectiveness of our method.
We evaluate unlearning on TinyImageNet under FL. Table~\ref{table:fedrated_tinyimage_multi} shows that FIUn achieves complete forgetting of target classes while keeping retained-class accuracy high (92.4\%), demonstrating strong stability and generalization via the MFIM function. FIUn also greatly reduces cumulative unlearning time compared to retraining and maintains consistently low unlearning time across label distributions. Additional experiments are provided in \textbf{Appendix~\ref{appendix_b2}}.

We evaluate the MFIM function's ability to manage merged label unlearning tasks within FL. As shown in Fig.~\ref{fig:overlap_analysis}, our method achieves an average 88.54\% performance improvement in AlexNet and DenseNet161 models over direct FIM use. In the ResNet18 model, its unlearning effectiveness is comparable to that of FIM. This is because MFIM reduces cumulative errors (caused by the manual threshold rule) by merging FIMs from all relevant roots into a single calculation, allowing parameter adjustments to be made in one step instead of applying the threshold rule to each root~\cite{foster2024fast,golatkar2020eternal}, leading to more accurate and stable unlearning outcomes.

In experiments with BERT-base-cased and GCNN on the Yahoo! Answers dataset within FL (Table~\ref{table:Yahoo}), we tested 50 clients for multi-label unlearning. The $AD_f$ metric showed small increases in some cases (15.13\% and 26.73\%), but overall remained low. For GCNN, under $C_f$=2 and 4, $AD_r$ exceeded 63\%, though FIUn slightly reduced retained accuracy due to the large, unoptimized FIM hyperparameter space. Nevertheless, FIUn was highly time-efficient, significantly faster than re-training, and achieved effective unlearning with only minor accuracy trade-offs.

\begin{center}
\fbox{%
\begin{minipage}{0.96\linewidth}
\textbf{Takeaway--capable of multi-root unlearning.}
With up to 88.54\% improvement, the MFIM function enables FIUn to consistently manage merged unlearning across different tasks and models, a capability that standard FIM approaches cannot achieve due to their inability to aggregate multiple FIMs into a single and unified FIM that encapsulates all root updates.
\end{minipage}
}
\end{center}

\vspace{-1em}
\subsection{Inherited Depth of Model Analysis}

We assess FIUn’s performance impact as inheritance depth increases in FL and IL, using consistent hyperparameters across all experiments.
\begin{packeditemize}
    \item \textbf{Federated Learning.} 
    (\textcolor{light_cyan}{Cyan} in Fig.~\ref{fig:total_Experi_Scenarios_2}) In FL, the CIFAR100 and TinyImageNet datasets are randomly divided into five segments. Each model trains using one segment of the data that adopts a binary tree structure. We evaluate whether our method can effectively unlearn specified labels as the depth of the binary tree increases.

    \item \textbf{Incremental Learning.} (\textcolor{light_blue}{Blue} in Fig.~\ref{fig:total_Experi_Scenarios_2}) In IL, the starting model $w_g$ includes data from 170 classes. With each learning step, a new class is added to evaluate the effectiveness of our method in unlearning specific labels.

\end{packeditemize}


Figs.~\ref{fig:Reference_deepth} and~\ref{fig:Reference_deepth1_1} evaluate inheritance depth under FL and IL for $\#C_f\in{1,10}$. Overall, our method remains stable across depths; in FL, the unlearning accuracy $\bigtriangleup_{acc}$ stays near 100\% even at large depths. In IL, especially TinyImageNet + DenseNet161, $\bigtriangleup_{acc}$ declines and fluctuates as depth grows, likely because DenseNet161’s dense connectivity accumulates class-specific features that are harder to erase, and TinyImageNet’s strong inter-class overlap amplifies category and parameter drift; depth-dependent adaptation and data stochasticity further contribute to variability.

\vspace{-0.4em}
\begin{center}
\fbox{%
\begin{minipage}{0.96\linewidth}
\textbf{Takeaway--perfectly matching inheritance networks.}
FIUn exhibits the \textit{Hyper-Distance} property, effectively unlearning across up to 50 levels of inheritance depth in various models and datasets, enabling simultaneous updates of all inherited models during unlearning tasks.
\end{minipage}
}
\end{center}

 \vspace{-1em}
\subsection{Parameter Layer Selection in Models}
We evaluate the impact of the number of model layers used for FIM calculation on algorithm performance. Experiments were conducted in FL and IL frameworks with 15 label categories selected for unlearning. 
Fig.~\ref{fig:total_Experi_Scenarios_2} illustrates the DAG for these frameworks with Yellow and Blue models ($w_g$, $w_a$, $w_b$). FIM calculations were conducted on the last layer, last seven layers, and all layers using 10 random parameter sets.

We apply the methods from~\cite{foster2024fast,golatkar2020eternal} (using all layers) for unlearning in inherited models. As shown in Fig.~\ref{fig:unlearning_layer}, computing FIM on all layers leads to high variance in $\bigtriangleup_{acc}$. In contrast, the ``Last 1'' and ``Last 7'' settings show much smaller variance and achieve over 100\% better unlearning. This suggests that focusing on partial layers—especially the last layer—yields more stable and consistent results. The reason is that early layers capture general features, while later layers, particularly the last, encode task-specific features. Thus, targeting the last layer removes task-specific information more efficiently, which fits well with our MFIM design.

\vspace{-0.4em}
\begin{center}
\fbox{%
\begin{minipage}{0.96\linewidth}
\textbf{Takeaway--focusing on the last layer is the best for efficiency and unlearning effectiveness.}
FIUn achieves more efficient and effective unlearning by strategically performing FIM calculations on the weights of the last layer of models that the calculation of the MFIM function can benefit from.
\end{minipage}
}
\vspace{-1em}
\end{center}

\subsection{Unlearning Speed Analysis}
\begin{figure}[t]
    \centering
    \includegraphics[width=0.45\textwidth]{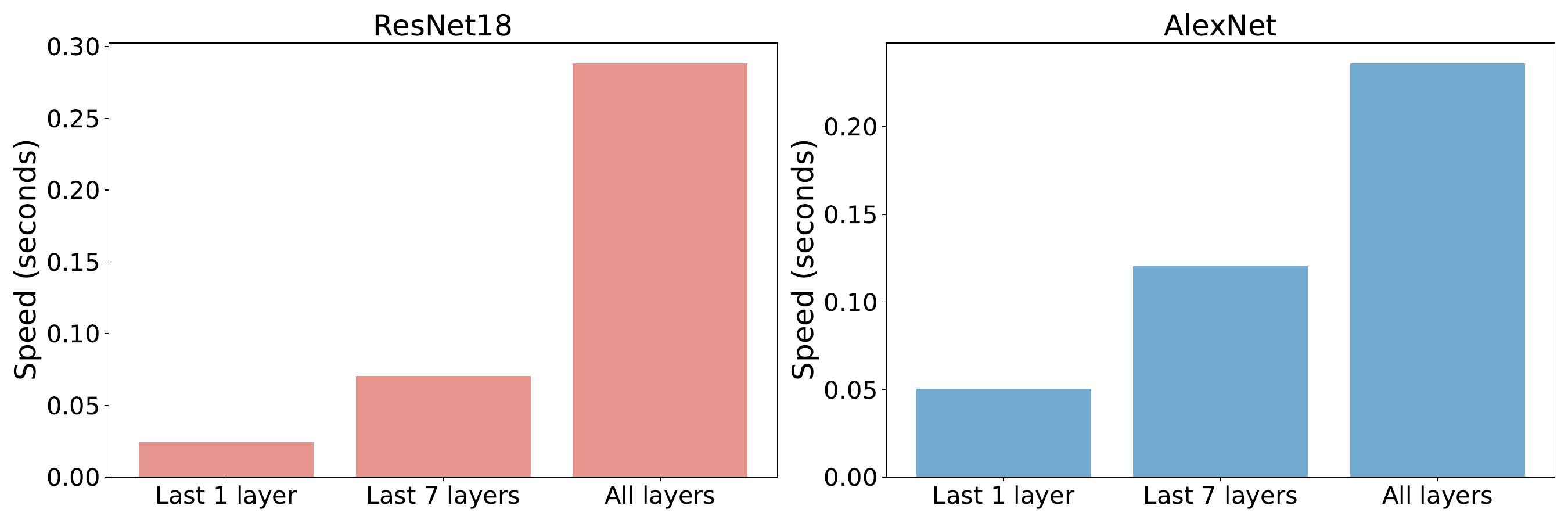}
    \caption{Unlearning speed of different layers of models. FIUn speed is based on the FL framework with CIFAR100.}
    \label{parameter_speed}
    \vspace{-1em}
\end{figure}

We conduct experiments on the inherited model $w_b$ using AlexNet and ResNet models, as shown in Fig.~\ref{fig:total_Experi_Scenarios_2}, while the unlearning speed is presented in Fig.~\ref{parameter_speed}. The results show that the unlearning time progressively increases with the number of unlearning layers, the parameters for AlexNet are 409600, 18833,764, and 20495268 for 1, 7, and all layers, respectively, while for ResNet18, they are 51200, 4903524, and 11227812, respectively. The results confirm that our proposed FIUn method with the last layer effectively enhances the efficiency of the unlearning process.

\vspace{-1em}

\subsection{Scalability Experiments}

To validate the scalability of FIUn, we conduct parallel experiments with the following key configurations:
\begin{itemize}
    \item \textbf{Unlearned data volume:} Requests range from 1,000 to 10,000, in increments of 1,000.
    \item \textbf{Concurrent requests:} A concurrency level of 8,000 is maintained.
    \item \textbf{Distributed participants:} Experiments use a DAG depth of 3, with scalability tested on federated setups across models such as TinyImageNet (ResNet18, DenseNet161) and Yahoo (Bert, GCNN).
\end{itemize}

Fig.~\ref{fig:scalibility} reports FL scalability with FIUn, measuring throughput and latency on TinyImageNet (ResNet18, DenseNet161) and Yahoo (BERT, GCNN). FIUn shows near-linear throughput growth and stable latency as transaction volume increases, reflecting its efficient parallel unlearning; the setup is reproducible and indicates real-world applicability. Additional experiments appear in \textbf{Appendix~\ref{appendix_c}}.

\vspace{-0.4em}
\section{Conclusion}\label{sec:conclusion}

We developed a novel parallel unlearning framework for models with dependence and inheritance, utilizing a chronological DAG to abstract and visualize model inheritance in FL, DDPL, IL, and TL. The FIUn method and MFIM function efficiently identified affected models and enabled parallel unlearning through the DAG. 
Experiments across various models and datasets demonstrated that FIUn achieved complete unlearning for single-class labels while maintaining an average retained accuracy of \textbf{94.53\%}. For multi-class labels, unlearning accuracy was just \textbf{1.07\%}, with retained accuracy at \textbf{84.77\%}. Our method is \textbf{99\%} faster than existing ones and handles model updates efficiently at all inheritance depths.







\bibliographystyle{IEEEtran}
\bibliography{main}

@article{dean2012large,
  title={Large scale distributed deep networks},
  author={Dean, Jeffrey and Corrado, Greg and Monga, Rajat and Chen, Kai and Devin, Matthieu and Mao, Mark and Ranzato, Marc'aurelio and Senior, Andrew and Tucker, Paul and Yang, Ke and others},
  journal={Advances in neural information processing systems},
  volume={25},
  year={2012}
}

@inproceedings{
wang2025gru,
title={{GRU}: Mitigating the Trade-off between Unlearning and Retention for {LLM}s},
author={Yue Wang and Qizhou Wang and Feng Liu and Wei Huang and Yali Du and Xiaojiang Du and Bo Han},
booktitle={Forty-second International Conference on Machine Learning},
year={2025}
}

@inproceedings{
yang2025exploring,
title={Exploring Criteria of Loss Reweighting to Enhance {LLM} Unlearning},
author={Puning Yang and Qizhou Wang and Zhuo Huang and Tongliang Liu and Chengqi Zhang and Bo Han},
booktitle={Forty-second International Conference on Machine Learning},
year={2025}
}

@inproceedings{polino2018model,
  title   ={Model compression via distillation and quantization},
  author  ={Polino, Antonio and Pascanu, Razvan and Alistarh, Dan},
  booktitle ={International Conference on Learning Representations},
  year  ={2018},
}

@inproceedings{jayaraman,
  title={Evaluating differentially private machine learning in practice},
  author={Jayaraman, Bargav and Evans, David},
  booktitle={28th USENIX Security Symposium (USENIX Security 19)},
  pages={1895--1912},
  year={2019},
}

@article{distil1Unl,
  title={Distill to Delete: Unlearning in Graph Networks with Knowledge Distillation},
  author={Sinha, Yash and Mandal, Murari and Kankanhalli, Mohan},
  journal={arXiv preprint arXiv:2309.16173},
  year={2023}
}

@article{li2020review,
  title={A review of applications in federated learning},
  author={Li, Li and Fan, Yuxi and Tse, Mike and Lin, Kuo-Yi},
  journal={Computers \& Industrial Engineering},
  volume={149},
  pages={106854},
  year={2020},
  publisher={Elsevier}
}

@article{ga,
  title={Negative preference optimization: From catastrophic collapse to effective unlearning},
  author={Zhang, Ruiqi and Lin, Licong and Bai, Yu and Mei, Song},
  journal={arXiv preprint arXiv:2404.05868},
  year={2024}
}

@inproceedings{retrain,
  title={Efficient two-stage model retraining for machine unlearning},
  author={Kim, Junyaup and Woo, Simon S},
  booktitle={Proceedings of the IEEE/CVF Conference on Computer Vision and Pattern Recognition},
  pages={4361--4369},
  year={2022}
}

@inproceedings{sisa,
  title={Machine unlearning},
  author={Bourtoule, Lucas and Chandrasekaran, Varun and Choquette-Choo, Christopher A and Jia, Hengrui and Travers, Adelin and Zhang, Baiwu and Lie, David and Papernot, Nicolas},
  booktitle={2021 IEEE Symposium on Security and Privacy (SP)},
  pages={141--159},
  year={2021},
  organization={IEEE}
}

@article{TL,
  title={A survey of transfer learning},
  author={Weiss, Karl and Khoshgoftaar, Taghi M and Wang, DingDing},
  journal={Journal of Big data},
  volume={3},
  pages={1--40},
  year={2016},
  publisher={Springer}
}

@article{IL,
  title={Three types of incremental learning},
  author={Van de Ven, Gido M and Tuytelaars, Tinne and Tolias, Andreas S},
  journal={Nature Machine Intelligence},
  volume={4},
  number={12},
  pages={1185--1197},
  year={2022},
  publisher={Nature Publishing Group UK London}
}

@article{ddap,
  title={Pytorch distributed: Experiences on accelerating data parallel training},
  author={Li, Shen and Zhao, Yanli and Varma, Rohan and Salpekar, Omkar and Noordhuis, Pieter and Li, Teng and Paszke, Adam and Smith, Jeff and Vaughan, Brian and Damania, Pritam and others},
  journal={arXiv preprint arXiv:2006.15704},
  year={2020}
}

@article{odusami2021analysis,
  title={Analysis of features of Alzheimer’s disease: Detection of early stage from functional brain changes in magnetic resonance images using a finetuned ResNet18 network},
  author={Odusami, Modupe and Maskeli{\=u}nas, Rytis and Dama{\v{s}}evi{\v{c}}ius, Robertas and Krilavi{\v{c}}ius, Tomas},
  journal={Diagnostics},
  volume={11},
  number={6},
  pages={1071},
  year={2021},
  publisher={MDPI}
}

@article{yu2023ironforge,
  title={Ironforge: An open, secure, fair, decentralized federated learning},
  author={Yu, Guangsheng and Wang, Xu and Sun, Caijun and Wang, Qin and Yu, Ping and Ni, Wei and Liu, Ren Ping},
  journal={IEEE Transactions on Neural Networks and Learning Systems},
  year={2023},
  publisher={IEEE}
}

@article{steven1993fundamentals,
  title={Fundamentals of statistical signal processing},
  author={Steven, M Kay},
  journal={PTR Prentice-Hall, Englewood Cliffs, NJ},
  volume={10},
  number={151045},
  pages={148},
  year={1993}
}

@inproceedings{golatkar2020eternal,
  title={Eternal sunshine of the spotless net: Selective forgetting in deep networks},
  author={Golatkar, Aditya and Achille, Alessandro and Soatto, Stefano},
  booktitle={Proceedings of the IEEE/CVF Conference on Computer Vision and Pattern Recognition},
  pages={9304--9312},
  year={2020}
}

@article{foster2024zero,
  title={Zero-shot machine unlearning at scale via lipschitz regularization},
  author={Foster, Jack and Fogarty, Kyle and Schoepf, Stefan and {\"O}ztireli, Cengiz and Brintrup, Alexandra},
  journal={arXiv preprint arXiv:2402.01401},
  year={2024}
}

@article{li2020federated,
  title={Federated optimization in heterogeneous networks},
  author={Li, Tian and Sahu, Anit Kumar and Zaheer, Manzil and Sanjabi, Maziar and Talwalkar, Ameet and Smith, Virginia},
  journal={Proceedings of Machine learning and systems},
  volume={2},
  pages={429--450},
  year={2020}
}

@inproceedings{krauss2024automatic,
  title={Automatic adversarial adaption for stealthy poisoning attacks in federated learning},
  author={Krau{\ss}, Torsten and K{\"o}nig, Jan and Dmitrienko, Alexandra and Kanzow, Christian},
  booktitle={To appear soon at the Network and Distributed System Security Symposium (NDSS)},
  year={2024}
}

@inproceedings{han2023anomaly,
  title={Anomaly Detection in the Open World: Normality Shift Detection, Explanation, and Adaptation.},
  author={Han, Dongqi and Wang, Zhiliang and Chen, Wenqi and Wang, Kai and Yu, Rui and Wang, Su and Zhang, Han and Wang, Zhihua and Jin, Minghui and Yang, Jiahai and others},
  booktitle={NDSS},
  year={2023}
}

@inproceedings{rajapaksha2023improving,
  title={Improving in-vehicle networks intrusion detection using on-device transfer learning},
  author={Rajapaksha, Sampath and Kalutarage, Harsha and Al-Kadri, M Omar and Petrovski, Andrei and Madzudzo, Garikayi},
  booktitle={Symposium on vehicles security and privacy},
  volume={10},
  year={2023}
}

@inproceedings{lee2023layer,
  title={Layer-wise adaptive model aggregation for scalable federated learning},
  author={Lee, Sunwoo and Zhang, Tuo and Avestimehr, A Salman},
  booktitle={Proceedings of the AAAI Conference on Artificial Intelligence},
  volume={37},
  number={7},
  pages={8491--8499},
  year={2023}
}

@inproceedings{foster2024fast,
  title={Fast machine unlearning without retraining through selective synaptic dampening},
  author={Foster, Jack and Schoepf, Stefan and Brintrup, Alexandra},
  booktitle={Proceedings of the AAAI Conference on Artificial Intelligence},
  volume={38},
  number={11},
  pages={12043--12051},
  year={2024}
}

@inproceedings{hoang2024learn,
  title={Learn to unlearn for deep neural networks: Minimizing unlearning interference with gradient projection},
  author={Hoang, Tuan and Rana, Santu and Gupta, Sunil and Venkatesh, Svetha},
  booktitle={Proceedings of the IEEE/CVF Winter Conference on Applications of Computer Vision},
  pages={4819--4828},
  year={2024}
}

@article{sepahvand2024data,
  title={Data Selection for Transfer Unlearning},
  author={Sepahvand, Nazanin Mohammadi and Dumoulin, Vincent and Triantafillou, Eleni and Dziugaite, Gintare Karolina},
  journal={arXiv preprint arXiv:2405.10425},
  year={2024}
}

@article{tarun2023fast,
  title={Fast yet effective machine unlearning},
  author={Tarun, Ayush K and Chundawat, Vikram S and Mandal, Murari and Kankanhalli, Mohan},
  journal={IEEE Transactions on Neural Networks and Learning Systems},
  year={2023},
  publisher={IEEE}
}

@article{liu2024,
  title={Decentralized Federated Unlearning on Blockchain},
  author={Xiao Liu and Mingyuan Li and Xu Wang and Guangsheng Yu and Wei Ni and Lixiang Li and Haipeng Peng and Renping Liu},
  journal={arXiv preprint arXiv:2402.16294},
  year={2024}
}

@article{li2021survey,
  title={A survey on federated learning systems: Vision, hype and reality for data privacy and protection},
  author={Li, Qinbin and Wen, Zeyi and Wu, Zhaomin and Hu, Sixu and Wang, Naibo and Li, Yuan and Liu, Xu and He, Bingsheng},
  journal={IEEE Transactions on Knowledge and Data Engineering},
  volume={35},
  number={4},
  pages={3347--3366},
  year={2021},
  publisher={IEEE}
}

@inproceedings{mbma,
author = {Matsui, Beatriz M. A. and Goya, Denise H.},
title = {{MLOps}: a guide to its adoption in the context of responsible AI},
year = {2023},
isbn = {9781450393195},
publisher = {Association for Computing Machinery},
address = {New York, NY, USA},
doi = {10.1145/3526073.3527591},
booktitle = {Proceedings of the 1st Workshop on Software Engineering for Responsible AI},
pages = {45–49},
numpages = {5},
location = {Pittsburgh, Pennsylvania},
series = {SE4RAI '22}
}

@article{voigt2017eu,
  author={Voigt, Paul and Von dem Bussche, Axel},
  title={The eu general data protection regulation (gdpr)},
  journal={A Practical Guide, 1st Ed., Cham: Springer International Publishing},
  volume={10},
  number={3152676},
  pages={10--5555},
  year={2017},
  publisher={Springer}
}

@misc{yu2024splitunlearning,
      title={Split Unlearning}, 
      author={Guangsheng Yu and Yanna Jiang and Qin Wang and Xu Wang and Baihe Ma and Caijun Sun and Wei Ni and Ren Ping Liu},
      year={2024},
      eprint={2308.10422},
      archivePrefix={arXiv},
      primaryClass={cs.CR},
      url={https://arxiv.org/abs/2308.10422}, 
}

@article{liu2024fishers,
  title={Fishers Harvest Parallel Unlearning in Inherited Model Networks},
  author={Liu, Xiao and Li, Mingyuan and Wang, Xu and Yu, Guangsheng and Ni, Wei and Li, Lixiang and Peng, Haipeng and Liu, Renping},
  journal={arXiv preprint arXiv:2408.08493},
  year={2024}
}

@article{gao2024verifi,
  title={Verifi: Towards verifiable federated unlearning},
  author={Gao, Xiangshan and Ma, Xingjun and Wang, Jingyi and Sun, Youcheng and Li, Bo and Ji, Shouling and Cheng, Peng and Chen, Jiming},
  journal={IEEE Transactions on Dependable and Secure Computing},
  volume={21},
  number={6},
  pages={5720--5736},
  year={2024},
  publisher={IEEE}
}

@inproceedings{pan2025federated,
  title={Federated unlearning with gradient descent and conflict mitigation},
  author={Pan, Zibin and Wang, Zhichao and Li, Chi and Zheng, Kaiyan and Wang, Boqi and Tang, Xiaoying and Zhao, Junhua},
  booktitle={Proceedings of the AAAI Conference on Artificial Intelligence},
  volume={39},
  number={19},
  pages={19804--19812},
  year={2025}
}

@article{tao2024communication,
  title={Communication efficient and provable federated unlearning},
  author={Tao, Youming and Wang, Cheng-Long and Pan, Miao and Yu, Dongxiao and Cheng, Xiuzhen and Wang, Di},
  journal={arXiv preprint arXiv:2401.11018},
  year={2024}
}

@article{wang2024server,
  title={Server-initiated federated unlearning to eliminate impacts of low-quality data},
  author={Wang, Pengfei and Song, Wei and Qi, Heng and Zhou, Changjun and Li, Fuliang and Wang, Yong and Sun, Peng and Zhang, Qiang},
  journal={IEEE Transactions on Services Computing},
  volume={17},
  number={3},
  pages={1196--1211},
  year={2024},
  publisher={IEEE}
}

@inproceedings{zuo2024ecil,
  title={Ecil-mu: Embedding based class incremental learning and machine unlearning},
  author={Zuo, Zhiwei and Tang, Zhuo and Wang, Bin and Li, Kenli and Datta, Anwitaman},
  booktitle={ICASSP 2024-2024 IEEE International Conference on Acoustics, Speech and Signal Processing (ICASSP)},
  pages={6275--6279},
  year={2024},
  organization={IEEE}
}

@article{ginart2019making,
  title={Making ai forget you: Data deletion in machine learning},
  author={Ginart, Antonio and Guan, Melody and Valiant, Gregory and Zou, James Y},
  journal={Advances in neural information processing systems},
  volume={32},
  year={2019}
}

@article{kurmanji2023towards,
  title={Towards unbounded machine unlearning},
  author={Kurmanji, Meghdad and Triantafillou, Peter and Hayes, Jamie and Triantafillou, Eleni},
  journal={Advances in neural information processing systems},
  volume={36},
  pages={1957--1987},
  year={2023}
}

@article{nguyen2020variational,
  title={Variational bayesian unlearning},
  author={Nguyen, Quoc Phong and Low, Bryan Kian Hsiang and Jaillet, Patrick},
  journal={Advances in Neural Information Processing Systems},
  volume={33},
  pages={16025--16036},
  year={2020}
}

@article{zhu2024decoupling,
  title={Decoupling the class label and the target concept in machine unlearning},
  author={Zhu, Jianing and Han, Bo and Yao, Jiangchao and Xu, Jianliang and Niu, Gang and Sugiyama, Masashi},
  journal={arXiv preprint arXiv:2406.08288},
  year={2024}
}

@article{amari1998natural,
  title={Natural gradient works efficiently in learning},
  author={Amari, Shun-Ichi},
  journal={Neural computation},
  volume={10},
  number={2},
  pages={251--276},
  year={1998},
  publisher={MIT Press}
}

@inproceedings{koh2017understanding,
  title={Understanding black-box predictions via influence functions},
  author={Koh, Pang Wei and Liang, Percy},
  booktitle={International conference on machine learning},
  pages={1885--1894},
  year={2017},
  organization={PMLR}
}

@inproceedings{lee2023undo,
  title={Undo: Effective and accurate unlearning method for deep neural networks},
  author={Lee, Sangyong and Woo, Simon S},
  booktitle={Proceedings of the 32nd ACM International Conference on Information and Knowledge Management},
  pages={4043--4047},
  year={2023}
}




\newpage

\appendices

\label{appendix_b}

\begin{table*}
\centering
\setlength{\extrarowheight}{0pt}
\addtolength{\extrarowheight}{\aboverulesep}
\addtolength{\extrarowheight}{\belowrulesep}
\setlength{\aboverulesep}{0pt}
\setlength{\belowrulesep}{0pt}
\caption{Federated Unlearning Performance on CIFAR100}
\label{table:fedrated_cifar100}
\resizebox{0.78\textwidth}{!}{
\begin{tabular}{c|c|c|ccccccccc|cccccc} 
\hline
\multirow{3}{*}{Model}              & \multirow{3}{*}{\#$C_f$} & \multirow{3}{*}{Metrics}                                                & \multicolumn{3}{c|}{\multirow{2}{*}{Original~(\%)}}                                                                   & \multicolumn{3}{c|}{\multirow{2}{*}{Re-training~(\%)}}                                                             & \multicolumn{3}{c|}{\multirow{2}{*}{FIUn~(\%)}}                                                                                      & \multicolumn{6}{c}{Cumulative Unlearning Time (s)}                                                                                                                                                                                                                              \\ 
\cline{13-18}
                                    &                          &                                                                         & \multicolumn{3}{c|}{}                                                                                                 & \multicolumn{3}{c|}{}                                                                                              & \multicolumn{3}{c|}{}                                                                                                                & \multicolumn{3}{c|}{Re-training}                                                     & \multicolumn{3}{c}{FIUn}                                                                                                                                                                 \\ 
\cline{4-18}
                                    &                          &                                                                         & \multicolumn{1}{c|}{$w_g$}            & \multicolumn{1}{c|}{$w_a$}            & \multicolumn{1}{c|}{$w_b$}            & \multicolumn{1}{c|}{$w_g$}           & \multicolumn{1}{c|}{$w_a$}           & \multicolumn{1}{c|}{$w_b$}           & \multicolumn{1}{c|}{$w_g$}                 & \multicolumn{1}{c|}{$w_a$}                 & $w_b$                                      & \multicolumn{1}{c|}{$w_g$} & \multicolumn{1}{c|}{$w_a$} & \multicolumn{1}{c|}{$w_b$} & \multicolumn{1}{c|}{$w_g$}                                  & \multicolumn{1}{c|}{$w_a$}                                  & $w_b$                                                        \\ 
\hline
\multirow{6}{*}{\rotcell{AlexNet}}  & \multirow{2}{*}{1}       & $AD_r\uparrow$                                                          & 99.99                                 & 99.99                                 & 99.81                                 & 99.99                                & 99.99                                & 98.80                                & {\cellcolor[rgb]{1,1,0.878}}\textbf{97.13} & {\cellcolor[rgb]{1,1,0.878}}\textbf{96.41} & {\cellcolor[rgb]{1,1,0.878}}\textbf{97.12} & \multirow{2}{*}{29.70}     & \multirow{2}{*}{59.06}     & \multirow{2}{*}{89.62}     & {\cellcolor[rgb]{1,1,0.878}}                                & {\cellcolor[rgb]{1,1,0.878}}                                & {\cellcolor[rgb]{1,1,0.878}}                                 \\
                                    &                          & \multicolumn{1}{l|}{{\cellcolor[rgb]{0.929,0.933,1}}${AD}_f\downarrow$} & {\cellcolor[rgb]{0.929,0.933,1}}99.99 & {\cellcolor[rgb]{0.929,0.933,1}}99.99 & {\cellcolor[rgb]{0.929,0.933,1}}99.66 & {\cellcolor[rgb]{0.929,0.933,1}}0.00 & {\cellcolor[rgb]{0.929,0.933,1}}0.00 & {\cellcolor[rgb]{0.929,0.933,1}}0.00 & {\cellcolor[rgb]{1,1,0.878}}\textbf{0.00}  & {\cellcolor[rgb]{1,1,0.878}}\textbf{0.00}  & {\cellcolor[rgb]{1,1,0.878}}\textbf{0.00}  &                            &                            &                            & \multirow{-2}{*}{{\cellcolor[rgb]{1,1,0.878}}\textbf{0.09}} & \multirow{-2}{*}{{\cellcolor[rgb]{1,1,0.878}}\textbf{0.11}} & \multirow{-2}{*}{{\cellcolor[rgb]{1,1,0.878}}\textbf{0.13}}  \\ 
\hhline{~-----------------}
                                    & \multirow{2}{*}{2}       & $AD_r\uparrow$~                                                         & 99.99                                 & 99.99                                 & 99.86                                 & 99.99                                & 99.99                                & 99.51                                & {\cellcolor[rgb]{1,1,0.878}}\textbf{91.89} & {\cellcolor[rgb]{1,1,0.878}}\textbf{90.78} & {\cellcolor[rgb]{1,1,0.878}}\textbf{95.66} & \multirow{2}{*}{28.69}     & \multirow{2}{*}{57.81}     & \multirow{2}{*}{84.97}     & {\cellcolor[rgb]{1,1,0.878}}                                & {\cellcolor[rgb]{1,1,0.878}}                                & {\cellcolor[rgb]{1,1,0.878}}                                 \\
                                    &                          & \multicolumn{1}{l|}{{\cellcolor[rgb]{0.929,0.933,1}}${AD}_f\downarrow$} & {\cellcolor[rgb]{0.929,0.933,1}}99.99 & {\cellcolor[rgb]{0.929,0.933,1}}99.99 & {\cellcolor[rgb]{0.929,0.933,1}}99.49 & {\cellcolor[rgb]{0.929,0.933,1}}0.00 & {\cellcolor[rgb]{0.929,0.933,1}}0.00 & {\cellcolor[rgb]{0.929,0.933,1}}0.00 & {\cellcolor[rgb]{1,1,0.878}}\textbf{0.00}  & {\cellcolor[rgb]{1,1,0.878}}\textbf{0.00}  & {\cellcolor[rgb]{1,1,0.878}}\textbf{0.00}  &                            &                            &                            & \multirow{-2}{*}{{\cellcolor[rgb]{1,1,0.878}}\textbf{0.11}} & \multirow{-2}{*}{{\cellcolor[rgb]{1,1,0.878}}\textbf{0.16}} & \multirow{-2}{*}{{\cellcolor[rgb]{1,1,0.878}}\textbf{0.13}}  \\ 
\hhline{~-----------------}
                                    & \multirow{2}{*}{4}       & $AD_r\uparrow$                                                          & 99.99                                 & 99.99                                 & 99.96                                 & 99.99                                & 99.99                                & 99.16                                & {\cellcolor[rgb]{1,1,0.878}}\textbf{85.79} & {\cellcolor[rgb]{1,1,0.878}}\textbf{78.54} & {\cellcolor[rgb]{1,1,0.878}}\textbf{89.40} & \multirow{2}{*}{26.39}     & \multirow{2}{*}{53.10}     & \multirow{2}{*}{81.96}     & {\cellcolor[rgb]{1,1,0.878}}                                & {\cellcolor[rgb]{1,1,0.878}}                                & {\cellcolor[rgb]{1,1,0.878}}                                 \\
                                    &                          & \multicolumn{1}{l|}{{\cellcolor[rgb]{0.929,0.933,1}}${AD}_f\downarrow$} & {\cellcolor[rgb]{0.929,0.933,1}}99.99 & {\cellcolor[rgb]{0.929,0.933,1}}99.99 & {\cellcolor[rgb]{0.929,0.933,1}}99.89 & {\cellcolor[rgb]{0.929,0.933,1}}0.00 & {\cellcolor[rgb]{0.929,0.933,1}}0.00 & {\cellcolor[rgb]{0.929,0.933,1}}0.00 & {\cellcolor[rgb]{1,1,0.878}}\textbf{0.00}  & {\cellcolor[rgb]{1,1,0.878}}\textbf{0.00}  & {\cellcolor[rgb]{1,1,0.878}}\textbf{0.00}  &                            &                            &                            & \multirow{-2}{*}{{\cellcolor[rgb]{1,1,0.878}}\textbf{0.10}} & \multirow{-2}{*}{{\cellcolor[rgb]{1,1,0.878}}\textbf{0.16}} & \multirow{-2}{*}{{\cellcolor[rgb]{1,1,0.878}}\textbf{0.13}}  \\ 
\hline
\multirow{6}{*}{\rotcell{ResNet18}} & \multirow{2}{*}{1}       & $AD_r\uparrow$                                                          & 99.99                                 & 99.99                                 & 99.99                                 & 99.99                                & 99.99                                & 99.99                                & {\cellcolor[rgb]{1,1,0.878}}\textbf{97.09} & {\cellcolor[rgb]{1,1,0.878}}\textbf{97.13} & {\cellcolor[rgb]{1,1,0.878}}\textbf{96.93} & \multirow{2}{*}{30.36}     & \multirow{2}{*}{61.40}     & \multirow{2}{*}{90.94}     & {\cellcolor[rgb]{1,1,0.878}}                                & {\cellcolor[rgb]{1,1,0.878}}                                & {\cellcolor[rgb]{1,1,0.878}}                                 \\
                                    &                          & \multicolumn{1}{l|}{{\cellcolor[rgb]{0.929,0.933,1}}${AD}_f\downarrow$} & {\cellcolor[rgb]{0.929,0.933,1}}99.99 & {\cellcolor[rgb]{0.929,0.933,1}}99.99 & {\cellcolor[rgb]{0.929,0.933,1}}99.99 & {\cellcolor[rgb]{0.929,0.933,1}}0.00 & {\cellcolor[rgb]{0.929,0.933,1}}0.00 & {\cellcolor[rgb]{0.929,0.933,1}}0.00 & {\cellcolor[rgb]{1,1,0.878}}\textbf{0.00}  & {\cellcolor[rgb]{1,1,0.878}}\textbf{0.00}  & {\cellcolor[rgb]{1,1,0.878}}\textbf{0.00}  &                            &                            &                            & \multirow{-2}{*}{{\cellcolor[rgb]{1,1,0.878}}\textbf{0.68}} & \multirow{-2}{*}{{\cellcolor[rgb]{1,1,0.878}}\textbf{1.00}} & \multirow{-2}{*}{{\cellcolor[rgb]{1,1,0.878}}\textbf{0.99}}  \\ 
\hhline{~-----------------}
                                    & \multirow{2}{*}{2}       & $AD_r\uparrow$                                                          & 99.99                                 & 99.99                                 & 99.99                                 & 99.99                                & 99.99                                & 99.99                                & {\cellcolor[rgb]{1,1,0.878}}\textbf{96.66} & {\cellcolor[rgb]{1,1,0.878}}\textbf{98.10} & {\cellcolor[rgb]{1,1,0.878}}\textbf{97.44} & \multirow{2}{*}{28.30}     & \multirow{2}{*}{57.39}     & \multirow{2}{*}{84.06}     & {\cellcolor[rgb]{1,1,0.878}}                                & {\cellcolor[rgb]{1,1,0.878}}                                & {\cellcolor[rgb]{1,1,0.878}}                                 \\
                                    &                          & \multicolumn{1}{l|}{{\cellcolor[rgb]{0.929,0.933,1}}${AD}_f\downarrow$} & {\cellcolor[rgb]{0.929,0.933,1}}99.99 & {\cellcolor[rgb]{0.929,0.933,1}}99.99 & {\cellcolor[rgb]{0.929,0.933,1}}99.99 & {\cellcolor[rgb]{0.929,0.933,1}}0.00 & {\cellcolor[rgb]{0.929,0.933,1}}0.00 & {\cellcolor[rgb]{0.929,0.933,1}}0.00 & {\cellcolor[rgb]{1,1,0.878}}\textbf{0.00}  & {\cellcolor[rgb]{1,1,0.878}}\textbf{0.00}  & {\cellcolor[rgb]{1,1,0.878}}\textbf{0.00}  &                            &                            &                            & \multirow{-2}{*}{{\cellcolor[rgb]{1,1,0.878}}\textbf{0.69}} & \multirow{-2}{*}{{\cellcolor[rgb]{1,1,0.878}}\textbf{1.03}} & \multirow{-2}{*}{{\cellcolor[rgb]{1,1,0.878}}\textbf{1.09}}  \\ 
\hhline{~-----------------}
                                    & \multirow{2}{*}{4}       & $AD_r\uparrow$                                                          & 99.99                                 & 99.99                                 & 99.99                                 & 99.96                                & 99.99                                & 99.99                                & {\cellcolor[rgb]{1,1,0.878}}\textbf{97.17} & {\cellcolor[rgb]{1,1,0.878}}\textbf{99.45} & {\cellcolor[rgb]{1,1,0.878}}\textbf{99.19} & \multirow{2}{*}{26.39}     & \multirow{2}{*}{52.69}     & \multirow{2}{*}{81.39}     & {\cellcolor[rgb]{1,1,0.878}}                                & {\cellcolor[rgb]{1,1,0.878}}                                & {\cellcolor[rgb]{1,1,0.878}}                                 \\
                                    &                          & \multicolumn{1}{l|}{{\cellcolor[rgb]{0.933,0.933,1}}${AD}_f\downarrow$} & {\cellcolor[rgb]{0.933,0.933,1}}99.99 & {\cellcolor[rgb]{0.933,0.933,1}}99.99 & {\cellcolor[rgb]{0.933,0.933,1}}99.99 & {\cellcolor[rgb]{0.933,0.933,1}}0.00 & {\cellcolor[rgb]{0.933,0.933,1}}0.00 & {\cellcolor[rgb]{0.933,0.933,1}}0.00 & {\cellcolor[rgb]{1,1,0.878}}\textbf{0.00}  & {\cellcolor[rgb]{1,1,0.878}}\textbf{0.00}  & {\cellcolor[rgb]{1,1,0.878}}\textbf{0.00}  &                            &                            &                            & \multirow{-2}{*}{{\cellcolor[rgb]{1,1,0.878}}\textbf{0.75}} & \multirow{-2}{*}{{\cellcolor[rgb]{1,1,0.878}}\textbf{1.09}} & \multirow{-2}{*}{{\cellcolor[rgb]{1,1,0.878}}\textbf{1.06}}  \\
\hline
\end{tabular}}
\vspace{-1em}
\end{table*}

\begin{table*}
\centering
\setlength{\extrarowheight}{0pt}
\addtolength{\extrarowheight}{\aboverulesep}
\addtolength{\extrarowheight}{\belowrulesep}
\setlength{\aboverulesep}{0pt}
\setlength{\belowrulesep}{0pt}
\caption{Incremental Unlearning Performance on CIFAR100}
\label{table:continue_unlearn}
\resizebox{0.78\textwidth}{!}{
\begin{tabular}{c|c|c|ccccccccc|cccccc} 
\hline
\multirow{3}{*}{Model}              & \multirow{3}{*}{\#$C_f$} & \multirow{3}{*}{Metrics}                                                & \multicolumn{3}{c|}{\multirow{2}{*}{Original (\%)}}                                                                   & \multicolumn{3}{c|}{\multirow{2}{*}{Re-training (\%)}}                                                             & \multicolumn{3}{c|}{\multirow{2}{*}{FIUn (\%)}}                                                                                      & \multicolumn{6}{c}{Cumulative Unlearning Time (s)}                                                                                                                                                                                                                              \\ 
\cline{13-18}
                                    &                          &                                                                         & \multicolumn{3}{c|}{}                                                                                                 & \multicolumn{3}{c|}{}                                                                                              & \multicolumn{3}{c|}{}                                                                                                                & \multicolumn{3}{c|}{Re-training}                                                     & \multicolumn{3}{c}{FIUn}                                                                                                                                                                 \\ 
\cline{4-18}
                                    &                          &                                                                         & \multicolumn{1}{c|}{$w_g$}            & \multicolumn{1}{c|}{$w_a$}            & \multicolumn{1}{c|}{$w_b$}            & \multicolumn{1}{c|}{$w_g$}           & \multicolumn{1}{c|}{$w_a$}           & \multicolumn{1}{c|}{$w_b$}           & \multicolumn{1}{c|}{$w_g$}                 & \multicolumn{1}{c|}{$w_a$}                 & $w_b$                                      & \multicolumn{1}{c|}{$w_g$} & \multicolumn{1}{c|}{$w_a$} & \multicolumn{1}{c|}{$w_b$} & \multicolumn{1}{c|}{$w_g$}                                  & \multicolumn{1}{c|}{$w_a$}                                  & $w_b$                                                        \\ 
\hline
\multirow{6}{*}{\rotcell{AlexNet}}  & \multirow{2}{*}{1}       & $AD_r\uparrow$                                                          & 99.99                                 & 99.99                                 & 99.81                                 & 99.99                                & 99.99                                & 98.80                                & {\cellcolor[rgb]{1,1,0.878}}\textbf{98.97} & {\cellcolor[rgb]{1,1,0.878}}\textbf{94.39} & {\cellcolor[rgb]{1,1,0.878}}\textbf{94.98} & \multirow{2}{*}{22.95}     & \multirow{2}{*}{43.83}     & \multirow{2}{*}{63.45}     & {\cellcolor[rgb]{1,1,0.878}}                                & {\cellcolor[rgb]{1,1,0.878}}                                & {\cellcolor[rgb]{1,1,0.878}}                                 \\
                                    &                          & \multicolumn{1}{l|}{{\cellcolor[rgb]{0.933,0.929,1}}${AD}_f\downarrow$} & {\cellcolor[rgb]{0.933,0.929,1}}99.99 & {\cellcolor[rgb]{0.933,0.929,1}}99.99 & {\cellcolor[rgb]{0.933,0.929,1}}99.66 & {\cellcolor[rgb]{0.933,0.929,1}}0.00 & {\cellcolor[rgb]{0.933,0.929,1}}0.00 & {\cellcolor[rgb]{0.933,0.929,1}}0.00 & {\cellcolor[rgb]{1,1,0.878}}\textbf{0.00}  & {\cellcolor[rgb]{1,1,0.878}}\textbf{0.00}  & {\cellcolor[rgb]{1,1,0.878}}\textbf{0.00}  &                            &                            &                            & \multirow{-2}{*}{{\cellcolor[rgb]{1,1,0.878}}\textbf{0.09}} & \multirow{-2}{*}{{\cellcolor[rgb]{1,1,0.878}}\textbf{0.39}} & \multirow{-2}{*}{{\cellcolor[rgb]{1,1,0.878}}\textbf{0.39}}  \\ 
\hhline{~-----------------}
                                    & \multirow{2}{*}{2}       & $AD_r\uparrow$~                                                         & 99.99                                 & 99.99                                 & 99.80                                 & 99.99                                & 99.99                                & 99.51                                & {\cellcolor[rgb]{1,1,0.878}}\textbf{92.93} & {\cellcolor[rgb]{1,1,0.878}}\textbf{71.58} & {\cellcolor[rgb]{1,1,0.878}}\textbf{69.14} & \multirow{2}{*}{21.35}     & \multirow{2}{*}{41.58}     & \multirow{2}{*}{62.10}     & {\cellcolor[rgb]{1,1,0.878}}                                & {\cellcolor[rgb]{1,1,0.878}}                                & {\cellcolor[rgb]{1,1,0.878}}                                 \\
                                    &                          & \multicolumn{1}{l|}{{\cellcolor[rgb]{0.933,0.929,1}}${AD}_f\downarrow$} & {\cellcolor[rgb]{0.933,0.929,1}}99.99 & {\cellcolor[rgb]{0.933,0.929,1}}99.99 & {\cellcolor[rgb]{0.933,0.929,1}}99.49 & {\cellcolor[rgb]{0.933,0.929,1}}0.00 & {\cellcolor[rgb]{0.933,0.929,1}}0.00 & {\cellcolor[rgb]{0.933,0.929,1}}0.00 & {\cellcolor[rgb]{1,1,0.878}}\textbf{0.00}  & {\cellcolor[rgb]{1,1,0.878}}\textbf{0.00}  & {\cellcolor[rgb]{1,1,0.878}}\textbf{0.00}  &                            &                            &                            & \multirow{-2}{*}{{\cellcolor[rgb]{1,1,0.878}}\textbf{0.11}} & \multirow{-2}{*}{{\cellcolor[rgb]{1,1,0.878}}\textbf{0.27}} & \multirow{-2}{*}{{\cellcolor[rgb]{1,1,0.878}}\textbf{0.26}}  \\ 
\hhline{~-----------------}
                                    & \multirow{2}{*}{4}       & $AD_r\uparrow$                                                          & 99.99                                 & 99.99                                 & 99.96                                 & 99.99                                & 99.99                                & 99.16                                & {\cellcolor[rgb]{1,1,0.878}}\textbf{88.77} & {\cellcolor[rgb]{1,1,0.878}}\textbf{59.62} & {\cellcolor[rgb]{1,1,0.878}}\textbf{58.81} & \multirow{2}{*}{20.52}     & \multirow{2}{*}{41.12}     & \multirow{2}{*}{60.69}     & {\cellcolor[rgb]{1,1,0.878}}                                & {\cellcolor[rgb]{1,1,0.878}}                                & {\cellcolor[rgb]{1,1,0.878}}                                 \\
                                    &                          & \multicolumn{1}{l|}{{\cellcolor[rgb]{0.933,0.933,1}}${AD}_f\downarrow$} & {\cellcolor[rgb]{0.933,0.933,1}}99.99 & {\cellcolor[rgb]{0.933,0.933,1}}99.99 & {\cellcolor[rgb]{0.933,0.933,1}}99.89 & {\cellcolor[rgb]{0.933,0.933,1}}0.00 & {\cellcolor[rgb]{0.933,0.933,1}}0.00 & {\cellcolor[rgb]{0.933,0.933,1}}0.00 & {\cellcolor[rgb]{1,1,0.878}}\textbf{0.00}  & {\cellcolor[rgb]{1,1,0.878}}\textbf{0.00}  & {\cellcolor[rgb]{1,1,0.878}}\textbf{0.00}  &                            &                            &                            & \multirow{-2}{*}{{\cellcolor[rgb]{1,1,0.878}}\textbf{0.12}} & \multirow{-2}{*}{{\cellcolor[rgb]{1,1,0.878}}\textbf{0.28}} & \multirow{-2}{*}{{\cellcolor[rgb]{1,1,0.878}}\textbf{0.29}}  \\ 
\hline
\multirow{6}{*}{\rotcell{ResNet18}} & \multirow{2}{*}{1}       & $AD_r\uparrow$                                                          & 99.99                                 & 99.99                                 & 99.99                                 & 99.99                                & 99.99                                & 99.99                                & {\cellcolor[rgb]{1,1,0.878}}\textbf{91.71} & {\cellcolor[rgb]{1,1,0.878}}\textbf{90.05} & {\cellcolor[rgb]{1,1,0.878}}\textbf{90.82} & \multirow{2}{*}{26.35}     & \multirow{2}{*}{47.12}     & \multirow{2}{*}{66.93}     & {\cellcolor[rgb]{1,1,0.878}}                                & {\cellcolor[rgb]{1,1,0.878}}                                & {\cellcolor[rgb]{1,1,0.878}}                                 \\
                                    &                          & \multicolumn{1}{l|}{{\cellcolor[rgb]{0.933,0.933,1}}${AD}_f\downarrow$} & {\cellcolor[rgb]{0.933,0.933,1}}99.99 & {\cellcolor[rgb]{0.933,0.933,1}}99.99 & {\cellcolor[rgb]{0.933,0.933,1}}99.99 & {\cellcolor[rgb]{0.933,0.933,1}}0.00 & {\cellcolor[rgb]{0.933,0.933,1}}0.00 & {\cellcolor[rgb]{0.933,0.933,1}}0.00 & {\cellcolor[rgb]{1,1,0.878}}\textbf{0.00}  & {\cellcolor[rgb]{1,1,0.878}}\textbf{0.00}  & {\cellcolor[rgb]{1,1,0.878}}\textbf{0.00}  &                            &                            &                            & \multirow{-2}{*}{{\cellcolor[rgb]{1,1,0.878}}\textbf{0.30}} & \multirow{-2}{*}{{\cellcolor[rgb]{1,1,0.878}}\textbf{0.80}} & \multirow{-2}{*}{{\cellcolor[rgb]{1,1,0.878}}\textbf{0.85}}  \\ 
\hhline{~-----------------}
                                    & \multirow{2}{*}{2}       & $AD_r\uparrow$                                                          & 99.99                                 & 99.99                                 & 99.99                                 & 99.99                                & 99.99                                & 99.99                                & {\cellcolor[rgb]{1,1,0.878}}\textbf{91.49} & {\cellcolor[rgb]{1,1,0.878}}\textbf{89.02} & {\cellcolor[rgb]{1,1,0.878}}\textbf{87.64} & \multirow{2}{*}{25.37}     & \multirow{2}{*}{45.93}     & \multirow{2}{*}{65.89}     & {\cellcolor[rgb]{1,1,0.878}}                                & {\cellcolor[rgb]{1,1,0.878}}                                & {\cellcolor[rgb]{1,1,0.878}}                                 \\
                                    &                          & \multicolumn{1}{l|}{{\cellcolor[rgb]{0.933,0.933,1}}${AD}_f\downarrow$} & {\cellcolor[rgb]{0.933,0.933,1}}99.99 & {\cellcolor[rgb]{0.933,0.933,1}}99.99 & {\cellcolor[rgb]{0.933,0.933,1}}99.99 & {\cellcolor[rgb]{0.933,0.933,1}}0.00 & {\cellcolor[rgb]{0.933,0.933,1}}0.00 & {\cellcolor[rgb]{0.933,0.933,1}}0.00 & {\cellcolor[rgb]{1,1,0.878}}\textbf{0.00}  & {\cellcolor[rgb]{1,1,0.878}}\textbf{0.00}  & {\cellcolor[rgb]{1,1,0.878}}\textbf{0.00}  &                            &                            &                            & \multirow{-2}{*}{{\cellcolor[rgb]{1,1,0.878}}\textbf{0.30}} & \multirow{-2}{*}{{\cellcolor[rgb]{1,1,0.878}}\textbf{0.84}} & \multirow{-2}{*}{{\cellcolor[rgb]{1,1,0.878}}\textbf{0.85}}  \\ 
\hhline{~-----------------}
                                    & \multirow{2}{*}{4}       & $AD_r\uparrow$                                                          & 99.99                                 & 99.99                                 & 99.99                                 & 99.96                                & 99.99                                & 99.99                                & {\cellcolor[rgb]{1,1,0.878}}\textbf{84.71} & {\cellcolor[rgb]{1,1,0.878}}\textbf{83.42} & {\cellcolor[rgb]{1,1,0.878}}\textbf{84.54} & \multirow{2}{*}{24.83}     & \multirow{2}{*}{43.23}     & \multirow{2}{*}{65.38}     & {\cellcolor[rgb]{1,1,0.878}}                                & {\cellcolor[rgb]{1,1,0.878}}                                & {\cellcolor[rgb]{1,1,0.878}}                                 \\
                                    &                          & \multicolumn{1}{l|}{{\cellcolor[rgb]{0.933,0.933,1}}${AD}_f\downarrow$} & {\cellcolor[rgb]{0.933,0.933,1}}99.99 & {\cellcolor[rgb]{0.933,0.933,1}}99.99 & {\cellcolor[rgb]{0.933,0.933,1}}99.99 & {\cellcolor[rgb]{0.933,0.933,1}}0.00 & {\cellcolor[rgb]{0.933,0.933,1}}0.00 & {\cellcolor[rgb]{0.933,0.933,1}}0.00 & {\cellcolor[rgb]{1,1,0.878}}\textbf{0.00}  & {\cellcolor[rgb]{1,1,0.878}}\textbf{0.00}  & {\cellcolor[rgb]{1,1,0.878}}\textbf{0.00}  &                            &                            &                            & \multirow{-2}{*}{{\cellcolor[rgb]{1,1,0.878}}\textbf{0.31}} & \multirow{-2}{*}{{\cellcolor[rgb]{1,1,0.878}}\textbf{0.86}} & \multirow{-2}{*}{{\cellcolor[rgb]{1,1,0.878}}\textbf{0.87}}  \\
\hline
\end{tabular}}
\vspace{-1em}
\end{table*}

\begin{table*}
\centering
\setlength{\extrarowheight}{0pt}
\addtolength{\extrarowheight}{\aboverulesep}
\addtolength{\extrarowheight}{\belowrulesep}
\setlength{\aboverulesep}{0pt}
\setlength{\belowrulesep}{0pt}
\caption{Transfer Unlearning Performance on TinyImageNet}
\label{table:transfer_unlearn_imagenet}
\resizebox{0.78\textwidth}{!}{
\begin{tabular}{c|c|c|ccccccccc|cccccc} 
\hline
\multirow{3}{*}{Model}                 & \multirow{3}{*}{\#$C_f$} & \multirow{3}{*}{Metrics}                                                & \multicolumn{3}{c|}{\multirow{2}{*}{Original (\%)}}                                                                   & \multicolumn{3}{c|}{\multirow{2}{*}{Re-training (\%)}}                                                             & \multicolumn{3}{c|}{\multirow{2}{*}{FIUn (\%)}}                                                                                      & \multicolumn{6}{c}{Cumulative Unlearning Time (s)}                                                                                                                                                                                                                              \\ 
\cline{13-18}
                                       &                          &                                                                         & \multicolumn{3}{c|}{}                                                                                                 & \multicolumn{3}{c|}{}                                                                                              & \multicolumn{3}{c|}{}                                                                                                                & \multicolumn{3}{c|}{Re-training}                                                     & \multicolumn{3}{c}{FIUn}                                                                                                                                                                 \\ 
\cline{4-18}
                                       &                          &                                                                         & \multicolumn{1}{c|}{$w_g$}            & \multicolumn{1}{c|}{$w_a$}            & \multicolumn{1}{c|}{$w_b$}            & \multicolumn{1}{c|}{$w_g$}           & \multicolumn{1}{c|}{$w_a$}           & \multicolumn{1}{c|}{$w_b$}           & \multicolumn{1}{c|}{$w_g$}                 & \multicolumn{1}{c|}{$w_a$}                 & $w_b$                                      & \multicolumn{1}{c|}{$w_g$} & \multicolumn{1}{c|}{$w_a$} & \multicolumn{1}{c|}{$w_b$} & \multicolumn{1}{c|}{$w_g$}                                  & \multicolumn{1}{c|}{$w_a$}                                  & $w_b$                                                        \\ 
\hline
\multirow{6}{*}{\rotcell{DenseNet161}} & \multirow{2}{*}{1}       & $AD_r\uparrow$                                                          & 86.33                                 & 91.24                                 & 89.34                                 & 97.75                                & 97.52                                & 96.75                                & {\cellcolor[rgb]{1,1,0.878}}\textbf{71.49} & {\cellcolor[rgb]{1,1,0.878}}\textbf{69.82} & {\cellcolor[rgb]{1,1,0.878}}\textbf{66.73} & \multirow{2}{*}{987.94}    & \multirow{2}{*}{1986.35}   & \multirow{2}{*}{1985.16}   & {\cellcolor[rgb]{1,1,0.878}}                                & {\cellcolor[rgb]{1,1,0.878}}                                & {\cellcolor[rgb]{1,1,0.878}}                                 \\
                                       &                          & \multicolumn{1}{l|}{{\cellcolor[rgb]{0.929,0.929,1}}${AD}_f\downarrow$} & {\cellcolor[rgb]{0.929,0.929,1}}94.50 & {\cellcolor[rgb]{0.929,0.929,1}}94.50 & {\cellcolor[rgb]{0.929,0.929,1}}96.70 & {\cellcolor[rgb]{0.929,0.929,1}}0.00 & {\cellcolor[rgb]{0.929,0.929,1}}0.00 & {\cellcolor[rgb]{0.929,0.929,1}}0.00 & {\cellcolor[rgb]{1,1,0.878}}\textbf{0.00}  & {\cellcolor[rgb]{1,1,0.878}}\textbf{0.00}  & {\cellcolor[rgb]{1,1,0.878}}\textbf{0.00}  &                            &                            &                            & \multirow{-2}{*}{{\cellcolor[rgb]{1,1,0.878}}\textbf{2.41}} & \multirow{-2}{*}{{\cellcolor[rgb]{1,1,0.878}}\textbf{4.15}} & \multirow{-2}{*}{{\cellcolor[rgb]{1,1,0.878}}\textbf{4.34}}  \\ 
\hhline{~-----------------}
                                       & \multirow{2}{*}{4}       & $AD_r\uparrow$~                                                         & 86.33                                 & 91.24                                 & 89.34                                 & 96.42                                & 96.41                                & 95.13                                & {\cellcolor[rgb]{1,1,0.878}}\textbf{64.24} & {\cellcolor[rgb]{1,1,0.878}}\textbf{55.32} & {\cellcolor[rgb]{1,1,0.878}}\textbf{51.18} & \multirow{2}{*}{1001.34}   & \multirow{2}{*}{1994.53}   & \multirow{2}{*}{1986.61}   & {\cellcolor[rgb]{1,1,0.878}}                                & {\cellcolor[rgb]{1,1,0.878}}                                & {\cellcolor[rgb]{1,1,0.878}}                                 \\
                                       &                          & \multicolumn{1}{l|}{{\cellcolor[rgb]{0.929,0.933,1}}${AD}_f\downarrow$} & {\cellcolor[rgb]{0.929,0.933,1}}89.12 & {\cellcolor[rgb]{0.929,0.933,1}}91.23 & {\cellcolor[rgb]{0.929,0.933,1}}91.84 & {\cellcolor[rgb]{0.929,0.933,1}}0.00 & {\cellcolor[rgb]{0.929,0.933,1}}0.00 & {\cellcolor[rgb]{0.929,0.933,1}}0.00 & {\cellcolor[rgb]{1,1,0.878}}\textbf{0.00}  & {\cellcolor[rgb]{1,1,0.878}}\textbf{0.00}  & {\cellcolor[rgb]{1,1,0.878}}\textbf{0.00}  &                            &                            &                            & \multirow{-2}{*}{{\cellcolor[rgb]{1,1,0.878}}\textbf{2.34}} & \multirow{-2}{*}{{\cellcolor[rgb]{1,1,0.878}}\textbf{4.71}} & \multirow{-2}{*}{{\cellcolor[rgb]{1,1,0.878}}\textbf{4.26}}  \\ 
\hhline{~-----------------}
                                       & \multirow{2}{*}{6}       & $AD_r\uparrow$                                                          & 86.33                                 & 91.24                                 & 89.34                                 & 95.72                                & 95.43                                & 93.58                                & {\cellcolor[rgb]{1,1,0.878}}\textbf{58.69} & {\cellcolor[rgb]{1,1,0.878}}\textbf{47.38} & {\cellcolor[rgb]{1,1,0.878}}\textbf{42.57} & \multirow{2}{*}{1018.14}   & \multirow{2}{*}{2008.89}   & \multirow{2}{*}{2009.31}   & {\cellcolor[rgb]{1,1,0.878}}                                & {\cellcolor[rgb]{1,1,0.878}}                                & {\cellcolor[rgb]{1,1,0.878}}                                 \\
                                       &                          & \multicolumn{1}{l|}{{\cellcolor[rgb]{0.929,0.933,1}}${AD}_f\downarrow$} & {\cellcolor[rgb]{0.929,0.933,1}}86.57 & {\cellcolor[rgb]{0.929,0.933,1}}89.25 & {\cellcolor[rgb]{0.929,0.933,1}}90.28 & {\cellcolor[rgb]{0.929,0.933,1}}0.00 & {\cellcolor[rgb]{0.929,0.933,1}}0.00 & {\cellcolor[rgb]{0.929,0.933,1}}0.00 & {\cellcolor[rgb]{1,1,0.878}}\textbf{0.00}  & {\cellcolor[rgb]{1,1,0.878}}\textbf{0.00}  & {\cellcolor[rgb]{1,1,0.878}}\textbf{0.00}  &                            &                            &                            & \multirow{-2}{*}{{\cellcolor[rgb]{1,1,0.878}}\textbf{2.16}} & \multirow{-2}{*}{{\cellcolor[rgb]{1,1,0.878}}\textbf{4.64}} & \multirow{-2}{*}{{\cellcolor[rgb]{1,1,0.878}}\textbf{4.46}}  \\ 
\hline
\multirow{6}{*}{\rotcell{ResNet18}}    & \multirow{2}{*}{1}       & $AD_r\uparrow$                                                          & 99.99                                 & 99.99                                 & 99.99                                 & 99.99                                & 99.99                                & 99.99                                & {\cellcolor[rgb]{1,1,0.878}}\textbf{98.66} & {\cellcolor[rgb]{1,1,0.878}}\textbf{98.07} & {\cellcolor[rgb]{1,1,0.878}}\textbf{96.74} & \multirow{2}{*}{142.15}    & \multirow{2}{*}{243.75}    & \multirow{2}{*}{241.61}    & {\cellcolor[rgb]{1,1,0.878}}                                & {\cellcolor[rgb]{1,1,0.878}}                                & {\cellcolor[rgb]{1,1,0.878}}                                 \\
                                       &                          & \multicolumn{1}{l|}{{\cellcolor[rgb]{0.929,0.933,1}}${AD}_f\downarrow$} & {\cellcolor[rgb]{0.929,0.933,1}}99.99 & {\cellcolor[rgb]{0.929,0.933,1}}99.99 & {\cellcolor[rgb]{0.929,0.933,1}}99.99 & {\cellcolor[rgb]{0.929,0.933,1}}0.00 & {\cellcolor[rgb]{0.929,0.933,1}}0.00 & {\cellcolor[rgb]{0.929,0.933,1}}0.00 & {\cellcolor[rgb]{1,1,0.878}}\textbf{0.00}  & {\cellcolor[rgb]{1,1,0.878}}\textbf{0.00}  & {\cellcolor[rgb]{1,1,0.878}}\textbf{0.00}  &                            &                            &                            & \multirow{-2}{*}{{\cellcolor[rgb]{1,1,0.878}}\textbf{0.42}} & \multirow{-2}{*}{{\cellcolor[rgb]{1,1,0.878}}\textbf{0.75}} & \multirow{-2}{*}{{\cellcolor[rgb]{1,1,0.878}}\textbf{0.78}}  \\ 
\hhline{~-----------------}
                                       & \multirow{2}{*}{4}       & $AD_r\uparrow$                                                          & 99.99                                 & 99.99                                 & 99.99                                 & 99.99                                & 99.99                                & 99.99                                & {\cellcolor[rgb]{1,1,0.878}}\textbf{98.84} & {\cellcolor[rgb]{1,1,0.878}}\textbf{98.38} & {\cellcolor[rgb]{1,1,0.878}}\textbf{97.38} & \multirow{2}{*}{143.46}    & \multirow{2}{*}{244.34}    & \multirow{2}{*}{245.74}    & {\cellcolor[rgb]{1,1,0.878}}                                & {\cellcolor[rgb]{1,1,0.878}}                                & {\cellcolor[rgb]{1,1,0.878}}                                 \\
                                       &                          & \multicolumn{1}{l|}{{\cellcolor[rgb]{0.933,0.933,1}}${AD}_f\downarrow$} & {\cellcolor[rgb]{0.933,0.933,1}}99.99 & {\cellcolor[rgb]{0.933,0.933,1}}99.99 & {\cellcolor[rgb]{0.933,0.933,1}}99.99 & {\cellcolor[rgb]{0.933,0.933,1}}0.00 & {\cellcolor[rgb]{0.933,0.933,1}}0.00 & {\cellcolor[rgb]{0.933,0.933,1}}0.00 & {\cellcolor[rgb]{1,1,0.878}}\textbf{0.00}  & {\cellcolor[rgb]{1,1,0.878}}\textbf{0.00}  & {\cellcolor[rgb]{1,1,0.878}}\textbf{0.00}  &                            &                            &                            & \multirow{-2}{*}{{\cellcolor[rgb]{1,1,0.878}}\textbf{0.42}} & \multirow{-2}{*}{{\cellcolor[rgb]{1,1,0.878}}\textbf{0.74}} & \multirow{-2}{*}{{\cellcolor[rgb]{1,1,0.878}}\textbf{0.75}}  \\ 
\hhline{~-----------------}
                                       & \multirow{2}{*}{6}       & $AD_r\uparrow$                                                          & 99.96                                 & 99.99                                 & 99.99                                 & 99.99                                & 99.99                                & 99.99                                & {\cellcolor[rgb]{1,1,0.878}}\textbf{98.82} & {\cellcolor[rgb]{1,1,0.878}}\textbf{98.91} & {\cellcolor[rgb]{1,1,0.878}}\textbf{98.98} & \multirow{2}{*}{147.27}    & \multirow{2}{*}{245.35}    & \multirow{2}{*}{246.41}    & {\cellcolor[rgb]{1,1,0.878}}                                & {\cellcolor[rgb]{1,1,0.878}}                                & {\cellcolor[rgb]{1,1,0.878}}                                 \\
                                       &                          & \multicolumn{1}{l|}{{\cellcolor[rgb]{0.933,0.933,1}}${AD}_f\downarrow$} & {\cellcolor[rgb]{0.933,0.933,1}}99.99 & {\cellcolor[rgb]{0.933,0.933,1}}99.99 & {\cellcolor[rgb]{0.933,0.933,1}}99.99 & {\cellcolor[rgb]{0.933,0.933,1}}0.00 & {\cellcolor[rgb]{0.933,0.933,1}}0.00 & {\cellcolor[rgb]{0.933,0.933,1}}0.00 & {\cellcolor[rgb]{1,1,0.878}}\textbf{0.00}  & {\cellcolor[rgb]{1,1,0.878}}\textbf{0.00}  & {\cellcolor[rgb]{1,1,0.878}}\textbf{0.00}  &                            &                            &                            & \multirow{-2}{*}{{\cellcolor[rgb]{1,1,0.878}}\textbf{0.56}} & \multirow{-2}{*}{{\cellcolor[rgb]{1,1,0.878}}\textbf{0.89}} & \multirow{-2}{*}{{\cellcolor[rgb]{1,1,0.878}}\textbf{0.74}}  \\
\hline
\end{tabular}}
\vspace{-1em}
\end{table*}

\begin{table*}
\centering
\setlength{\extrarowheight}{0pt}
\addtolength{\extrarowheight}{\aboverulesep}
\addtolength{\extrarowheight}{\belowrulesep}
\setlength{\aboverulesep}{0pt}
\setlength{\belowrulesep}{0pt}
\caption{Distributed Data-Parallel Unlearning Performance on TinyImageNet}
\label{table:distributed_unlearn_tinyimage}
\resizebox{0.78\textwidth}{!}{
\begin{tabular}{c|c|c|ccccccccc|cccccc} 
\hline
\multirow{3}{*}{Model}                 & \multirow{3}{*}{\#$C_f$} & \multirow{3}{*}{Metrics}                                                & \multicolumn{3}{c|}{\multirow{2}{*}{Original (\%)}}                                                                   & \multicolumn{3}{c|}{\multirow{2}{*}{Re-training~(\%)}}                                                             & \multicolumn{3}{c|}{\multirow{2}{*}{FIUn~(\%)}}                                                                                      & \multicolumn{6}{c}{Cumulative Unlearning Time (s)}                                                                                                                                                                                                                                                \\ 
\cline{13-18}
                                       &                          &                                                                         & \multicolumn{3}{c|}{}                                                                                                 & \multicolumn{3}{c|}{}                                                                                              & \multicolumn{3}{c|}{}                                                                                                                & \multicolumn{3}{c|}{Re-training}                                                     & \multicolumn{3}{c}{FIUn}                                                                                                                                                                                   \\ 
\cline{4-18}
                                       &                          &                                                                         & \multicolumn{1}{c|}{$w_g$}            & \multicolumn{1}{c|}{$w_a$}            & \multicolumn{1}{c|}{$w_b$}            & \multicolumn{1}{c|}{$w_g$}           & \multicolumn{1}{c|}{$w_a$}           & \multicolumn{1}{c|}{$w_b$}           & \multicolumn{1}{c|}{$w_g$}                 & \multicolumn{1}{c|}{$w_a$}                 & $w_b$                                      & \multicolumn{1}{c|}{$w_g$} & \multicolumn{1}{c|}{$w_a$} & \multicolumn{1}{c|}{$w_b$} & \multicolumn{1}{c|}{$w_g$}                                           & \multicolumn{1}{c|}{$w_a$}                                  & $w_b$                                                                 \\ 
\hline
\multirow{6}{*}{\rotcell{DenseNet161}} & \multirow{2}{*}{1}       & $AD_r\uparrow$                                                          & 99.99                                 & 87.16                                 & 85.21                                 & 99.99                                & 89.39                                & 87.24                                & {\cellcolor[rgb]{1,1,0.878}}\textbf{91.82} & {\cellcolor[rgb]{1,1,0.878}}\textbf{86.63} & {\cellcolor[rgb]{1,1,0.878}}\textbf{83.58} & \multirow{2}{*}{980.53}    & \multirow{2}{*}{988.42}    & \multirow{2}{*}{991.75}    & {\cellcolor[rgb]{1,1,0.878}}                                         & {\cellcolor[rgb]{1,1,0.878}}                                & {\cellcolor[rgb]{1,1,0.878}}                                          \\
                                       &                          & \multicolumn{1}{l|}{{\cellcolor[rgb]{0.933,0.933,1}}${AD}_f\downarrow$} & {\cellcolor[rgb]{0.933,0.933,1}}99.99 & {\cellcolor[rgb]{0.933,0.933,1}}92.75 & {\cellcolor[rgb]{0.933,0.933,1}}90.41 & {\cellcolor[rgb]{0.933,0.933,1}}0.00 & {\cellcolor[rgb]{0.933,0.933,1}}0.00 & {\cellcolor[rgb]{0.933,0.933,1}}0.00 & {\cellcolor[rgb]{1,1,0.878}}\textbf{0.00}  & {\cellcolor[rgb]{1,1,0.878}}\textbf{0.00}  & {\cellcolor[rgb]{1,1,0.878}}\textbf{0.00}  &                            &                            &                            & \multirow{-2}{*}{{\cellcolor[rgb]{1,1,0.878}}\textbf{\textbf{4.34}}} & \multirow{-2}{*}{{\cellcolor[rgb]{1,1,0.878}}\textbf{2.61}} & \multirow{-2}{*}{{\cellcolor[rgb]{1,1,0.878}}\textbf{\textbf{2.13}}}  \\ 
\hhline{~-----------------}
                                       & \multirow{2}{*}{4}       & $AD_r\uparrow$~                                                         & 99.96                                 & 87.00                                 & 85.21                                 & 99.96                                & 88.35                                & 89.34                                & {\cellcolor[rgb]{1,1,0.878}}\textbf{87.42} & {\cellcolor[rgb]{1,1,0.878}}\textbf{86.56} & {\cellcolor[rgb]{1,1,0.878}}\textbf{84.37} & \multirow{2}{*}{1004.63}   & \multirow{2}{*}{1006.32}   & \multirow{2}{*}{1005.332}  & {\cellcolor[rgb]{1,1,0.878}}                                         & {\cellcolor[rgb]{1,1,0.878}}                                & {\cellcolor[rgb]{1,1,0.878}}                                          \\
                                       &                          & \multicolumn{1}{l|}{{\cellcolor[rgb]{0.933,0.933,1}}${AD}_f\downarrow$} & {\cellcolor[rgb]{0.933,0.933,1}}99.96 & {\cellcolor[rgb]{0.933,0.933,1}}86.97 & {\cellcolor[rgb]{0.933,0.933,1}}88.85 & {\cellcolor[rgb]{0.933,0.933,1}}0.00 & {\cellcolor[rgb]{0.933,0.933,1}}0.00 & {\cellcolor[rgb]{0.933,0.933,1}}0.00 & {\cellcolor[rgb]{1,1,0.878}}\textbf{0.00}  & {\cellcolor[rgb]{1,1,0.878}}\textbf{0.00}  & {\cellcolor[rgb]{1,1,0.878}}\textbf{0.00}  &                            &                            &                            & \multirow{-2}{*}{{\cellcolor[rgb]{1,1,0.878}}\textbf{\textbf{4.31}}} & \multirow{-2}{*}{{\cellcolor[rgb]{1,1,0.878}}\textbf{2.05}} & \multirow{-2}{*}{{\cellcolor[rgb]{1,1,0.878}}\textbf{\textbf{2.35}}}  \\ 
\hhline{~-----------------}
                                       & \multirow{2}{*}{6}       & $AD_r\uparrow$                                                          & 99.99                                 & 87.00                                 & 85.21                                 & 99.96                                & 87.96                                & 88.15                                & {\cellcolor[rgb]{1,1,0.878}}\textbf{82.50} & {\cellcolor[rgb]{1,1,0.878}}\textbf{86.82} & {\cellcolor[rgb]{1,1,0.878}}\textbf{84.47} & \multirow{2}{*}{1024.25}   & \multirow{2}{*}{1036.14}   & \multirow{2}{*}{1026.74}   & {\cellcolor[rgb]{1,1,0.878}}                                         & {\cellcolor[rgb]{1,1,0.878}}                                & {\cellcolor[rgb]{1,1,0.878}}                                          \\
                                       &                          & \multicolumn{1}{l|}{{\cellcolor[rgb]{0.933,0.933,1}}${AD}_f\downarrow$} & {\cellcolor[rgb]{0.933,0.933,1}}99.99 & {\cellcolor[rgb]{0.933,0.933,1}}89.35 & {\cellcolor[rgb]{0.933,0.933,1}}86.08 & {\cellcolor[rgb]{0.933,0.933,1}}0.00 & {\cellcolor[rgb]{0.933,0.933,1}}0.00 & {\cellcolor[rgb]{0.933,0.933,1}}0.00 & {\cellcolor[rgb]{1,1,0.878}}\textbf{0.00}  & {\cellcolor[rgb]{1,1,0.878}}\textbf{0.00}  & {\cellcolor[rgb]{1,1,0.878}}\textbf{0.00}  &                            &                            &                            & \multirow{-2}{*}{{\cellcolor[rgb]{1,1,0.878}}\textbf{\textbf{4.16}}} & \multirow{-2}{*}{{\cellcolor[rgb]{1,1,0.878}}\textbf{2.24}} & \multirow{-2}{*}{{\cellcolor[rgb]{1,1,0.878}}\textbf{\textbf{2.21}}}  \\ 
\hline
\multirow{6}{*}{\rotcell{ResNet18}}    & \multirow{2}{*}{1}       & $AD_r\uparrow$                                                          & 99.99                                 & 99.99                                 & 99.99                                 & 99.99                                & 99.99                                & 99.99                                & {\cellcolor[rgb]{1,1,0.878}}\textbf{99.19} & {\cellcolor[rgb]{1,1,0.878}}\textbf{99.99} & {\cellcolor[rgb]{1,1,0.878}}\textbf{99.99} & \multirow{2}{*}{112.52}    & \multirow{2}{*}{117.34}    & \multirow{2}{*}{114.52}    & {\cellcolor[rgb]{1,1,0.878}}                                         & {\cellcolor[rgb]{1,1,0.878}}                                & {\cellcolor[rgb]{1,1,0.878}}                                          \\
                                       &                          & \multicolumn{1}{l|}{{\cellcolor[rgb]{0.933,0.933,1}}${AD}_f\downarrow$} & {\cellcolor[rgb]{0.933,0.933,1}}99.99 & {\cellcolor[rgb]{0.933,0.933,1}}99.99 & {\cellcolor[rgb]{0.933,0.933,1}}99.99 & {\cellcolor[rgb]{0.933,0.933,1}}0.00 & {\cellcolor[rgb]{0.933,0.933,1}}0.00 & {\cellcolor[rgb]{0.933,0.933,1}}0.00 & {\cellcolor[rgb]{1,1,0.878}}\textbf{0.00}  & {\cellcolor[rgb]{1,1,0.878}}\textbf{0.00}  & {\cellcolor[rgb]{1,1,0.878}}\textbf{0.00}  &                            &                            &                            & \multirow{-2}{*}{{\cellcolor[rgb]{1,1,0.878}}\textbf{\textbf{3.56}}} & \multirow{-2}{*}{{\cellcolor[rgb]{1,1,0.878}}\textbf{1.67}} & \multirow{-2}{*}{{\cellcolor[rgb]{1,1,0.878}}\textbf{\textbf{1.47}}}  \\ 
\hhline{~-----------------}
                                       & \multirow{2}{*}{4}       & $AD_r\uparrow$                                                          & 99.99                                 & 99.99                                 & 99.99                                 & 99.99                                & 99.99                                & 99.99                                & {\cellcolor[rgb]{1,1,0.878}}\textbf{99.99} & {\cellcolor[rgb]{1,1,0.878}}\textbf{99.99} & {\cellcolor[rgb]{1,1,0.878}}\textbf{99.99} & \multirow{2}{*}{109.32}    & \multirow{2}{*}{108.74}    & \multirow{2}{*}{110.53}    & {\cellcolor[rgb]{1,1,0.878}}                                         & {\cellcolor[rgb]{1,1,0.878}}                                & {\cellcolor[rgb]{1,1,0.878}}                                          \\
                                       &                          & \multicolumn{1}{l|}{{\cellcolor[rgb]{0.933,0.933,1}}${AD}_f\downarrow$} & {\cellcolor[rgb]{0.933,0.933,1}}99.99 & {\cellcolor[rgb]{0.933,0.933,1}}99.99 & {\cellcolor[rgb]{0.933,0.933,1}}99.99 & {\cellcolor[rgb]{0.933,0.933,1}}0.00 & {\cellcolor[rgb]{0.933,0.933,1}}0.00 & {\cellcolor[rgb]{0.933,0.933,1}}0.00 & {\cellcolor[rgb]{1,1,0.878}}\textbf{0.00}  & {\cellcolor[rgb]{1,1,0.878}}\textbf{0.00}  & {\cellcolor[rgb]{1,1,0.878}}\textbf{0.32}  &                            &                            &                            & \multirow{-2}{*}{{\cellcolor[rgb]{1,1,0.878}}\textbf{\textbf{3.54}}} & \multirow{-2}{*}{{\cellcolor[rgb]{1,1,0.878}}\textbf{1.74}} & \multirow{-2}{*}{{\cellcolor[rgb]{1,1,0.878}}\textbf{\textbf{1.37}}}  \\ 
\hhline{~-----------------}
                                       & \multirow{2}{*}{6}       & $AD_r\uparrow$                                                          & 99.96                                 & 99.99                                 & 99.99                                 & 99.96                                & 99.96                                & 99.96                                & {\cellcolor[rgb]{1,1,0.878}}\textbf{99.38} & {\cellcolor[rgb]{1,1,0.878}}\textbf{99.99} & {\cellcolor[rgb]{1,1,0.878}}\textbf{99.99} & \multirow{2}{*}{107.78}    & \multirow{2}{*}{108.32}    & \multirow{2}{*}{107.17}    & {\cellcolor[rgb]{1,1,0.878}}                                         & {\cellcolor[rgb]{1,1,0.878}}                                & {\cellcolor[rgb]{1,1,0.878}}                                          \\
                                       &                          & \multicolumn{1}{l|}{{\cellcolor[rgb]{0.933,0.933,1}}${AD}_f\downarrow$} & {\cellcolor[rgb]{0.933,0.933,1}}99.99 & {\cellcolor[rgb]{0.933,0.933,1}}99.99 & {\cellcolor[rgb]{0.933,0.933,1}}99.99 & {\cellcolor[rgb]{0.933,0.933,1}}0.00 & {\cellcolor[rgb]{0.933,0.933,1}}0.00 & {\cellcolor[rgb]{0.933,0.933,1}}0.00 & {\cellcolor[rgb]{1,1,0.878}}\textbf{0.00}  & {\cellcolor[rgb]{1,1,0.878}}\textbf{0.00}  & {\cellcolor[rgb]{1,1,0.878}}\textbf{0.00}  &                            &                            &                            & \multirow{-2}{*}{{\cellcolor[rgb]{1,1,0.878}}\textbf{\textbf{3.53}}} & \multirow{-2}{*}{{\cellcolor[rgb]{1,1,0.878}}\textbf{1.84}} & \multirow{-2}{*}{{\cellcolor[rgb]{1,1,0.878}}\textbf{\textbf{1.64}}}  \\
\hline
\end{tabular}}
\vspace{-1em}
\end{table*}

\section{General Label Unlearning Analysis}
\label{appendix_b1}
Table~\ref{table:fedrated_cifar100} to Table~\ref{table:distributed_unlearn_tinyimage} show that, on CIFAR100, FIUn achieves strong performance under federated
unlearning and incremental unlearning frameworks. Similarly, on TinyImageNet, FIUn exhibits excellent performance under transfer unlearning and distributed data-parallel unlearning. Re-training on
TinyImageNet takes a significantly longer time than CIFAR100. In comparison, FIUn maintains a high unlearning accuracy while keeping the unlearning time extremely short.


\section{Merged Label Unlearning Analysis}
\label{appendix_b2}

Tables~\ref{table:fedrated_cifar100_multi} to~\ref{table:DISTRI_tinyimage_multi} show that, in the federated unlearning setting on CIFAR100, our method achieves good performance on multi-label distributions. On TinyImageNet, under the distributed data-parallel unlearning setting, it demonstrates excellent performance on multi-label distributions. For multiple label distributions, our proposed MFIM
function accelerates the unlearning speed of inherited models
while maintaining high unlearning accuracy.

\section{More Datasets and Large Scale Analysis}
\label{appendix_c}
Tables~\ref{tab:xv} and~\ref{tab:xv1} present the experimental results of unlearning using a ViT model on ImageNet and TinyImageNet datasets. Table~\ref{tab:xv} shows that for ImageNet, when unlearning one class ($C_f=1$), the accuracy on retained data ($AD_r$) remains stable (73.8\% $\rightarrow$ 73.64\%) while the accuracy on unlearned data ($AD_f$) drops from 100\% to 0\%, with an unlearning time of 2.1s. For ten classes ($C_f=10$), $AD_f$ decreases from 89.04\% to 0\%, with $AD_r$ slightly reduces to 73.1\% and an unlearning time of 4.33s. In the TinyImageNet experiment, as shown in Table~\ref{tab:xv1}, we simulate a federated learning setup with 100 clients, each running the same ViT model. For both $C_f=1$ and $C_f=10$, $AD_f$ reaches 0\% after unlearning, while $AD_r$ remains above 95\%, with unlearning times ranging from 3.79s to 13.14s across clients.

\begin{table*}
\centering
\setlength{\extrarowheight}{0pt}
\addtolength{\extrarowheight}{\aboverulesep}
\addtolength{\extrarowheight}{\belowrulesep}
\setlength{\aboverulesep}{0pt}
\setlength{\belowrulesep}{0pt}
\caption{Federated Unlearning Performance on CIFAR-100 with Multiple Label Distribution}
\label{table:fedrated_cifar100_multi}
\resizebox{0.8\textwidth}{!}{
\begin{tabular}{c|c|c|ccccccccc|cccccc} 
\hline
\multirow{3}{*}{Model}              & \multirow{3}{*}{$\#\cap_{f}$} & \multirow{3}{*}{Metrics}                                                & \multicolumn{3}{c|}{\multirow{2}{*}{Original (\%)}}                                                                   & \multicolumn{3}{c|}{\multirow{2}{*}{Re-training~(\%)}}                                                             & \multicolumn{3}{c|}{\multirow{2}{*}{FIUn~(\%)}}                                                                                                        & \multicolumn{6}{c}{Cumulative Unlearning Time (s)}                                                                                                                                                                                                                                                \\ 
\cline{13-18}
                                    &                               &                                                                         & \multicolumn{3}{c|}{}                                                                                                 & \multicolumn{3}{c|}{}                                                                                              & \multicolumn{3}{c|}{}                                                                                                                                  & \multicolumn{3}{c|}{Re-training}                                                     & \multicolumn{3}{c}{FIUn}                                                                                                                                                                                   \\ 
\cline{4-18}
                                    &                               &                                                                         & \multicolumn{1}{c|}{$w_g$}            & \multicolumn{1}{c|}{$w_a$}            & \multicolumn{1}{c|}{$w_b$}            & \multicolumn{1}{c|}{$w_g$}           & \multicolumn{1}{c|}{$w_a$}           & \multicolumn{1}{c|}{$w_b$}           & \multicolumn{1}{c|}{$w_g$}                          & \multicolumn{1}{c|}{$w_a$}                 & $w_b$                                               & \multicolumn{1}{c|}{$w_g$} & \multicolumn{1}{c|}{$w_a$} & \multicolumn{1}{c|}{$w_b$} & \multicolumn{1}{c|}{$w_g$}                                           & \multicolumn{1}{c|}{$w_a$}                                  & $w_b$                                                                 \\ 
\hline
\multirow{6}{*}{\rotcell{AlexNet}}  & \multirow{2}{*}{60\%}         & $AD_r\uparrow$                                                          & 99.99                                 & 99.99                                 & 99.81                                 & 98.80                                & 99.99                                & 99.99                                & {\cellcolor[rgb]{1,1,0.878}}\textbf{69.42}          & {\cellcolor[rgb]{1,1,0.878}}\textbf{81.68} & {\cellcolor[rgb]{1,1,0.878}}\textbf{\textbf{75.05}} & \multirow{2}{*}{59.26}     & \multirow{2}{*}{30.13}     & \multirow{2}{*}{29.70}     & {\cellcolor[rgb]{1,1,0.878}}                                         & {\cellcolor[rgb]{1,1,0.878}}                                & {\cellcolor[rgb]{1,1,0.878}}                                          \\
                                    &                               & \multicolumn{1}{l|}{{\cellcolor[rgb]{0.878,0.878,1}}${AD}_f\downarrow$} & {\cellcolor[rgb]{0.878,0.878,1}}99.99 & {\cellcolor[rgb]{0.878,0.878,1}}99.99 & {\cellcolor[rgb]{0.878,0.878,1}}99.66 & {\cellcolor[rgb]{0.878,0.878,1}}0.00 & {\cellcolor[rgb]{0.878,0.878,1}}0.00 & {\cellcolor[rgb]{0.878,0.878,1}}0.00 & {\cellcolor[rgb]{1,1,0.878}}\textbf{0.00}           & {\cellcolor[rgb]{1,1,0.878}}\textbf{0.00}  & {\cellcolor[rgb]{1,1,0.878}}\textbf{\textbf{0.00}}  &                            &                            &                            & \multirow{-2}{*}{{\cellcolor[rgb]{1,1,0.878}}\textbf{\textbf{0.15}}} & \multirow{-2}{*}{{\cellcolor[rgb]{1,1,0.878}}\textbf{0.08}} & \multirow{-2}{*}{{\cellcolor[rgb]{1,1,0.878}}\textbf{\textbf{0.09}}}  \\ 
\hhline{~-----------------}
                                    & \multirow{2}{*}{40\%}         & $AD_r\uparrow$~                                                         & 99.99                                 & 99.99                                 & 99.80                                 & 99.51                                & 99.99                                & 99.99                                & {\cellcolor[rgb]{1,1,0.878}}\textbf{61.66}          & {\cellcolor[rgb]{1,1,0.878}}\textbf{79.77} & {\cellcolor[rgb]{1,1,0.878}}\textbf{\textbf{72.96}} & \multirow{2}{*}{53.84}     & \multirow{2}{*}{29.58}     & \multirow{2}{*}{28.69}     & {\cellcolor[rgb]{1,1,0.878}}                                         & {\cellcolor[rgb]{1,1,0.878}}                                & {\cellcolor[rgb]{1,1,0.878}}                                          \\
                                    &                               & \multicolumn{1}{l|}{{\cellcolor[rgb]{0.878,0.878,1}}${AD}_f\downarrow$} & {\cellcolor[rgb]{0.878,0.878,1}}99.99 & {\cellcolor[rgb]{0.878,0.878,1}}99.99 & {\cellcolor[rgb]{0.878,0.878,1}}99.49 & {\cellcolor[rgb]{0.878,0.878,1}}0.00 & {\cellcolor[rgb]{0.878,0.878,1}}0.00 & {\cellcolor[rgb]{0.878,0.878,1}}0.00 & {\cellcolor[rgb]{1,1,0.878}}\textbf{0.00}           & {\cellcolor[rgb]{1,1,0.878}}\textbf{0.00}  & {\cellcolor[rgb]{1,1,0.878}}\textbf{\textbf{0.00}}  &                            &                            &                            & \multirow{-2}{*}{{\cellcolor[rgb]{1,1,0.878}}\textbf{0.11}}          & \multirow{-2}{*}{{\cellcolor[rgb]{1,1,0.878}}\textbf{0.09}} & \multirow{-2}{*}{{\cellcolor[rgb]{1,1,0.878}}\textbf{0.19}}           \\ 
\hhline{~-----------------}
                                    & \multirow{2}{*}{20\%}         & $AD_r\uparrow$                                                          & 99.99                                 & 99.99                                 & 99.96                                 & 99.16                                & 99.99                                & 99.99                                & {\cellcolor[rgb]{1,1,0.878}}\textbf{59.57}          & {\cellcolor[rgb]{1,1,0.878}}\textbf{77.47} & {\cellcolor[rgb]{1,1,0.878}}\textbf{\textbf{77.97}} & \multirow{2}{*}{57.36}     & \multirow{2}{*}{27.48}     & \multirow{2}{*}{26.39}     & {\cellcolor[rgb]{1,1,0.878}}                                         & {\cellcolor[rgb]{1,1,0.878}}                                & {\cellcolor[rgb]{1,1,0.878}}                                          \\
                                    &                               & \multicolumn{1}{l|}{{\cellcolor[rgb]{0.878,0.878,1}}${AD}_f\downarrow$} & {\cellcolor[rgb]{0.878,0.878,1}}99.99 & {\cellcolor[rgb]{0.878,0.878,1}}99.99 & {\cellcolor[rgb]{0.878,0.878,1}}99.89 & {\cellcolor[rgb]{0.878,0.878,1}}0.00 & {\cellcolor[rgb]{0.878,0.878,1}}0.00 & {\cellcolor[rgb]{0.878,0.878,1}}0.00 & {\cellcolor[rgb]{1,1,0.878}}\textbf{0.12}           & {\cellcolor[rgb]{1,1,0.878}}\textbf{0.00}  & {\cellcolor[rgb]{1,1,0.878}}\textbf{\textbf{0.00}}  &                            &                            &                            & \multirow{-2}{*}{{\cellcolor[rgb]{1,1,0.878}}\textbf{0.10}}          & \multirow{-2}{*}{{\cellcolor[rgb]{1,1,0.878}}\textbf{0.07}} & \multirow{-2}{*}{{\cellcolor[rgb]{1,1,0.878}}\textbf{0.18}}           \\ 
\hline
\multirow{6}{*}{\rotcell{ResNet18}} & \multirow{2}{*}{60\%}         & $AD_r\uparrow$                                                          & 99.99                                 & 99.99                                 & 99.99                                 & 99.99                                & 99.99                                & 99.99                                & {\cellcolor[rgb]{1,1,0.878}}\textbf{\textbf{93.15}} & {\cellcolor[rgb]{1,1,0.878}}\textbf{98.76} & {\cellcolor[rgb]{1,1,0.878}}\textbf{\textbf{97.81}} & \multirow{2}{*}{67.31}     & \multirow{2}{*}{33.18}     & \multirow{2}{*}{30.36}     & {\cellcolor[rgb]{1,1,0.878}}                                         & {\cellcolor[rgb]{1,1,0.878}}                                & {\cellcolor[rgb]{1,1,0.878}}                                          \\
                                    &                               & \multicolumn{1}{l|}{{\cellcolor[rgb]{0.878,0.878,1}}${AD}_f\downarrow$} & {\cellcolor[rgb]{0.878,0.878,1}}99.99 & {\cellcolor[rgb]{0.878,0.878,1}}99.99 & {\cellcolor[rgb]{0.878,0.878,1}}99.99 & {\cellcolor[rgb]{0.878,0.878,1}}0.00 & {\cellcolor[rgb]{0.878,0.878,1}}0.00 & {\cellcolor[rgb]{0.878,0.878,1}}0.00 & {\cellcolor[rgb]{1,1,0.878}}\textbf{0.00}           & {\cellcolor[rgb]{1,1,0.878}}\textbf{0.00}  & {\cellcolor[rgb]{1,1,0.878}}\textbf{0.00}           &                            &                            &                            & \multirow{-2}{*}{{\cellcolor[rgb]{1,1,0.878}}\textbf{\textbf{0.60}}} & \multirow{-2}{*}{{\cellcolor[rgb]{1,1,0.878}}\textbf{0.32}} & \multirow{-2}{*}{{\cellcolor[rgb]{1,1,0.878}}\textbf{\textbf{0.29}}}  \\ 
\hhline{~-----------------}
                                    & \multirow{2}{*}{40\%}         & $AD_r\uparrow$                                                          & 99.99                                 & 99.99                                 & 99.99                                 & 99.99                                & 99.99                                & 99.99                                & {\cellcolor[rgb]{1,1,0.878}}\textbf{\textbf{93.78}} & {\cellcolor[rgb]{1,1,0.878}}\textbf{99.04} & {\cellcolor[rgb]{1,1,0.878}}\textbf{\textbf{98.98}} & \multirow{2}{*}{52.33}     & \multirow{2}{*}{28.13}     & \multirow{2}{*}{28.30}     & {\cellcolor[rgb]{1,1,0.878}}                                         & {\cellcolor[rgb]{1,1,0.878}}                                & {\cellcolor[rgb]{1,1,0.878}}                                          \\
                                    &                               & \multicolumn{1}{l|}{{\cellcolor[rgb]{0.878,0.878,1}}${AD}_f\downarrow$} & {\cellcolor[rgb]{0.878,0.878,1}}99.99 & {\cellcolor[rgb]{0.878,0.878,1}}99.99 & {\cellcolor[rgb]{0.878,0.878,1}}99.99 & {\cellcolor[rgb]{0.878,0.878,1}}0.00 & {\cellcolor[rgb]{0.878,0.878,1}}0.00 & {\cellcolor[rgb]{0.878,0.878,1}}0.00 & {\cellcolor[rgb]{1,1,0.878}}\textbf{0.00}           & {\cellcolor[rgb]{1,1,0.878}}\textbf{0.00}  & {\cellcolor[rgb]{1,1,0.878}}\textbf{0.00}           &                            &                            &                            & \multirow{-2}{*}{{\cellcolor[rgb]{1,1,0.878}}\textbf{\textbf{0.59}}} & \multirow{-2}{*}{{\cellcolor[rgb]{1,1,0.878}}\textbf{0.34}} & \multirow{-2}{*}{{\cellcolor[rgb]{1,1,0.878}}\textbf{\textbf{0.30}}}  \\ 
\hhline{~-----------------}
                                    & \multirow{2}{*}{20\%}         & $AD_r\uparrow$                                                          & 99.99                                 & 99.99                                 & 99.99                                 & 99.96                                & 99.99                                & 99.99                                & {\cellcolor[rgb]{1,1,0.878}}\textbf{\textbf{88.54}} & {\cellcolor[rgb]{1,1,0.878}}\textbf{98.79} & {\cellcolor[rgb]{1,1,0.878}}\textbf{\textbf{98.62}} & \multirow{2}{*}{56.61}     & \multirow{2}{*}{23.37}     & \multirow{2}{*}{26.39}     & {\cellcolor[rgb]{1,1,0.878}}                                         & {\cellcolor[rgb]{1,1,0.878}}                                & {\cellcolor[rgb]{1,1,0.878}}                                          \\
                                    &                               & \multicolumn{1}{l|}{{\cellcolor[rgb]{0.878,0.878,1}}${AD}_f\downarrow$} & {\cellcolor[rgb]{0.878,0.878,1}}99.99 & {\cellcolor[rgb]{0.878,0.878,1}}99.99 & {\cellcolor[rgb]{0.878,0.878,1}}99.99 & {\cellcolor[rgb]{0.878,0.878,1}}0.00 & {\cellcolor[rgb]{0.878,0.878,1}}0.00 & {\cellcolor[rgb]{0.878,0.878,1}}0.00 & {\cellcolor[rgb]{1,1,0.878}}\textbf{0.00}           & {\cellcolor[rgb]{1,1,0.878}}\textbf{0.00}  & {\cellcolor[rgb]{1,1,0.878}}\textbf{0.00}           &                            &                            &                            & \multirow{-2}{*}{{\cellcolor[rgb]{1,1,0.878}}\textbf{\textbf{0.58}}} & \multirow{-2}{*}{{\cellcolor[rgb]{1,1,0.878}}\textbf{0.28}} & \multirow{-2}{*}{{\cellcolor[rgb]{1,1,0.878}}\textbf{\textbf{0.30}}}  \\
\hline
\end{tabular}}
\vspace{-1em}
\end{table*}

\begin{table*}
\centering
\caption{ImageNet Experiment Results for Unlearning with Pretrained ViT Model}
\label{tab:xv}
\begin{tabular}{cccccc} 
\toprule
\textbf{Unlearning Label} & \textbf{AD\_r (\%)} & \textbf{AD\_f (\%)} & \textbf{AD\_r after Unlearning (\%)} & \textbf{AD\_f after Unlearning (\%)} & \textbf{Unlearning Time (s)}  \\ 
\midrule
One Class ($C_f = 1$)     & 73.8                & 100                 & 73.64                                & 0                                    & 2.1                           \\
Ten Classes ($C_f = 10$)  & 73.8                & 89.04               & 73.1                                 & 0                                    & 4.33                          \\
\bottomrule
\end{tabular}
\end{table*}

\begin{table*}
\centering
\caption{TinyImageNet Experiment Results for Unlearning in Federated Learning Setup with ViT Model}
\label{tab:xv1}
\begin{tabular}{ccccc} 
\toprule
\textbf{Unlearning Case}                  & \textbf{Client} & \textbf{Unlearning Time (s)} & \textbf{$AD_f$ after Unlearning (\%)} & \textbf{$AD_r$ after Unlearning (\%)}  \\ 
\midrule
\multirow{3}{*}{One Class ($C_f = 1$)}    & Client 1        & 3.79                         & 0.00                               & 95.63                               \\
                                          & Client 2        & 7.94                         & 0.00                               & 99.98                               \\
                                          & Client 3        & 12.45                        & 0.00                               & 99.63                               \\ 
\midrule
\multirow{3}{*}{Ten Classes ($C_f = 10$)} & Client 1        & 3.84                         & 0.00                               & 99.99                               \\
                                          & Client 2        & 7.96                         & 0.00                               & 99.99                               \\
                                          & Client 3        & 13.14                        & 0.00                               & 99.99                               \\
\bottomrule
\end{tabular}
\end{table*}

\begin{table*}
\centering
\setlength{\extrarowheight}{0pt}
\addtolength{\extrarowheight}{\aboverulesep}
\addtolength{\extrarowheight}{\belowrulesep}
\setlength{\aboverulesep}{0pt}
\setlength{\belowrulesep}{0pt}
\caption{Distributed Data-Parallel Unlearning Performance on TinyImageNet with Multiple Label Distribution}
\label{table:DISTRI_tinyimage_multi}
\resizebox{0.8\textwidth}{!}{
\begin{tabular}{c|c|c|ccccccccc|cccccc} 
\hline
\multirow{3}{*}{Model}                 & \multirow{3}{*}{\#$\cap_f$} & \multirow{3}{*}{Metrics}                                                & \multicolumn{3}{c|}{\multirow{2}{*}{Original (\%)}}                                                                   & \multicolumn{3}{c|}{\multirow{2}{*}{Re-training~(\%)}}                                                             & \multicolumn{3}{c|}{\multirow{2}{*}{FIUn~(\%)}}                                                                                                      & \multicolumn{6}{c}{Unlearning Time (s)}                                                                                                                                                                                                                                                           \\ 
\cline{13-18}
                                       &                             &                                                                         & \multicolumn{3}{c|}{}                                                                                                 & \multicolumn{3}{c|}{}                                                                                              & \multicolumn{3}{c|}{}                                                                                                                                & \multicolumn{3}{c|}{Re-training}                                                     & \multicolumn{3}{c}{FIUn}                                                                                                                                                                                   \\ 
\cline{4-18}
                                       &                             &                                                                         & \multicolumn{1}{c|}{$w_g$}            & \multicolumn{1}{c|}{$w_a$}            & \multicolumn{1}{c|}{$w_b$}            & \multicolumn{1}{c|}{$w_g$}           & \multicolumn{1}{c|}{$w_a$}           & \multicolumn{1}{c|}{$w_b$}           & \multicolumn{1}{c|}{$w_g$}                         & \multicolumn{1}{c|}{$w_a$}                 & $w_b$                                              & \multicolumn{1}{c|}{$w_g$} & \multicolumn{1}{c|}{$w_a$} & \multicolumn{1}{c|}{$w_b$} & \multicolumn{1}{c|}{$w_g$}                                           & \multicolumn{1}{c|}{$w_a$}                                  & $w_b$                                                                 \\ 
\hline
\multirow{6}{*}{\rotcell{DenseNet161}} & \multirow{2}{*}{60\%}       & $AD_r\uparrow$                                                          & 99.99                                 & 99.99                                 & 99.99                                 & 99.99                                & 99.99                                & 99.99                                & {\cellcolor[rgb]{1,1,0.882}}\textbf{76.86}         & {\cellcolor[rgb]{1,1,0.882}}\textbf{89.47} & {\cellcolor[rgb]{1,1,0.882}}\textbf{82.53}         & \multirow{2}{*}{994.32}    & \multirow{2}{*}{998.46}    & \multirow{2}{*}{1000.66}   & {\cellcolor[rgb]{1,1,0.882}}                                         & {\cellcolor[rgb]{1,1,0.882}}                                & {\cellcolor[rgb]{1,1,0.882}}                                          \\
                                       &                             & \multicolumn{1}{l|}{{\cellcolor[rgb]{0.882,0.882,1}}${AD}_f\downarrow$} & {\cellcolor[rgb]{0.882,0.882,1}}99.99 & {\cellcolor[rgb]{0.882,0.882,1}}99.99 & {\cellcolor[rgb]{0.882,0.882,1}}99.99 & {\cellcolor[rgb]{0.882,0.882,1}}0.00 & {\cellcolor[rgb]{0.882,0.882,1}}0.00 & {\cellcolor[rgb]{0.882,0.882,1}}0.00 & {\cellcolor[rgb]{1,1,0.882}}\textbf{\textbf{0.00}} & {\cellcolor[rgb]{1,1,0.882}}\textbf{0.00}  & {\cellcolor[rgb]{1,1,0.882}}\textbf{\textbf{0.00}} &                            &                            &                            & \multirow{-2}{*}{{\cellcolor[rgb]{1,1,0.882}}\textbf{\textbf{4.69}}} & \multirow{-2}{*}{{\cellcolor[rgb]{1,1,0.882}}\textbf{2.39}} & \multirow{-2}{*}{{\cellcolor[rgb]{1,1,0.882}}\textbf{\textbf{2.30}}}  \\ 
\hhline{~-----------------}
                                       & \multirow{2}{*}{40\%}       & $AD_r\uparrow$~                                                         & 99.99                                 & 99.99                                 & 99.50                                 & 99.99                                & 99.99                                & 99.99                                & {\cellcolor[rgb]{1,1,0.882}}\textbf{74.03}         & {\cellcolor[rgb]{1,1,0.882}}\textbf{89.47} & {\cellcolor[rgb]{1,1,0.882}}\textbf{83.18}         & \multirow{2}{*}{1010.45}   & \multirow{2}{*}{1011.14}   & \multirow{2}{*}{1009.53}   & {\cellcolor[rgb]{1,1,0.882}}                                         & {\cellcolor[rgb]{1,1,0.882}}                                & {\cellcolor[rgb]{1,1,0.882}}                                          \\
                                       &                             & \multicolumn{1}{l|}{{\cellcolor[rgb]{0.882,0.882,1}}${AD}_f\downarrow$} & {\cellcolor[rgb]{0.882,0.882,1}}99.99 & {\cellcolor[rgb]{0.882,0.882,1}}99.99 & {\cellcolor[rgb]{0.882,0.882,1}}99.99 & {\cellcolor[rgb]{0.882,0.882,1}}0.00 & {\cellcolor[rgb]{0.882,0.882,1}}0.00 & {\cellcolor[rgb]{0.882,0.882,1}}0.00 & {\cellcolor[rgb]{1,1,0.882}}\textbf{\textbf{0.00}} & {\cellcolor[rgb]{1,1,0.882}}\textbf{0.00}  & {\cellcolor[rgb]{1,1,0.882}}\textbf{\textbf{0.00}} &                            &                            &                            & \multirow{-2}{*}{{\cellcolor[rgb]{1,1,0.882}}\textbf{\textbf{4.12}}} & \multirow{-2}{*}{{\cellcolor[rgb]{1,1,0.882}}\textbf{2.69}} & \multirow{-2}{*}{{\cellcolor[rgb]{1,1,0.882}}\textbf{\textbf{2.14}}}  \\ 
\hhline{~-----------------}
                                       & \multirow{2}{*}{20\%}       & $AD_r\uparrow$                                                          & 99.99                                 & 99.99                                 & 99.99                                 & 99.99                                & 99.99                                & 99.99                                & {\cellcolor[rgb]{1,1,0.882}}\textbf{73.48}         & {\cellcolor[rgb]{1,1,0.882}}\textbf{89.13} & {\cellcolor[rgb]{1,1,0.882}}\textbf{83.78}         & \multirow{2}{*}{1021.34}   & \multirow{2}{*}{1020.67}   & \multirow{2}{*}{1022.64}   & {\cellcolor[rgb]{1,1,0.882}}                                         & {\cellcolor[rgb]{1,1,0.882}}                                & {\cellcolor[rgb]{1,1,0.882}}                                          \\
                                       &                             & \multicolumn{1}{l|}{{\cellcolor[rgb]{0.882,0.882,1}}${AD}_f\downarrow$} & {\cellcolor[rgb]{0.882,0.882,1}}99.99 & {\cellcolor[rgb]{0.882,0.882,1}}99.99 & {\cellcolor[rgb]{0.882,0.882,1}}99.91 & {\cellcolor[rgb]{0.882,0.882,1}}0.00 & {\cellcolor[rgb]{0.882,0.882,1}}0.00 & {\cellcolor[rgb]{0.882,0.882,1}}0.00 & {\cellcolor[rgb]{1,1,0.882}}\textbf{\textbf{0.00}} & {\cellcolor[rgb]{1,1,0.882}}\textbf{0.00}  & {\cellcolor[rgb]{1,1,0.882}}\textbf{\textbf{0.00}} &                            &                            &                            & \multirow{-2}{*}{{\cellcolor[rgb]{1,1,0.882}}\textbf{\textbf{4.98}}} & \multirow{-2}{*}{{\cellcolor[rgb]{1,1,0.882}}\textbf{2.64}} & \multirow{-2}{*}{{\cellcolor[rgb]{1,1,0.882}}\textbf{\textbf{2.59}}}  \\ 
\hline
\multirow{6}{*}{\rotcell{ResNet18}}    & \multirow{2}{*}{60\%}       & $AD_r\uparrow$                                                          & 99.99                                 & 99.99                                 & 99.99                                 & 99.99                                & 99.99                                & 99.99                                & {\cellcolor[rgb]{1,1,0.882}}\textbf{95.16}         & {\cellcolor[rgb]{1,1,0.882}}\textbf{95.92} & {\cellcolor[rgb]{1,1,0.882}}\textbf{98.86}         & \multirow{2}{*}{123.74}    & \multirow{2}{*}{126.43}    & \multirow{2}{*}{149.32}    & {\cellcolor[rgb]{1,1,0.882}}                                         & {\cellcolor[rgb]{1,1,0.882}}                                & {\cellcolor[rgb]{1,1,0.882}}                                          \\
                                       &                             & \multicolumn{1}{l|}{{\cellcolor[rgb]{0.882,0.882,1}}${AD}_f\downarrow$} & {\cellcolor[rgb]{0.882,0.882,1}}99.99 & {\cellcolor[rgb]{0.882,0.882,1}}99.99 & {\cellcolor[rgb]{0.882,0.882,1}}99.99 & {\cellcolor[rgb]{0.882,0.882,1}}0.00 & {\cellcolor[rgb]{0.882,0.882,1}}0.00 & {\cellcolor[rgb]{0.882,0.882,1}}0.00 & {\cellcolor[rgb]{1,1,0.882}}\textbf{\textbf{0.00}} & {\cellcolor[rgb]{1,1,0.882}}\textbf{0.00}  & {\cellcolor[rgb]{1,1,0.882}}\textbf{\textbf{0.00}} &                            &                            &                            & \multirow{-2}{*}{{\cellcolor[rgb]{1,1,0.882}}\textbf{\textbf{3.16}}} & \multirow{-2}{*}{{\cellcolor[rgb]{1,1,0.882}}\textbf{1.65}} & \multirow{-2}{*}{{\cellcolor[rgb]{1,1,0.882}}\textbf{\textbf{1.68}}}  \\ 
\hhline{~-----------------}
                                       & \multirow{2}{*}{40\%}       & $AD_r\uparrow$                                                          & 99.99                                 & 99.99                                 & 99.99                                 & 99.99                                & 99.99                                & 99.99                                & {\cellcolor[rgb]{1,1,0.882}}\textbf{94.43}         & {\cellcolor[rgb]{1,1,0.882}}\textbf{95.92} & {\cellcolor[rgb]{1,1,0.882}}\textbf{98.27}         & \multirow{2}{*}{128.43}    & \multirow{2}{*}{127.64}    & \multirow{2}{*}{155.32}    & {\cellcolor[rgb]{1,1,0.882}}                                         & {\cellcolor[rgb]{1,1,0.882}}                                & {\cellcolor[rgb]{1,1,0.882}}                                          \\
                                       &                             & \multicolumn{1}{l|}{{\cellcolor[rgb]{0.882,0.882,1}}${AD}_f\downarrow$} & {\cellcolor[rgb]{0.882,0.882,1}}99.99 & {\cellcolor[rgb]{0.882,0.882,1}}99.99 & {\cellcolor[rgb]{0.882,0.882,1}}99.99 & {\cellcolor[rgb]{0.882,0.882,1}}0.00 & {\cellcolor[rgb]{0.882,0.882,1}}0.00 & {\cellcolor[rgb]{0.882,0.882,1}}0.00 & {\cellcolor[rgb]{1,1,0.882}}\textbf{\textbf{0.00}} & {\cellcolor[rgb]{1,1,0.882}}\textbf{0.00}  & {\cellcolor[rgb]{1,1,0.882}}\textbf{\textbf{0.00}} &                            &                            &                            & \multirow{-2}{*}{{\cellcolor[rgb]{1,1,0.882}}\textbf{\textbf{3.97}}} & \multirow{-2}{*}{{\cellcolor[rgb]{1,1,0.882}}\textbf{1.46}} & \multirow{-2}{*}{{\cellcolor[rgb]{1,1,0.882}}\textbf{\textbf{1.59}}}  \\ 
\hhline{~-----------------}
                                       & \multirow{2}{*}{20\%}       & $AD_r\uparrow$                                                          & 99.99                                 & 99.99                                 & 99.99                                 & 99.96                                & 99.99                                & 99.99                                & {\cellcolor[rgb]{1,1,0.882}}\textbf{94.71}         & {\cellcolor[rgb]{1,1,0.882}}\textbf{95.92} & {\cellcolor[rgb]{1,1,0.882}}\textbf{98.48}         & \multirow{2}{*}{129.35}    & \multirow{2}{*}{126.43}    & \multirow{2}{*}{154.83}    & {\cellcolor[rgb]{1,1,0.882}}                                         & {\cellcolor[rgb]{1,1,0.882}}                                & {\cellcolor[rgb]{1,1,0.882}}                                          \\
                                       &                             & \multicolumn{1}{l|}{{\cellcolor[rgb]{0.882,0.882,1}}${AD}_f\downarrow$} & {\cellcolor[rgb]{0.882,0.882,1}}99.99 & {\cellcolor[rgb]{0.882,0.882,1}}99.99 & {\cellcolor[rgb]{0.882,0.882,1}}99.99 & {\cellcolor[rgb]{0.882,0.882,1}}0.00 & {\cellcolor[rgb]{0.882,0.882,1}}0.00 & {\cellcolor[rgb]{0.882,0.882,1}}0.00 & {\cellcolor[rgb]{1,1,0.882}}\textbf{\textbf{0.00}} & {\cellcolor[rgb]{1,1,0.882}}\textbf{0.00}  & {\cellcolor[rgb]{1,1,0.882}}\textbf{\textbf{0.00}} &                            &                            &                            & \multirow{-2}{*}{{\cellcolor[rgb]{1,1,0.882}}\textbf{\textbf{3.89}}} & \multirow{-2}{*}{{\cellcolor[rgb]{1,1,0.882}}\textbf{1.54}} & \multirow{-2}{*}{{\cellcolor[rgb]{1,1,0.882}}\textbf{\textbf{1.38}}}  \\
\hline
\end{tabular}}
\vspace{-1em}
\end{table*}

 




\vfill

\end{document}